\documentclass[journal]{IEEEtran}
\usepackage{times}
\usepackage{epsfig}
\usepackage{graphicx}
\usepackage{amsmath}
\usepackage{amssymb}
\usepackage{multirow}
\usepackage{diagbox}
\usepackage{algorithm, algorithmic}
\usepackage{booktabs}
\usepackage[pagebackref=true,breaklinks=true,letterpaper=true,colorlinks,bookmarks=false]{hyperref}

\ifCLASSINFOpdf
\else
\fi
\begin{document}
\title{Progressive Deep Video Dehazing without Explicit Alignment Estimation}

\author{Runde Li \\
School of Computer Science and Engineering, Nanjing University of Science and
Technology, China.

}


%Journal of \LaTeX\ Class Files,~Vol.~14, No.~8, August~2019
\markboth{}%
{Shell \MakeLowercase{\textit{et al.}}: Bare Demo of IEEEtran.cls for IEEE Journals}



\maketitle

\begin{abstract}
To solve the issue of video dehazing, there are two main tasks to attain: how to align adjacent frames to the reference frame; how to restore the reference frame.
Some papers adopt explicit approaches (e.g., the Markov random field, optical flow, deformable convolution, 3D convolution) to align neighboring frames with the reference frame in feature space or image space, they then use various restoration methods to achieve the final dehazing results.
In this paper, we propose a progressive alignment and restoration method for video dehazing.
The alignment process aligns consecutive neighboring frames stage by stage without using the optical flow estimation.
The restoration process is not only implemented under the alignment process but also  uses a refinement network to improve the dehazing performance of the whole network.
The proposed networks include four fusion networks and one refinement network.
To decrease the parameters of networks, three fusion networks in the first fusion stage share the same parameters.
Extensive experiments demonstrate that the proposed video dehazing method achieves outstanding performance against the-state-of-art methods.
\end{abstract}


\IEEEpeerreviewmaketitle

\section{Introduction}

\IEEEPARstart{V}{ideos} captured in outdoor scenes usually suffer from severe degradation due to hazy weather.
Thus, video dehazing is of great significance to surveillance, traffic monitoring, autonomous driving and so on.
The imaging process of the hazy frame can refer to that of single hazy image.
The atmospheric scattering model~\cite{DBLP:journals/tog/Fattal08,DBLP:conf/cvpr/NarasimhanN00,DBLP:journals/ijcv/NarasimhanN02,DBLP:journals/pami/He0T11} is usually used to describe the physical process of the hazy frame, which is shown as follow:
\begin{equation}
	\label{eq:haze process}
	{\emph{I}}(x)={\emph{J}}(x)t(x)+{\emph{A(x)}}(1-t(x))
\end{equation}
Where ${\emph{I}}(x)$ and $\emph{{J}}(x)$ denote the hazy frame and corresponding haze-free frame, respectively.
$\emph{{A}}(x)$ is the global airlight and $t(x)$ the medium transmission map.
The hazy video is considered as the concatenation of all adjacent haze frames.
However, video dehazing is different from single image dehazing, which needs to remove haze from frames and deal with the spatial and temporal information existing in consecutive adjacent frames.
Thus,  there are two issues to solve: how to remove haze from each frame; how to align adjacent frames to the reference frame.

There are two main video restoration methods including separately solving the restoration and alignment, jointly solving the restoration and alignment.
The method solving the restoration and alignment separately  first adopt various alignment approaches (e.g., Markov random field, optical flow, deformable convolution, recurrent neural networks) to wrap the adjacent frames to the reference frame. Then they use the restoration methods to recover the aligned reference frame~\cite{DBLP:journals/ijcv/XueCWWF19,DBLP:journals/tip/XiangWP20,DBLP:conf/cvpr/WangCYDL19,DBLP:conf/cvpr/WangCYDL19,DBLP:conf/eccv/ZhongGZZ20}.
For example, Cai et al.~\cite{DBLP:conf/pcm/CaiXT16} build the Markov random field (MRF) model based on the intensity value prior to address the spatial and temporal information on the adjacent three frames. They then restore the reference frame by optimizing the MRF likelihood function.
Xiang et al.~\cite{DBLP:journals/tip/XiangWP20} first use the optical flow to estimate the motion information between two frames and warp them to the reference frame. They then use a residual network to restore the aligned reference frame. The method not only increases model parameters resulting from the optical flow estimation network but also decreases the performance of the trained model when the motion information is not accurately estimated.
Wang et al.~\cite{DBLP:conf/cvpr/WangCYDL19} propose to use the enhanced deformable convolution with the offset estimation to align adjacent frames to the reference frame in feature space and use the temporal and spatial attention module to fuse these features. They then use a reconstruction module to restore the reference frame. But the method is not accurate to estimate the offset when the large motion or the non-homogeneous haze appears.
Zhong et al.~\cite{DBLP:conf/eccv/ZhongGZZ20} first use the recurrent neural network and a global spatial and temporal attention module to align the adjacent frames to the reference frame, they then adopt a reconstruction network to restore the aligned reference frames.

The other method solves the restoration and alignment jointly to align the adjacent frames to the reference frame and restore the reference frame in the same process or by an end-to-end network~\cite{DBLP:conf/cvpr/CaballeroLAATWS17,DBLP:conf/cvpr/LiHDZXT19,DBLP:journals/tip/ZhangLZMLL19}.
For example, Zhang et al.~\cite{DBLP:journals/tip/ZhangLZMLL19} adopt the 3D convolution to jointly deal with the spatial and temporal information by slowly fusing multiple consecutive video frames, but the 3D convolution significantly increases the computer complex and memory which decreases the efficiency of processing videos.
Li et al.~\cite{DBLP:conf/aaai/LiPWXF18} propose to transfer three adjacent frames into feature space by three different convolution layers and cascade these features, followed by sending them into a dehazing network. The method makes the alignment process of adjacent frames and the restoration process of the reference frame into an end-to-end trainable network. But the single-stage process is not able to restore haze videos in severe degradation.

\begin{figure*}[t]
	\footnotesize
	\begin{center}
		\begin{tabular}{cccc}
			%\vspace{-0.2cm}
			\includegraphics[width = 0.24\linewidth, height = 0.2\linewidth]{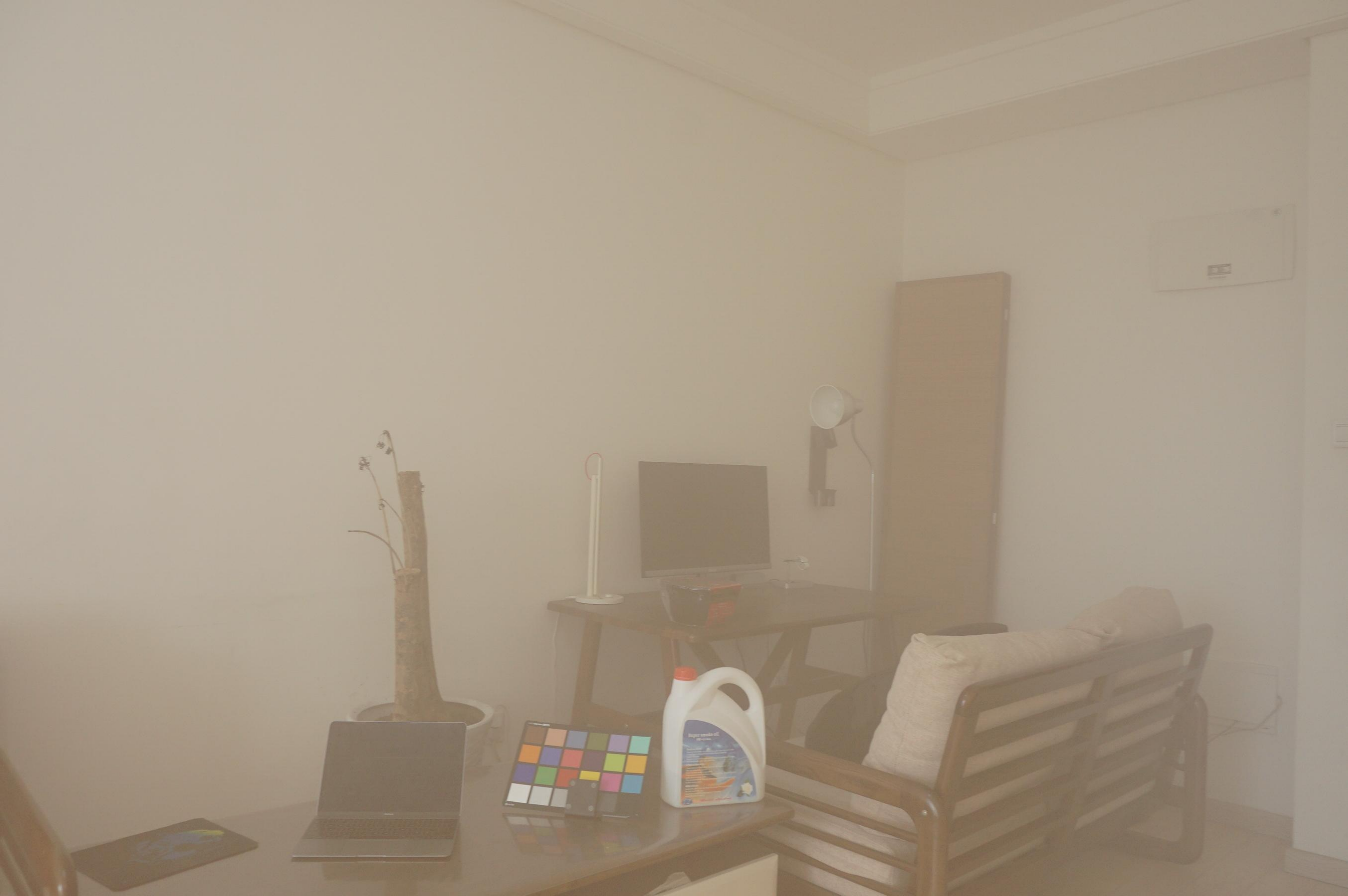}& \hspace{-0.46cm}
			\includegraphics[width = 0.24\linewidth, height = 0.2\linewidth]{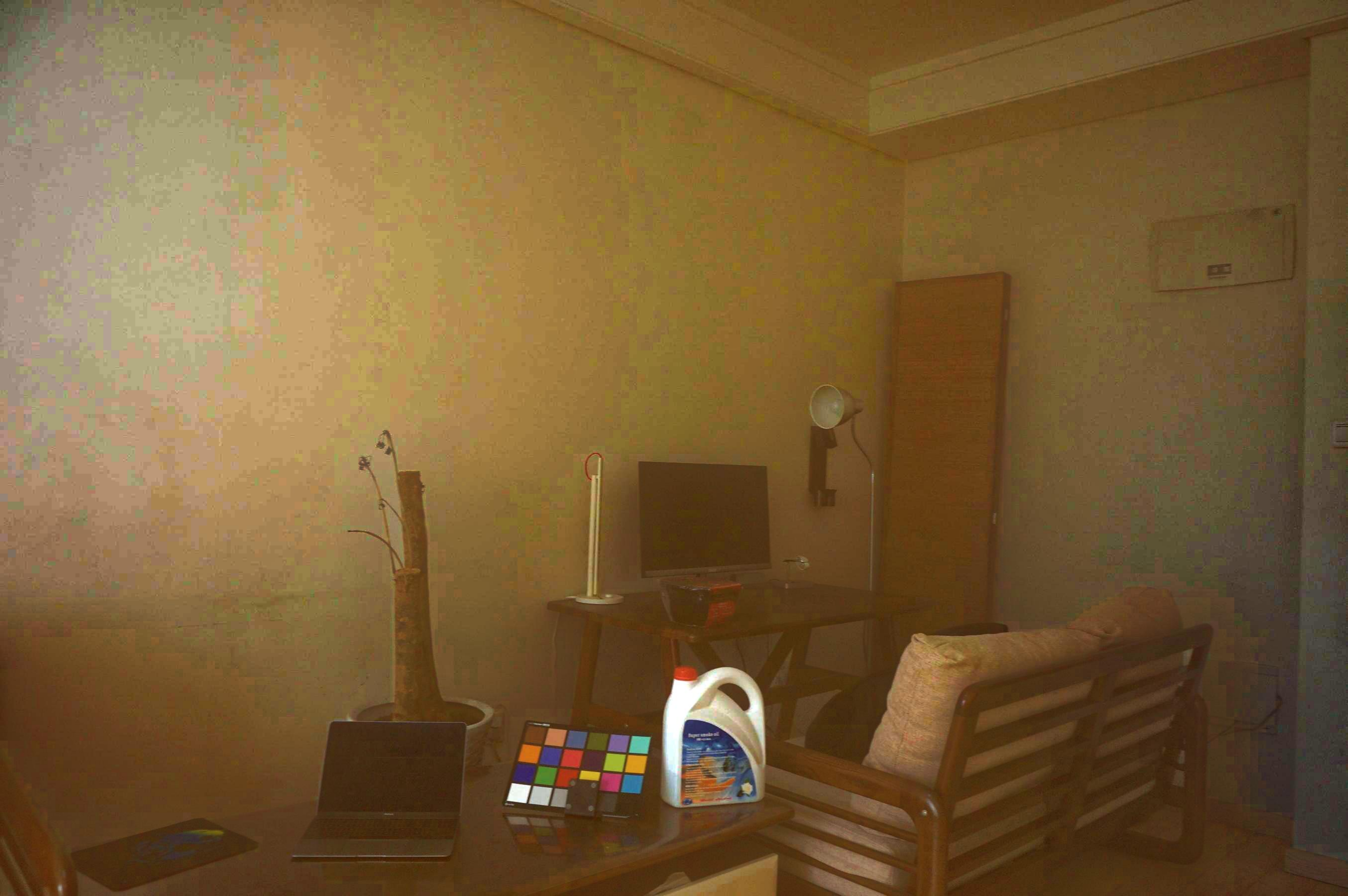}  & \hspace{-0.46cm}
			\includegraphics[width = 0.24\linewidth, height = 0.2\linewidth]{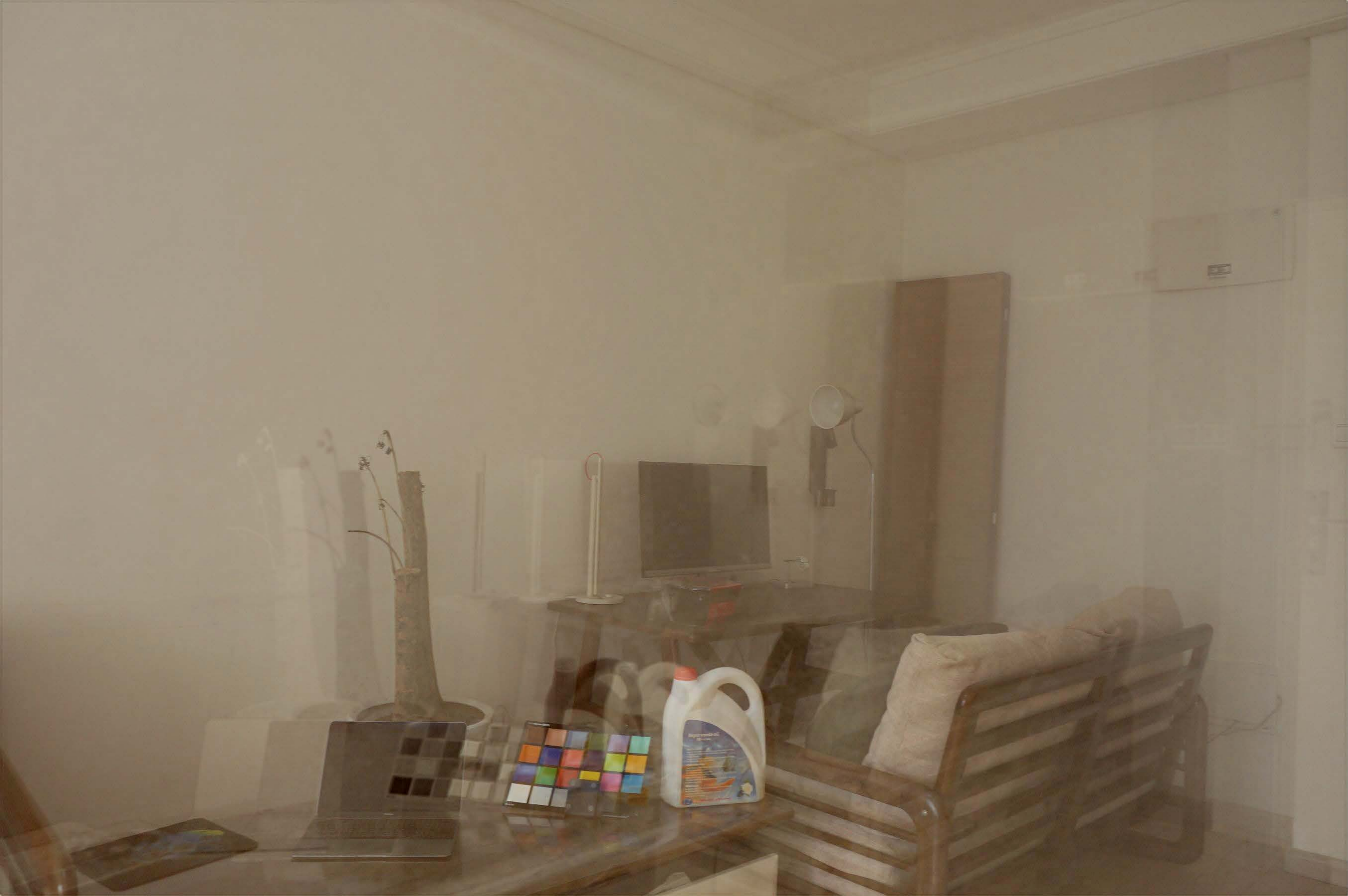} & \hspace{-0.46cm}
			\includegraphics[width = 0.24\linewidth, height = 0.2\linewidth]{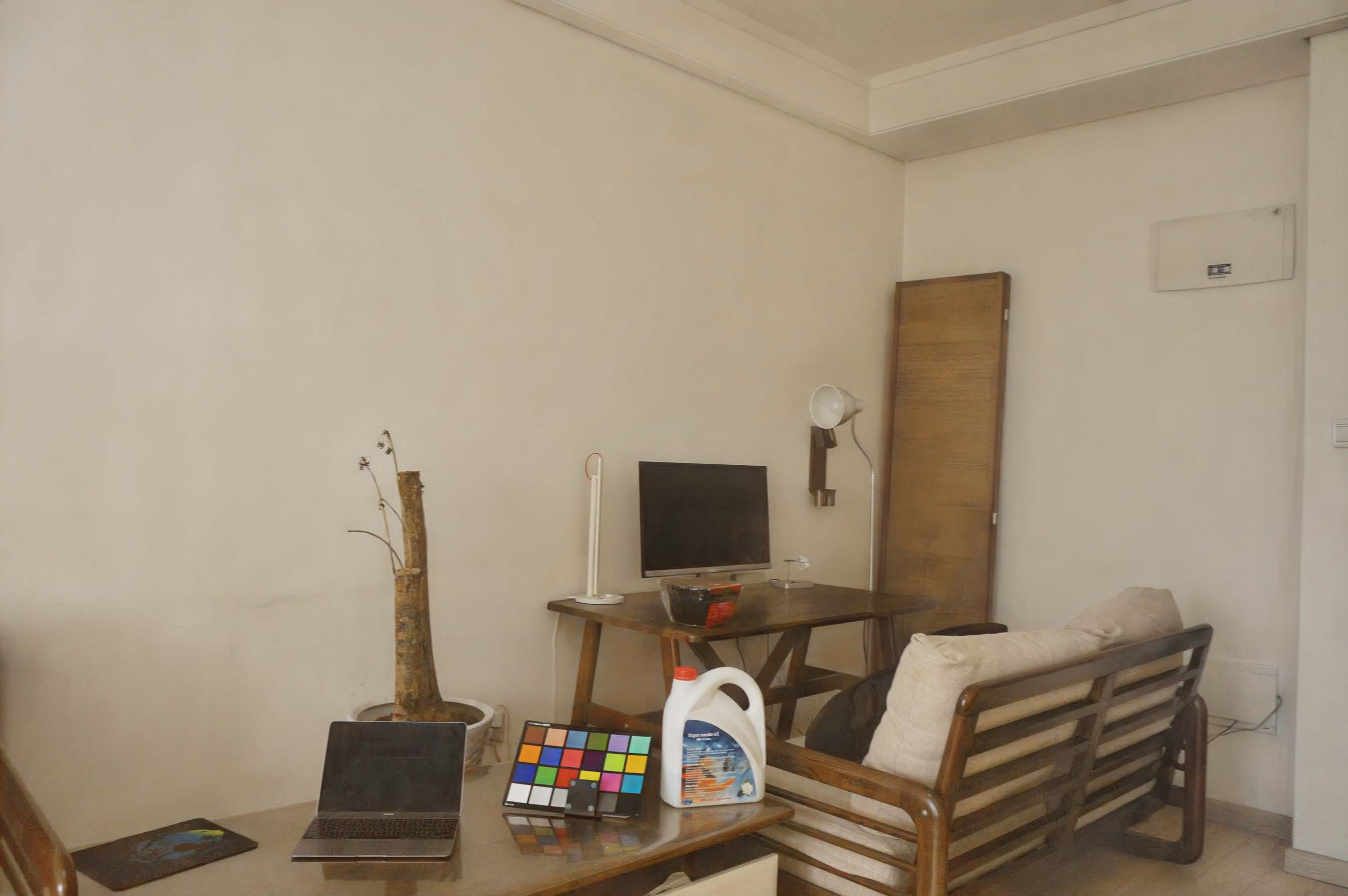}\\
			(a) Original  & \hspace{-0.46cm} (b) DCP~\cite{DBLP:journals/pami/He0T11} & \hspace{-0.46cm} (c) EDVNet~\cite{DBLP:conf/aaai/LiPWXF18} & \hspace{-0.46cm} (d) Ours \\
		\end{tabular}
	\end{center}
	%\vspace{-0.5cm}
	\caption{Examples from real video frame and corresponding dehazed frames by the state-of-the-art methods. Our dehazed frame contains fewer hazy residuals and artifacts.}
	\label{fig:demo}
\end{figure*}

In this paper, we aim to develop a video dehazing method in jointly solving alignment and restoration manner without using optical flow estimation.
Specifically, we intend to align the adjacent frames to the reference frame and restore the reference frame by an end-to-end network.
Inspired by the video restoration method~\cite{DBLP:conf/cvpr/TassanoDV20}, we develop a progressive video dehazing network to achieve the alignment and restoration processes, which fuses consecutive adjacent frames and removes haze stage by stage. Then it refines the preliminary dehazed frame by a refinement network.
Fig.~\ref{fig:demo} shows a real haze frame and corresponding dehazed frames by different methods. Our dehazed frame contain fewer haze residuals and artifacts.

The contributions of the work are as follows:
\begin{itemize}
	\item We propose an end-to-end video dehazing method to align neighboring frames to the reference frame and remove haze stage by stage without using the optical flow estimation.
	
	\item The proposed networks include four fusion networks and one refinement network.
	To decrease the parameters of networks, three fusion networks in the first fusion stage share the same parameters.
	
	\item Extensive experiments demonstrate that the proposed video dehazing method achieves favorable performance against the-state-of-art methods.
\end{itemize}

%-------------------------------------------------------------------------
\section{Related Work}

To address the problem of video dehazing, a direct approach is to use the image dehazing methods to restore the hazy video frame by frame.
For example, Chen et al.~\cite{DBLP:conf/eccv/ChenDW16} propose to use the gradient residual minimization to constrain the estimation of the transmission map, which helps to suppress the artifacts and halo. They then estimate the global atmospheric light as the method~\cite{DBLP:journals/pami/He0T11}. The dehazed image or frame is achieved according to the atmospheric scattering model.
Although the method is effective to remove haze on images or frames, the flickering artifacts appear on the dehazed videos when the camera shakes or the scene changes rapidly.
To exploit the temporal information from consecutive adjacent frames, some prior-based video dehazing methods~\cite{DBLP:journals/vc/ZhangLZYCS11,DBLP:conf/eccv/ChenDW16,DBLP:journals/jvcir/KimJSK13,DBLP:conf/vcip/YuQXC16,DBLP:conf/cvip/DasPSVS18,DBLP:conf/skima/UllahM19} introduce the temporal information of adjacent frames on the estimation process of the transmission map.
For example, Zhang et al.~\cite{DBLP:journals/vc/ZhangLZYCS11} first use the guided filter to extract the transmission map of each frame. They then adopt the optical flow to find the pixel changes of  two adjacent frames. To improve the spatial and temporal coherence, a  Markov random field model of the transmission map is built on the flow fields. The dehazed frames are finally achieved by the atmospheric scattering model.
Kim et al.~\cite{DBLP:journals/jvcir/KimJSK13} use the block-based transmission estimation method to improve the image dehazing effect.
When they deal with hazy videos, they first transfer each frame into YUV space and only restore the Y channel to reduce computation complexity. To estimate the relationship between the  current frame and the previous frame, they introduce a temporal coherence factor to correct the transmission map of the current frame. This method is benefit to suppress flickering artifacts.
Cai et al.~\cite{DBLP:conf/pcm/CaiXT16} build the Markov random field (MRF) model based on the intensity value prior to address the spatial and temporal information on the adjacent three frames. They then restore the reference frame by optimizing the MRF likelihood function.

As the development of the deep learning-based methods, various neural networks are introduced to address the issue of image and video dehazing.
For example, Ren et al~\cite{DBLP:conf/pcm/RenC17} use a convolution neural network (CNN) to estimate the transmission map from consecutive haze frames in an end-to-end trainable manner. They first cascade five adjacent haze frames as the input of the CNN, and it then outputs three consecutive transmission maps which are directly used to estimate three adjacent dehazed frames.
Although this method uses the temporal information from consecutive haze frames to suppress the flickering of the transmission maps, the dehazed frames may contain some artifacts due to fast scene change or big relative motion.
To improve the dehazing effect of each frame, they then provide more spatial information into the CNN by  embedding a semantic map of each frame~\cite{DBLP:journals/tip/RenZXMCML19}.
To effectively use the temporal information and suppress flickering artifacts, Li et al.~\cite{DBLP:conf/aaai/LiPWXF18} develop three approaches to fuse consecutive adjacent frames and demonstrate the K-Level fusion method is the best fusion structure for their proposed network. They then concatenate a dehazing network to restore the reference frame.
This method fuses the alignment process and restoration process by an end-to-end network, but it contains some hazy residuals in dense haze scenes.
In addition, some deep learning-based methods~\cite{DBLP:conf/cvpr/SuDWSHW17,DBLP:journals/tip/XiangWP20,DBLP:conf/cvpr/WangCYDL19,DBLP:conf/cvpr/CaballeroLAATWS17,DBLP:conf/cvpr/LiHDZXT19,DBLP:journals/tip/ZhangLZMLL19} employ other approaches (e.g., optical flow, deformable convolution, 3D convolution, recurrent neural network) to align frames in solving low-level vision issues.
Xiang et al.~\cite{DBLP:journals/tip/XiangWP20} first use the optical flow to estimate the motion information between two adjacent frames and warp them to the current frame. They are then stacked in channels and sent a CNN to restore blurred videos.
Wang et al.~\cite{DBLP:conf/cvpr/WangCYDL19} use the deformable convolution to align the five adjacent frames in feature space.
They also propose a temporal and spatial attention fusion module to jointly deal with the temporal and spatial information.
Zhang et al.~\cite{DBLP:journals/tip/ZhangLZMLL19} adopt the 3D convolution to jointly deal with the spatial and temporal information by slowly fusing multiple consecutive video frames, but the 3D convolution significantly increases the computer complex and memory which decreases the efficiency of processing videos.
The recurrent neural network (RNN) is an alternative tool to deal with the temporal information, which recursively passes the information of consecutive adjacent frames from the previous frame to the current frame and then the next frame~\cite{DBLP:conf/iccv/WieschollekHSL17,DBLP:conf/iccv/KimLSH17,DBLP:conf/eccv/ZhongGZZ20}.

Different from the methods mentioned above, we propose an end-to-end video dehazing method to align neighboring frames to the reference frame and remove haze stage by stage without using the optical flow estimation. To improve the dehazing performance, we also introduce the refinement network to further remove hazy residuals.

%%%%%%%%%%%%%%%%%%%%%%%%%%%%%%%%%%%%%%
\section{Proposed Algorithm}

In this section, we first describe the overview process of the proposed deep video dehazing method.
Furthermore, we show the proposed network architecture stage by stage and each sub-network in detail.
Finally, we describe the loss functions used to constrain  the training process of the proposed network.
\begin{figure*}[!htb]
	%\footnotesize
	\begin{center}
		\begin{tabular}{c}
			%\vspace{-0.2cm}
			\includegraphics[width = 0.95\linewidth,height = 0.5\linewidth]{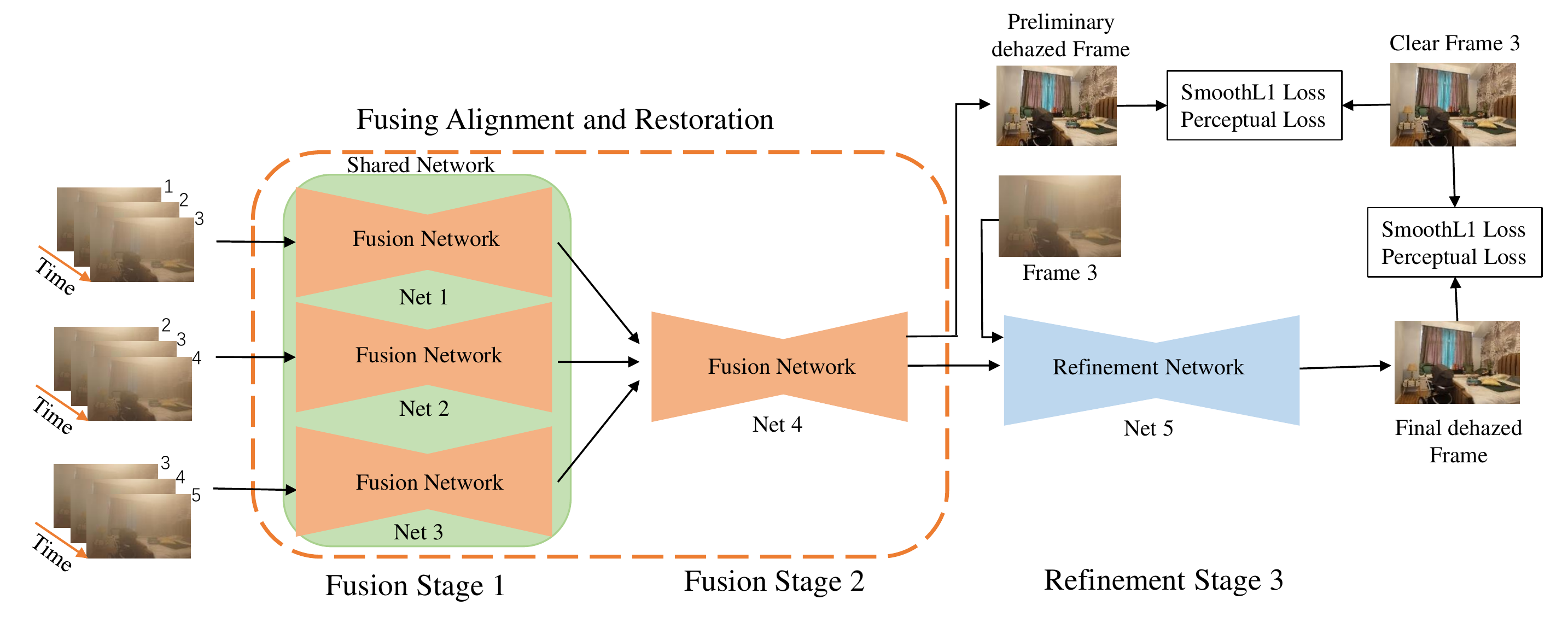}
		\end{tabular}
	\end{center}
	%\vspace{-1cm}
	\caption{The flow chart of the proposed progressive video dehazing method. In the first stage, the five consecutive adjacent frames are sequentially split into three groups.  Frames of each group are cascaded and sent into the fusion networks (Nets 1, 2, 3) in the first stage. In the second stage, three adjacent frames from the first stage are sent into another fusion network (Net 4), which achieves the alignment process and preliminary restoration process. In the third stage, the refinement network (Net 5) further removes haze residuals for the preliminary dehazed result.}
	\label{fig:model}
\end{figure*}

\subsection{Overview}

Restoring the current frame or referent frame $f_{t}$ from a hazy video, we need to not only remove the haze from the reference frame but also address the spatial and temporal information of consecutive adjacent frames.
We suppose  a consecutive frame set as $\{f_{t-n},..., f_{t-1}, f_{t}, f_{t+1},..., f_{t+n}\}$, where t denotes the current time, {t-n, n=1,2,3,...,N} denotes the past time, {t+n, n=1,2,3,...,N} denotes the future time. $f_{t-n}$ denotes the past $n-th$ frame, $f_{t+n}$ the future $n-th$ frame.
$O^{n}_{t}$ denotes the output of the $n-th$ stage of the proposed network.
In this paper, we adopt five consecutive adjacent frames to form a time unit.
To address the spatial and temporal coherence,  we fuse a time unit by an implicitly multi-stage alignment manner.
To be specific, the five consecutive adjacent frames are sequentially separated into three groups, and each group includes three consecutive adjacent frames.
%Since the maximum value of the hazy image is usually considered as the global atmospheric light~\cite{DBLP:journals/pami/He0T11}, we use the bright channel of the reference frame as the haze map $h_{t}$ to guide training process.
In the first fusion stage,  three consecutive frames of each group and the haze map are concatenated and sent into the corresponding fusion network $\mathcal{F}_{n}$. The fusion network then outputs one frame for each group. The process of the first stage is formulated as follow:

\begin{equation}
	\label{eq:stage1}
	\begin{split}
		&O^{1}_{t-1} = \mathcal{F}_{1}(f_{t-2}, f_{t-1}, f_{t}), \\
		&O^{1}_{t}   = \mathcal{F}_{2}(f_{t-1}, f_{t}, f_{t+1}), \\
		&O^{1}_{t+1} = \mathcal{F}_{3}(f_{t}, f_{t+1}, f_{t+2}).
	\end{split}
\end{equation}

In the second fusion stage, three frames from the first stage are also concatenated as the input of the fusion network $\mathcal{F}_{4}$. The fusion network outputs an aligned frame. The process of the second stage is formulated as follow:
%%%%
\begin{equation}
	\label{eq:stage2}
	\begin{split}
		O^{2}_{t} = \mathcal{F}_{4}(O^{1}_{t-1}, O^{1}_{t}, O^{1}_{t+1}).
	\end{split}
\end{equation}
The fusion networks aim to address  the spatial and temporal coherence and align the adjacent frames to the reference frame, but we also use them to remove haze for the reference frame.
Thus, the output $O^{2}_{t}$ of the second stage is used as the preliminary dehazing result.
In training process, we embed the loss functions to constrain the training of the fusion networks.

The third stage aims to refine the preliminary dehazing result and remove more haze residuals. Thus, the input of the refinement network $\mathcal{F}_{6}$ needs to fuse the output of the second stage and the reference frame.
The process of the third stage is formulated as follow:
\begin{equation}
	\label{eq:stage3}
	\begin{split}
		O^{3}_{t} = \mathcal{F}_{5}(O^{2}_{t}, f_{t}).
	\end{split}
\end{equation}

The proposed video dehazing process includes two fusion stages and one refinement stage. The fusion stages include four fusion networks used to align the adjacent frames to the reference frame and remove haze. The refinement stage uses a refinement network which are adopted to further improve the dehazing performance.

\subsection{Network Architecture}
\label{Networks}
\begin{figure}[!htp]
	%\footnotesize
	\begin{center}
		\begin{tabular}{c}
			%\vspace{-0.2cm}
			\includegraphics[width = 1\linewidth,height = 0.5\linewidth]{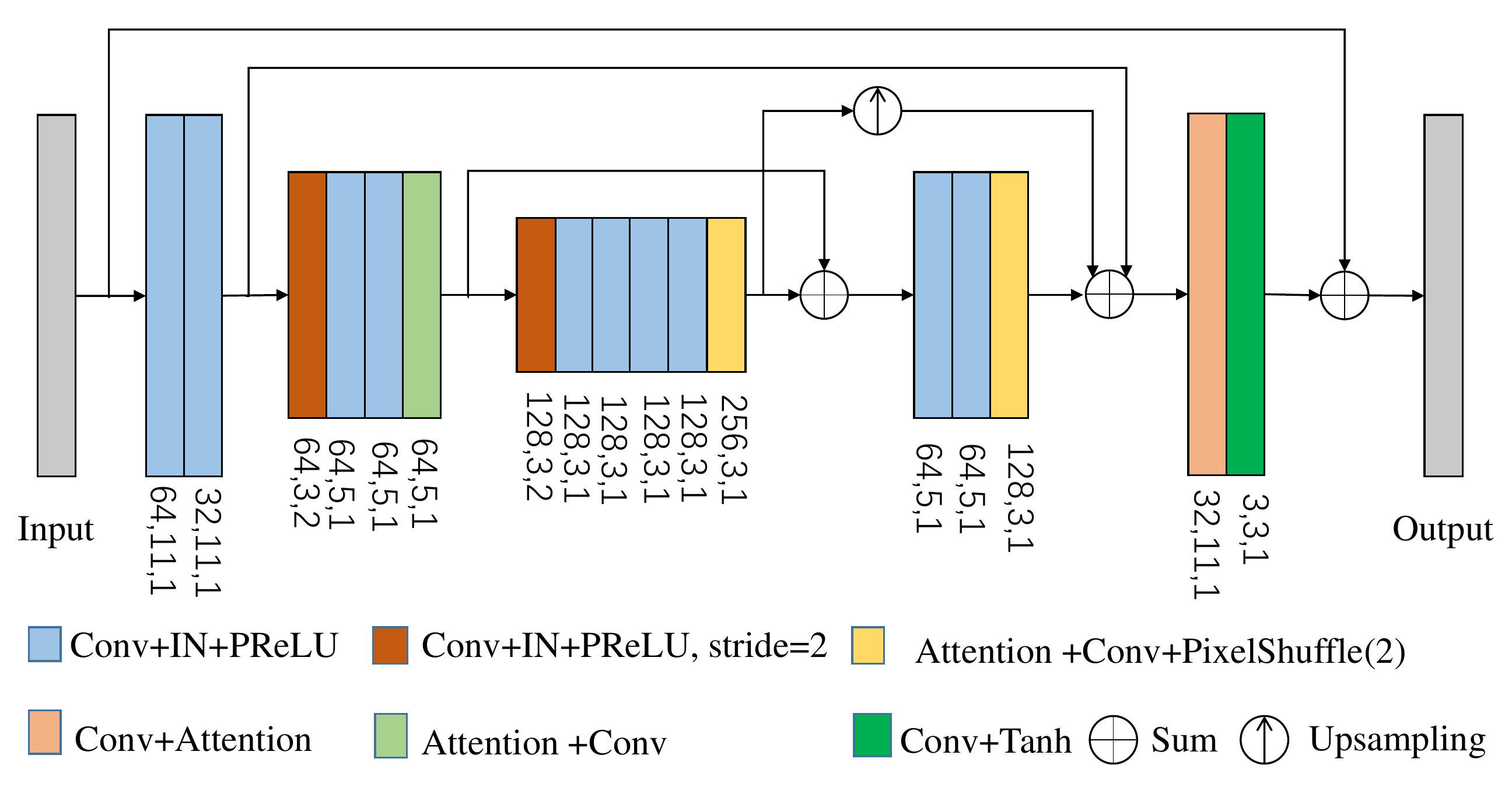}
		\end{tabular}
	\end{center}
	%\vspace{-1cm}
	\caption{The network architecture of the proposed UNet. The numerals bellow each module denote channel number, kernel size, stride of the convolution layer.}
	\label{fig:unet}
\end{figure}

As described above, the proposed method refers to five networks including four fusion networks (Nets 1, 2, 3, 4) and a refinement network (Net 5). The whole process is shown in Fig.~\ref{fig:model}. In the first fusion stage, there exist three fusion networks(Nets 1, 2, 3). Since they aim to achieve the similar fusion task, we adopt the parameter sharing strategy to reduce the parameters of the proposed networks. In other words, the three networks have the same network structure and share the same parameters.
In the second fusion stage, another fusion network (Net 4) is to fuse the outputs of the first fusion stage, which has the similar function as other fusion networks. Thus, we adopt the same network structure for four fusion networks.
The architecture of the fusion network is shown in Fig.~\ref{fig:unet}.
Because the U-Net structure~\cite{DBLP:conf/miccai/RonnebergerFB15} has been proved to help to align neighboring frames in video restoration~\cite{DBLP:conf/cvpr/TassanoDV20,DBLP:conf/eccv/WuXTT18}, we also develop the fusion network based on the U-Net architecture. In the encoder process, we first use a convolution layer with big kernel size to fast fuse consecutive adjacent frames and the haze map.
Except for the down-sampling operator, we also introduce an attention mechanism~\cite{DBLP:conf/nips/ChenKLYF18} to further increase the receptive field of the convolution layer.
In the decoder process, the PixelShuffle layer~\cite{DBLP:conf/cvpr/ShiCHTABRW16} is adopted to attain the up-sampling operator to avoid the grid artifacts.
To decrease the  memory requirements, we use the summation operation instead of the concatenation operation in the skip process.
We also introduce a multi-scale skip operator to fuse more scale features.
To decrease the difficulty of the training process, the middle frame is added with the output of the fusion network.

\begin{figure}[!htp]
	%\footnotesize
	\begin{center}
		\begin{tabular}{c}
			%\vspace{-0.2cm}
			\includegraphics[width = 1\linewidth,height = 0.6\linewidth]{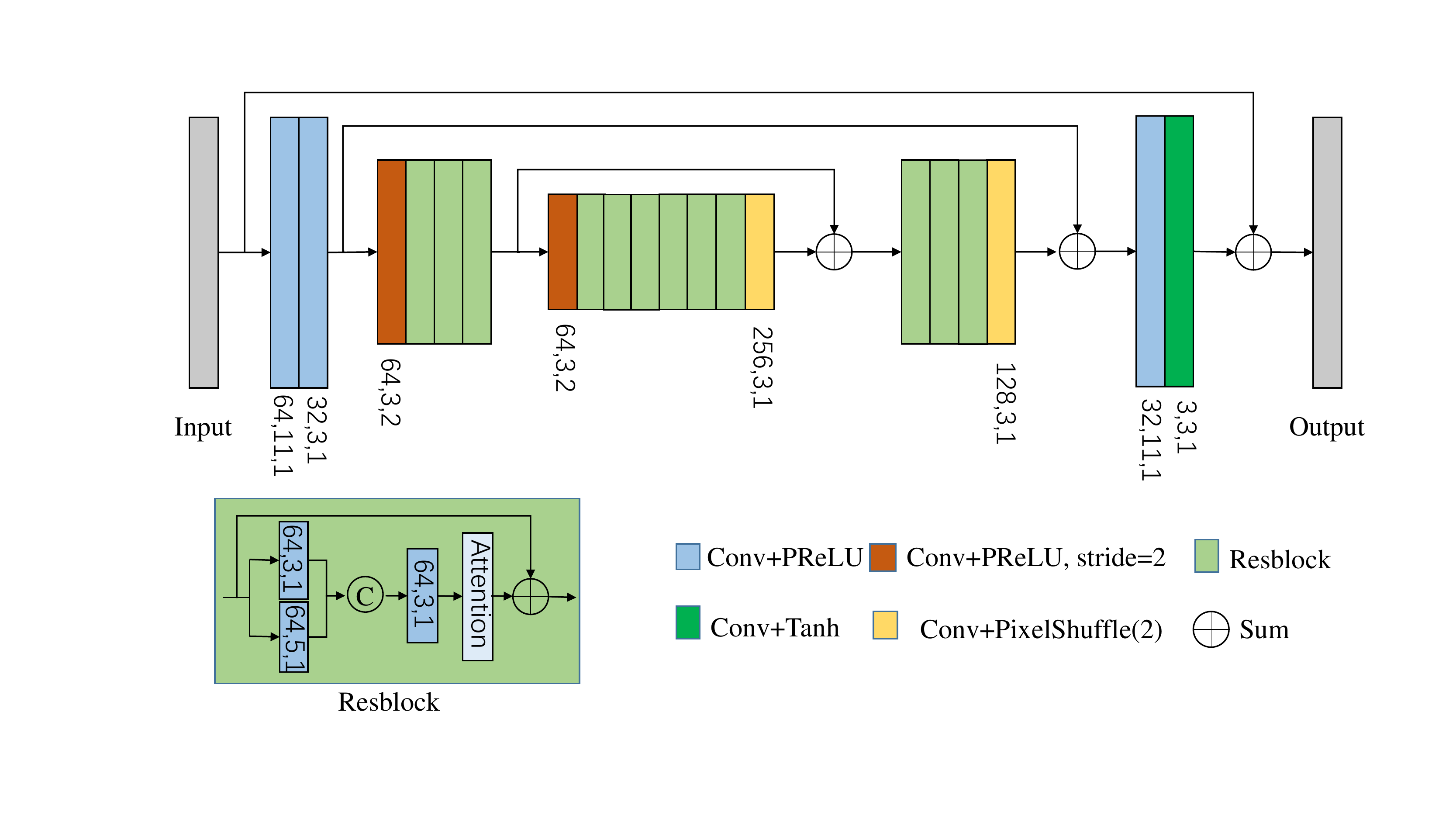}
		\end{tabular}
	\end{center}
	%\vspace{-1cm}
	\caption{The network architecture of the proposed ResNet. The numerals bellow each module denote channel number, kernel size, stride of the convolution layer.}
	\label{fig:resnet}
\end{figure}

The refinement network is introduced to further remove haze and restore more details for the reference frame. We also adopt the UNet architecture to build the refinement network.
The architecture of the refinement network is shown in Fig.~\ref{fig:resnet}.
Different from the fusion network, we adopt the residual blocks instead of some convolution layers and increase the network depth to improve the dehazing performance. The residual block introduces the multi-scale kernels and the attention mechanism to increase the receptive field.

\subsection{Loss Functions}
In the training process, we use the same loss functions to both the preliminary dehazing result $O^{2}_{t}$ and the final dehazing result $O^{3}_{t}$.
The smooth $L_{1}$ loss function and the perceptual loss are adopted to jointly constrain the training process of multi-stage networks.
The  smooth $L_{1}$ loss function is used as the pixel-wise constraint and shown as follow:
\begin{equation}
	\label{eq:loss-1}
	\begin{split}
		L_{SoomthL1}= \frac{1}{N}\sum^{N}_{x=1}\sum^{3}_{i=1}\phi(\tilde{J}_{i}(x) - J_{i}(x))
	\end{split}
\end{equation}
\begin{equation}
	\label{eq:loss-1-1}
	\phi(z)=\left\{
	\begin{aligned}
		&0.5z^{2}  , & if \mid z \mid<1, \\
		&\mid z \mid-0.5  , & otherwise.
	\end{aligned}
	\right.
\end{equation}
Here, $\tilde{J}_{i}(x)$ denotes the intensity of the $i$-th channel of pixel $x$ in the dehazed frame,  ${J}_{i}(x)$  the ground truth of the reference frame. $N$ is the total number of pixels.

The perceptual loss has been demonstrated to help to improve visual effect in many image restoration tasks, such as image super-resolution~\cite{DBLP:conf/eccv/JohnsonAF16}, image deblurring~\cite{DBLP:conf/cvpr/KupynBMMM18}, image dehazing~\cite{DBLP:conf/cvpr/LiPLT18}.
Thus, we also introduce the perceptual loss to constrain the training process of the proposed networks.
The perceptual loss function is shown as follow:

\begin{equation}
	\label{eq:loss-2}
	\begin{split}
		L_{P}=\frac{1}{N}\sum^{N}_{x=1}\sum^{M}_{i=1}\lambda_{i}\|\Phi_{i}(\tilde{J}(x)) - \Phi_{i}(J(x))\|_{1}
	\end{split}
\end{equation}
Here, $\Phi_{i}$, $M$,  $\lambda_{i}$ denote the $i$-th feature maps extracted from the pre-trained VGG\_19~\cite{DBLP:journals/corr/SimonyanZ14a},  the total number of the feature layers, the weight of the $i$-th feature layer, respectively.

The total loss function is shown as follow:
\begin{equation}
	\label{eq:loss-4}
	\begin{split}
		\mathcal{L}= \alpha L_{SoomthL1} + \beta L_{P}
	\end{split}
\end{equation}
Here, $\alpha$ and $\beta$ are  balance weights for each loss function.

%%%%%%%%%%%%%%%%%%%%%%%%%%%%%%%
\section{Experiments}

In this section, we first describe a dataset used to train and test the multi-scale deep video dehazing network and the experiment settings in detail.
Furthermore, we quantitatively and qualitatively evaluate the proposed method against the state-of-the-art methods. Finally, the ablation experiments are attained to demonstrate the effectiveness of the proposed method.

\subsection{Datasets}

\begin{figure}[!htp]
	\footnotesize
	\begin{center}
		\begin{tabular}{cccc}
			%\vspace{-0.2cm}
			\includegraphics[width = 0.24\linewidth, height = 0.21\linewidth]{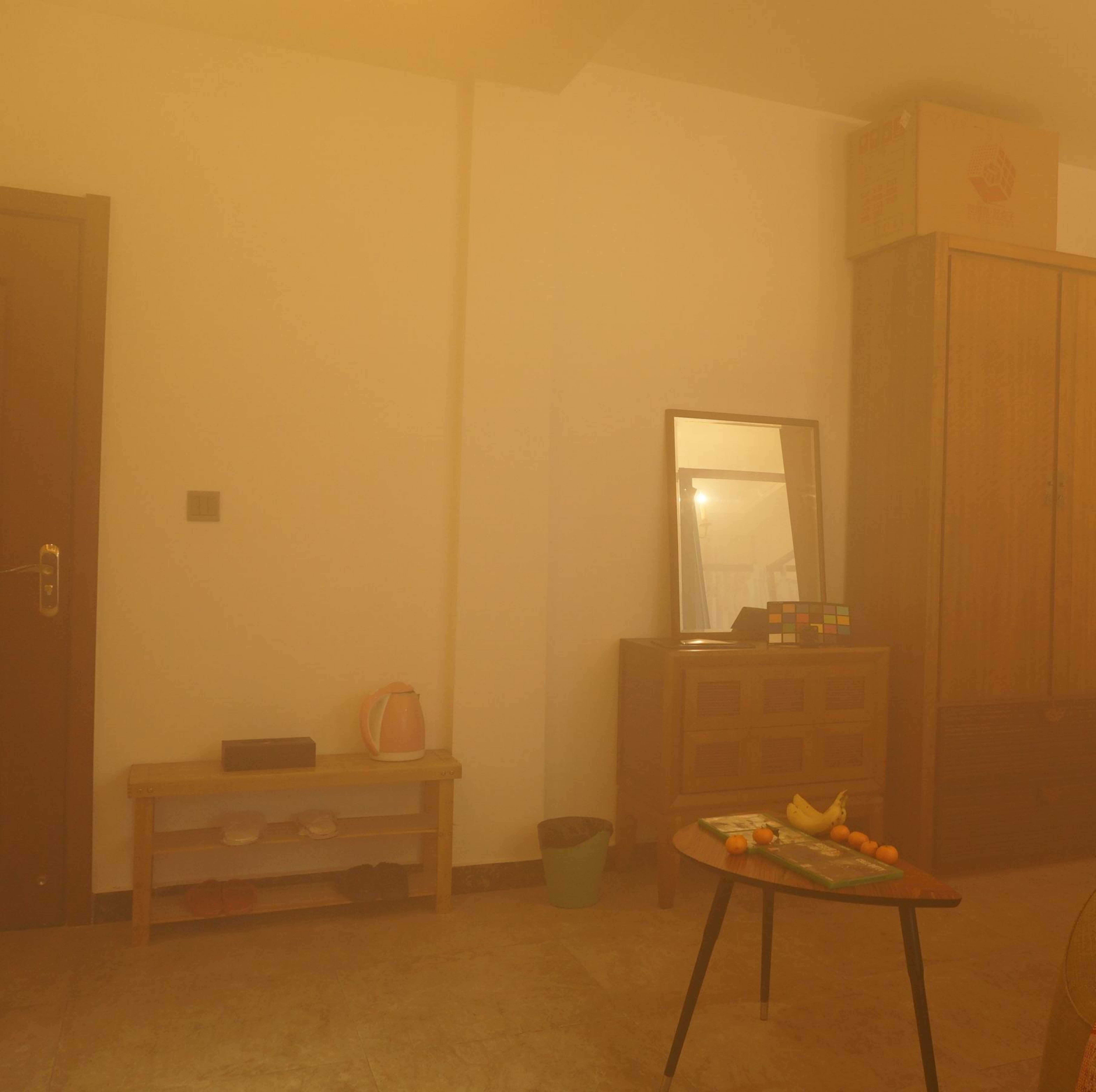} & \hspace{-0.46cm}
			\includegraphics[width = 0.24\linewidth, height = 0.21\linewidth]{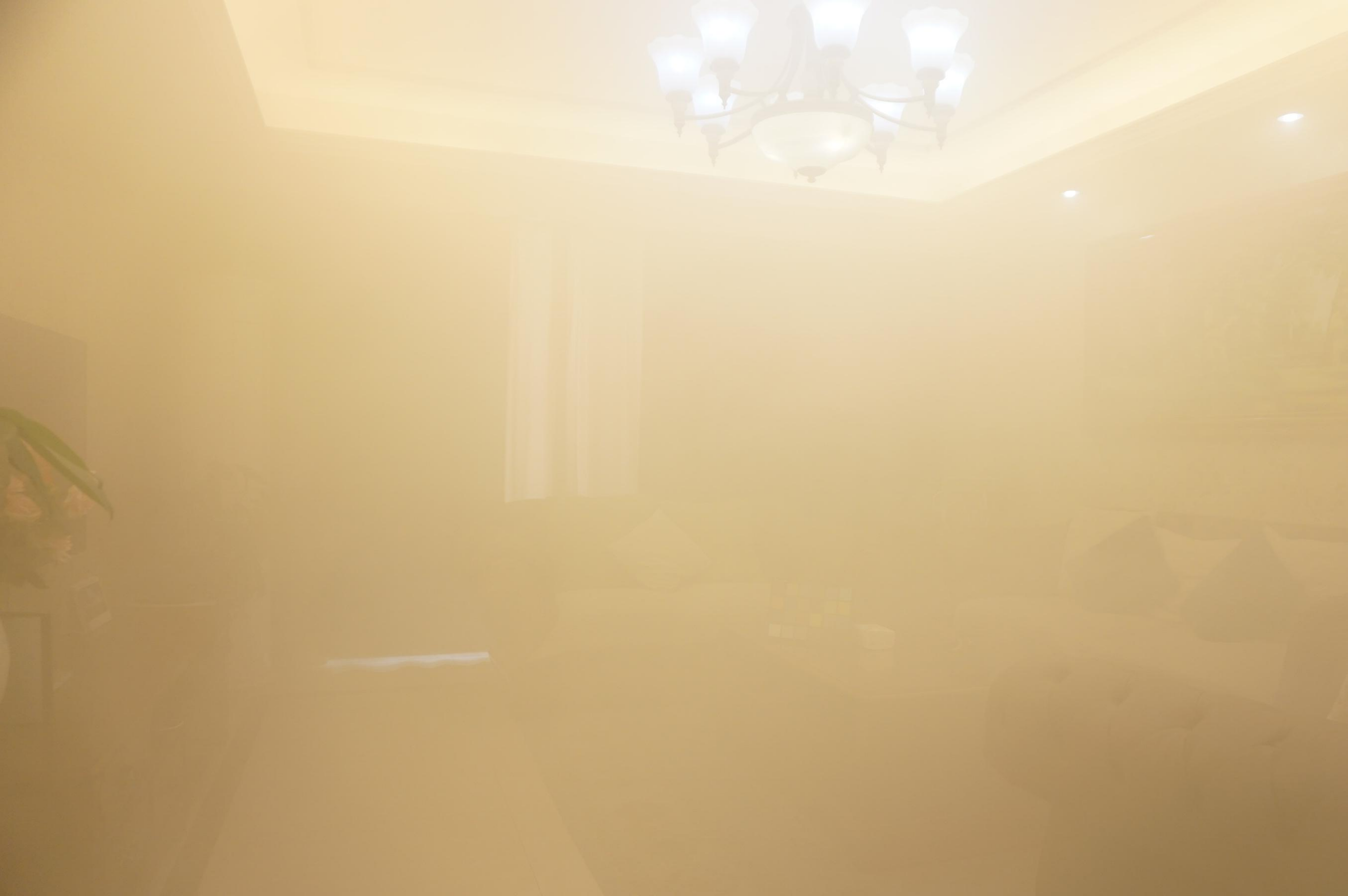} & \hspace{-0.46cm}
			\includegraphics[width = 0.24\linewidth, height = 0.21\linewidth]{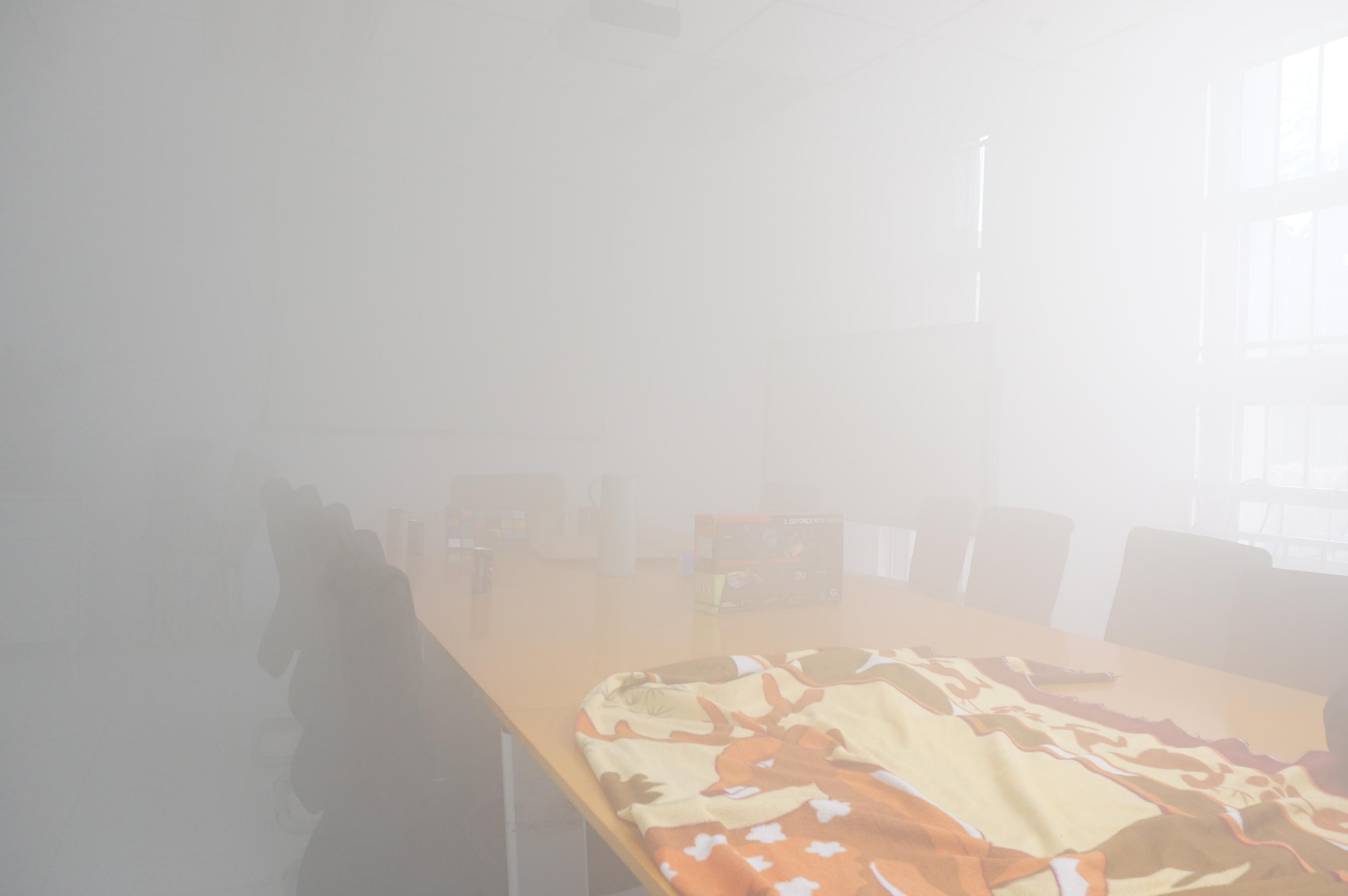} & \hspace{-0.46cm}
			\includegraphics[width = 0.24\linewidth, height = 0.21\linewidth]{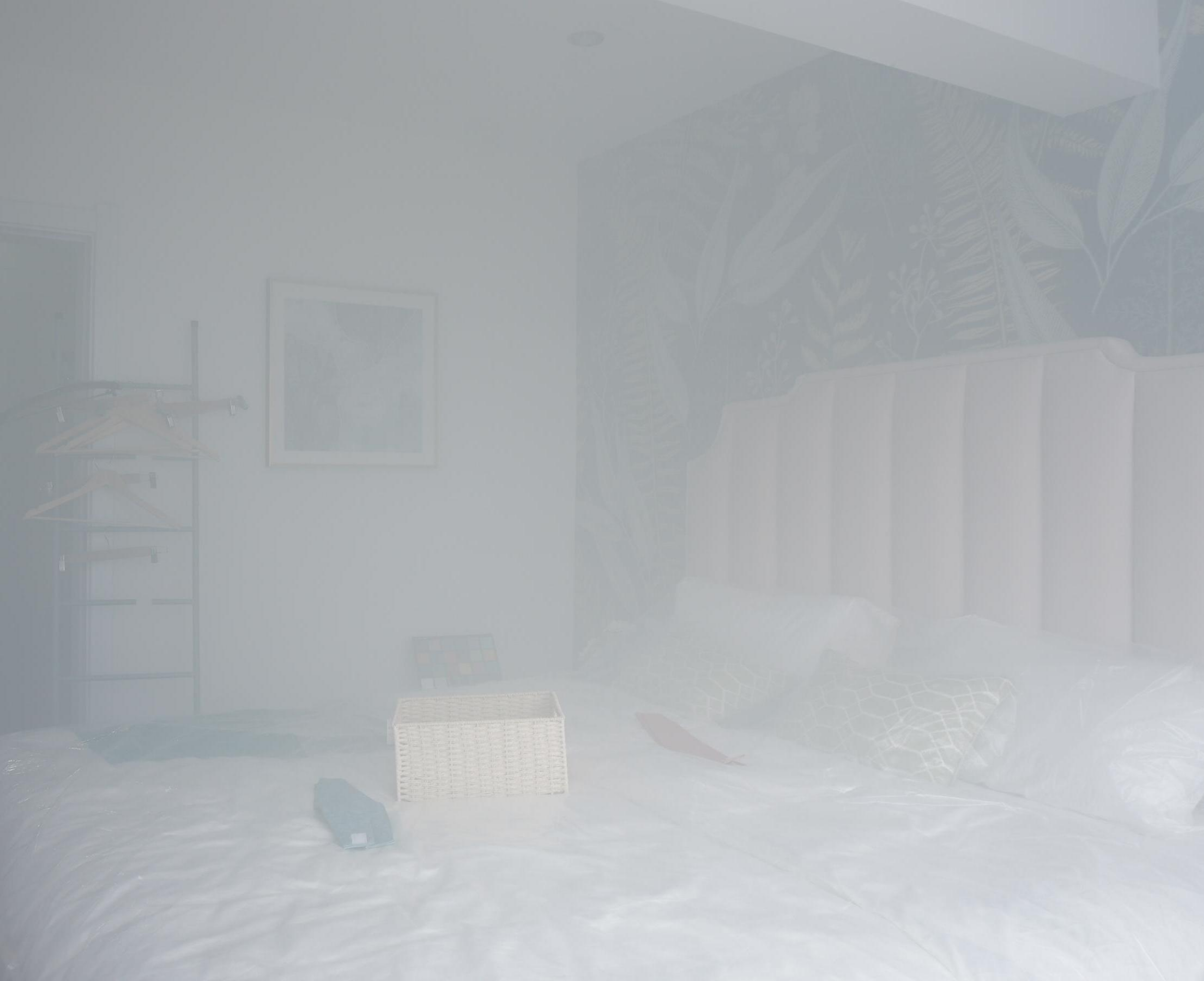} \\
			\includegraphics[width = 0.24\linewidth, height = 0.21\linewidth]{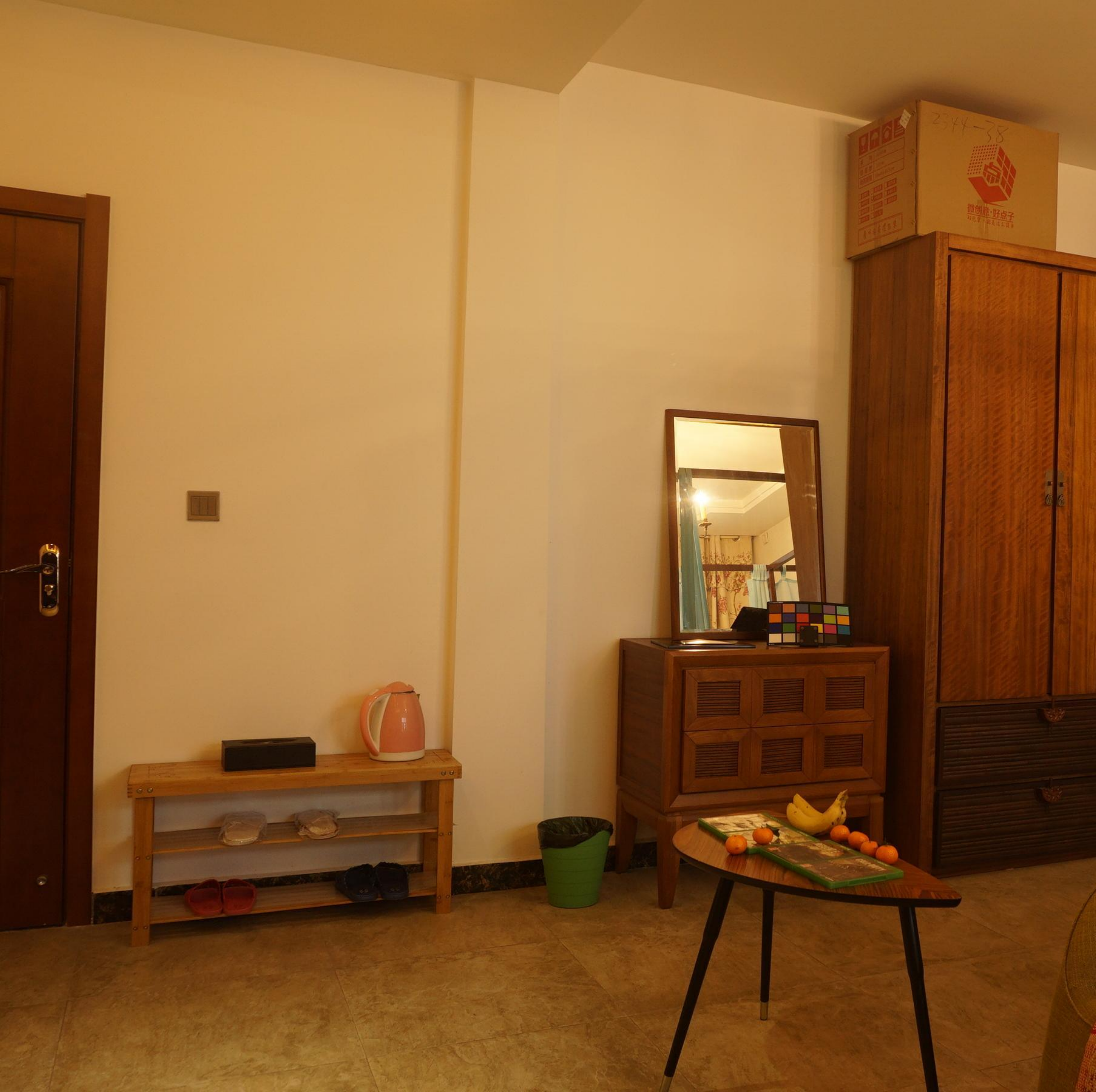} & \hspace{-0.46cm}
			\includegraphics[width = 0.24\linewidth, height = 0.21\linewidth]{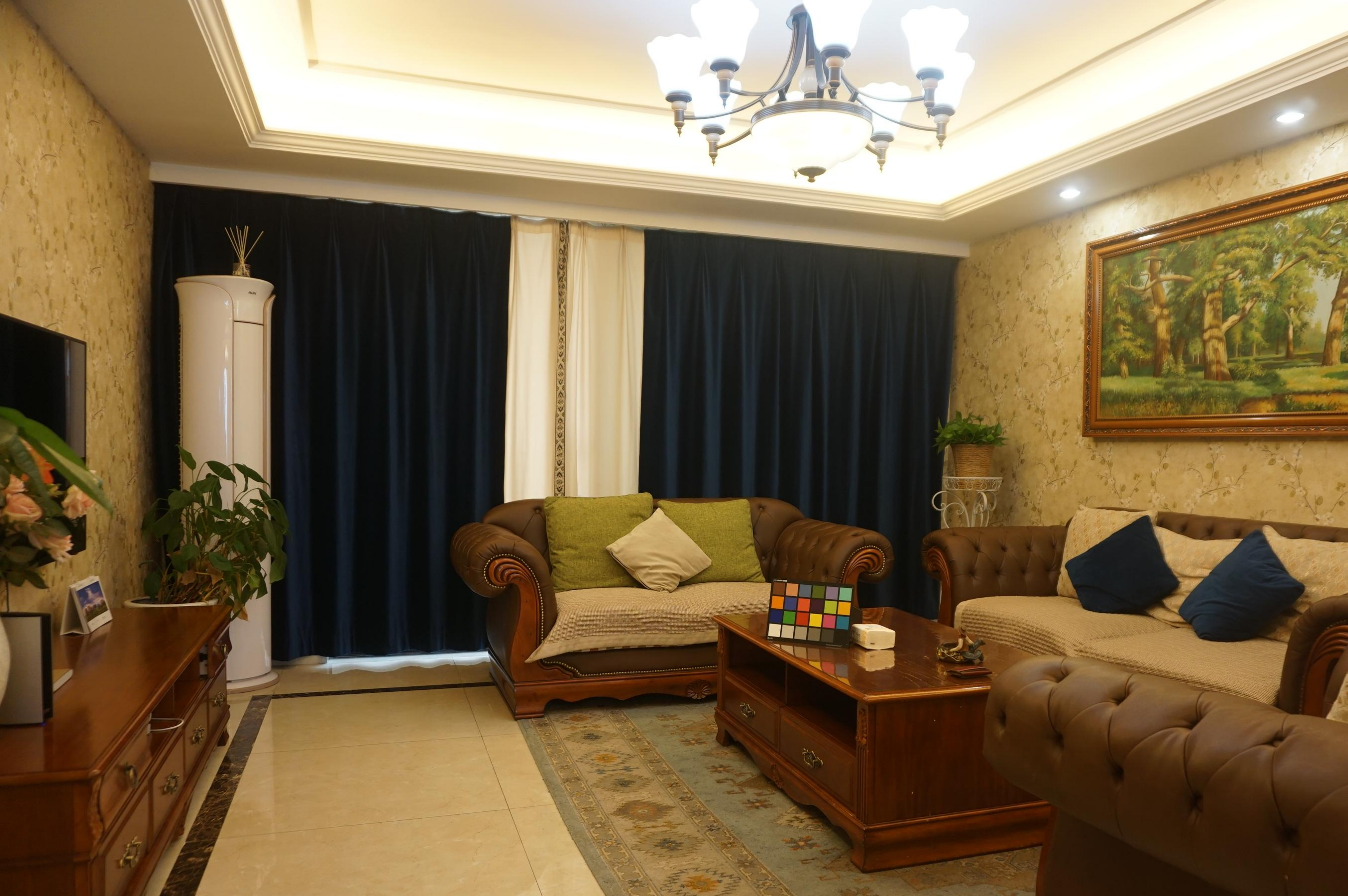} & \hspace{-0.46cm}
			\includegraphics[width = 0.24\linewidth, height = 0.21\linewidth]{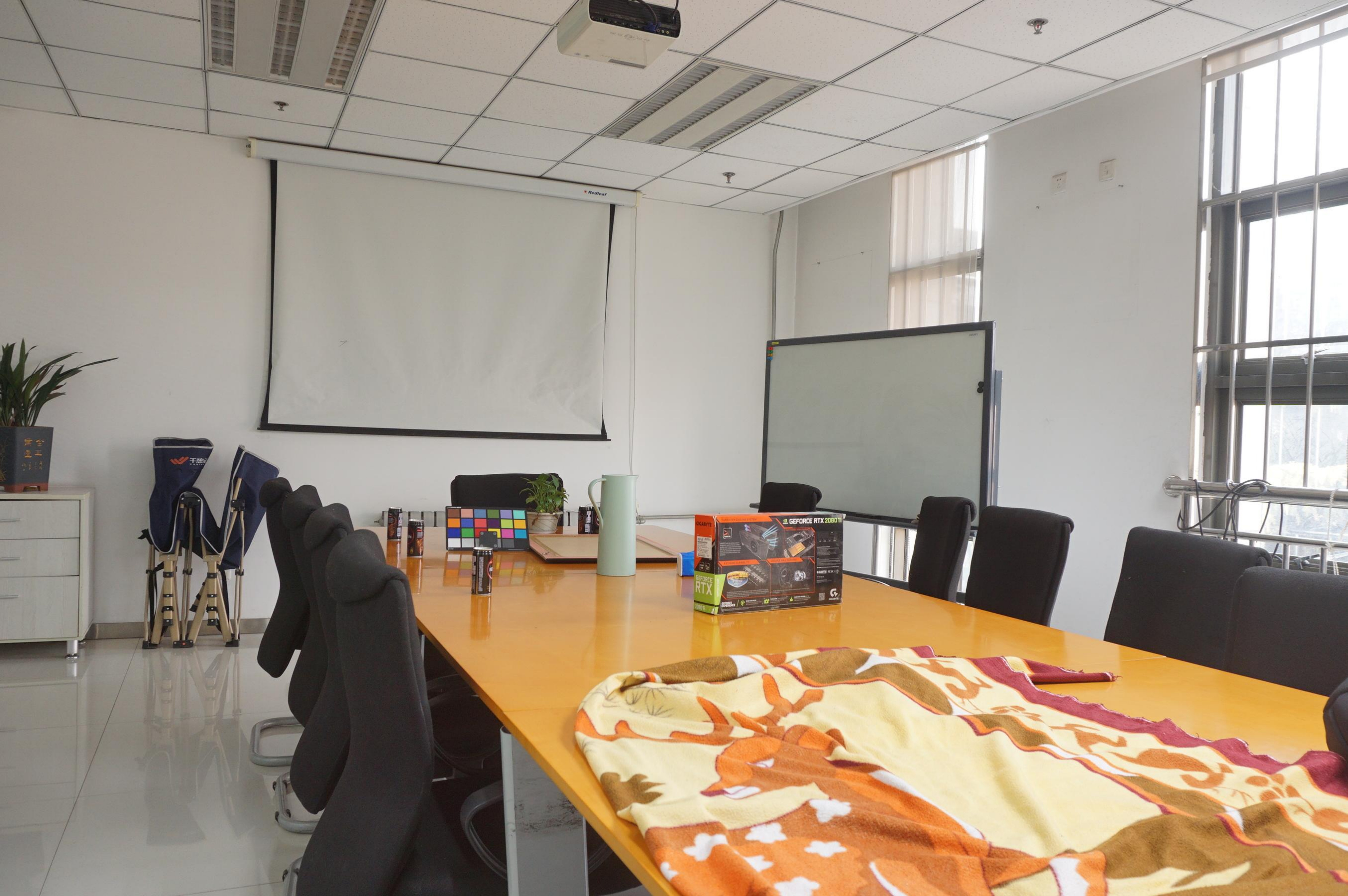} & \hspace{-0.46cm}
			\includegraphics[width = 0.24\linewidth, height = 0.21\linewidth]{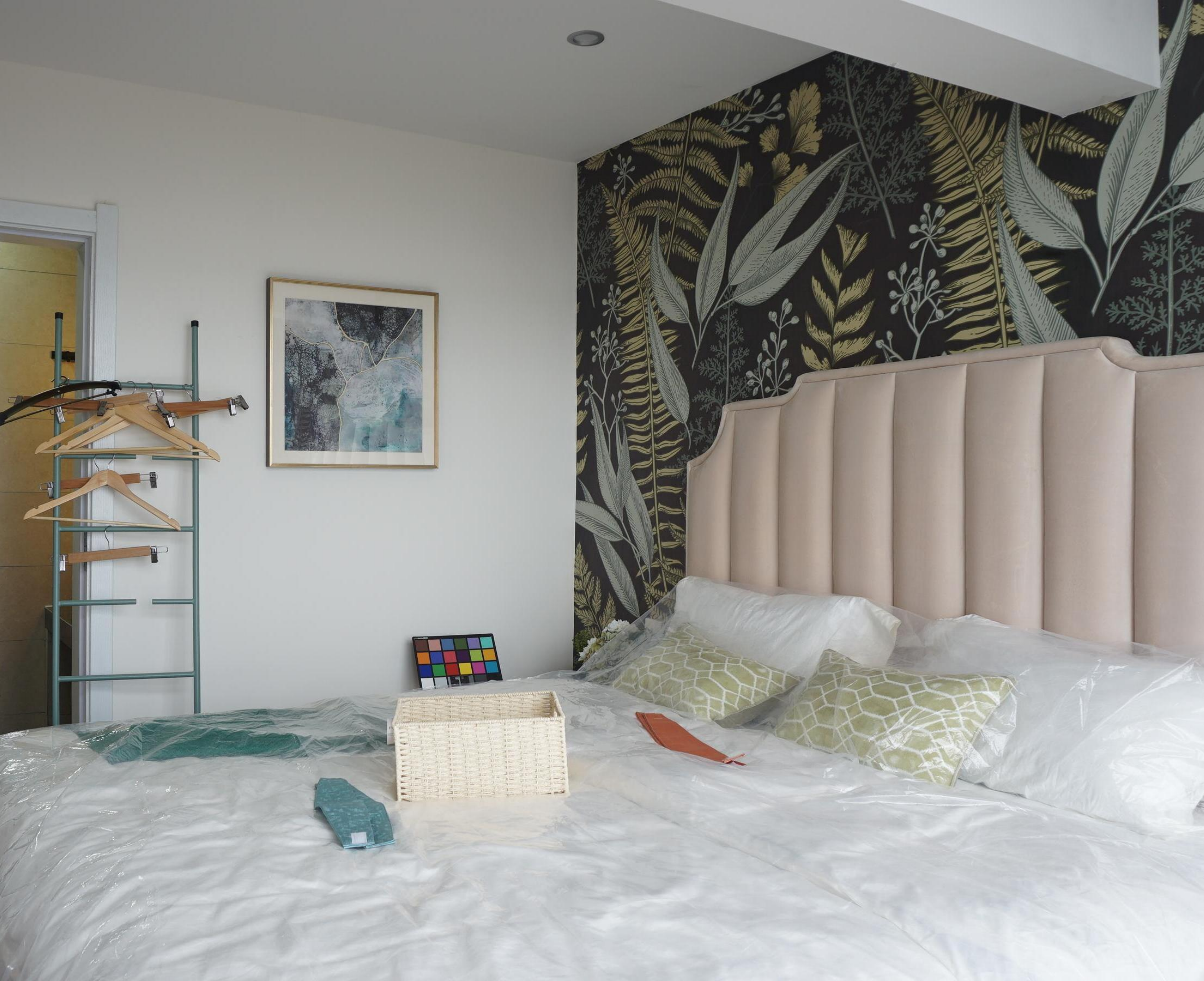} \\
			(a)   & \hspace{-0.36cm} (b)  &\hspace{-0.36cm} (c)  &\hspace{-0.36cm} (d) \\
		\end{tabular}
	\end{center}
	%\vspace{-0.3cm}
	\caption{Real hazy frames from the REVIDE dataset. The first row shows hazy frames from four different scenes. The second row shows corresponding clear frames.}
	%\vspace{-0.3cm}
	\label{fig:dataset}
\end{figure}

To train the proposed video dehazing network, we select a real haze video dataset: the REVIDE dataset~\cite{DBLP:conf/cvpr/videodehazing21}. The REVIDE dataset is a high definition dataset captured in indoor scenes to keep the space position of objects consistent between hazy scenes and haze-free scenes.
It includes four types of indoor haze videos: Eastern style, Western style, Laboratory style, Corridor style.
They are extracted into 2031  hazy/clear frame-pairs. Those are then split into a training dataset including 1747 frame-pairs and a test dataset including 284 frame-pairs from 6 video clips. The resolution of each frame is $2708\times1800$ pixel. To further extend the training dataset and hold some small objects, we adopt 0.75 and 0.5 ratios to downsample each frame-pair and the number of training frame-pairs achieves to 5241.
Fig.~\ref{fig:dataset} show four examples from the training dataset.

\subsection{Experiment Settings}
\begin{table*}[!htp]
	\footnotesize
	%\scriptsize
	\begin{center}
		%\vspace{-0.3cm}
		\caption{Quantitative evaluations on the REVIDE test dataset.}
		\label{table:REVIDE}
		\begin{tabular}{cccccccccc}
			\toprule
			Methods&Input &DCP~\cite{DBLP:journals/pami/He0T11} &FFA~\cite{DBLP:conf/aaai/QinWBXJ20} & GridDehazeNet~\cite{DBLP:conf/iccv/LiuMSC19} &RIVD~\cite{DBLP:conf/eccv/ChenDW16} &STMRF~\cite{DBLP:conf/pcm/CaiXT16} &EVDNet~\cite{DBLP:conf/aaai/LiPWXF18} &FastDVDnet~\cite{DBLP:conf/cvpr/TassanoDV20} &Ours \\  \midrule
			PSNR& 15.05 & 12.24 &18.57  &18.36 &14.36  &15.54  &17.41 &16.37 &22.69 \\
			SSIM& 0.770& 0.600&0.800 &0.830 &0.701 &0.692&0.808&0.656 &0.875\\
			\bottomrule
		\end{tabular}
		%\vspace{-0.5cm}
	\end{center}
\end{table*}
%%%%%%%%%%%%%%%%%%
\begin{figure*}[!htp]
	\footnotesize
	\begin{center}
		\begin{tabular}{cccccccc}
			%\vspace{-0.2cm}
			\includegraphics[width = 0.12\linewidth, height = 0.1\linewidth]{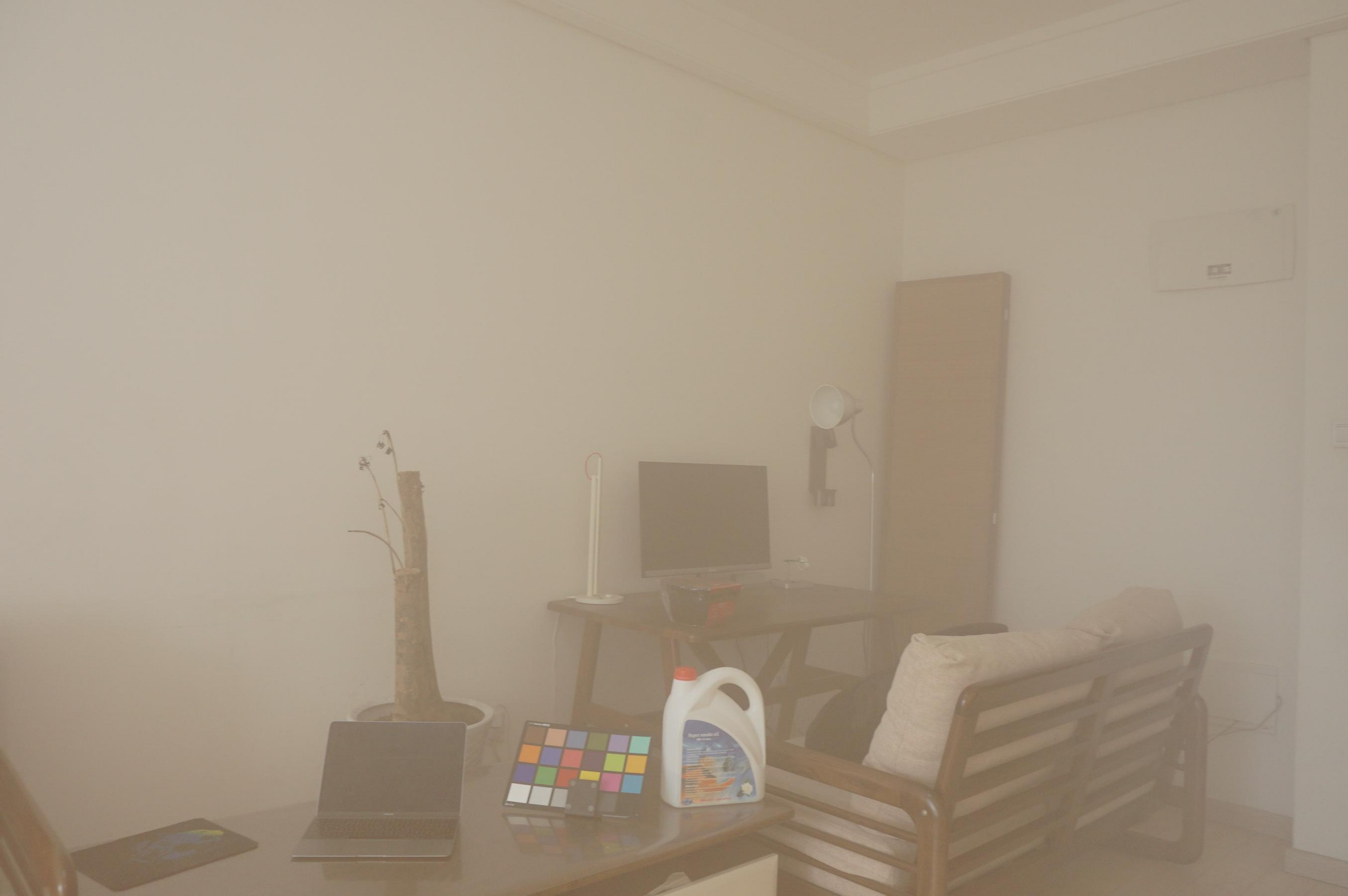}& \hspace{-0.46cm}
			\includegraphics[width = 0.12\linewidth, height = 0.1\linewidth]{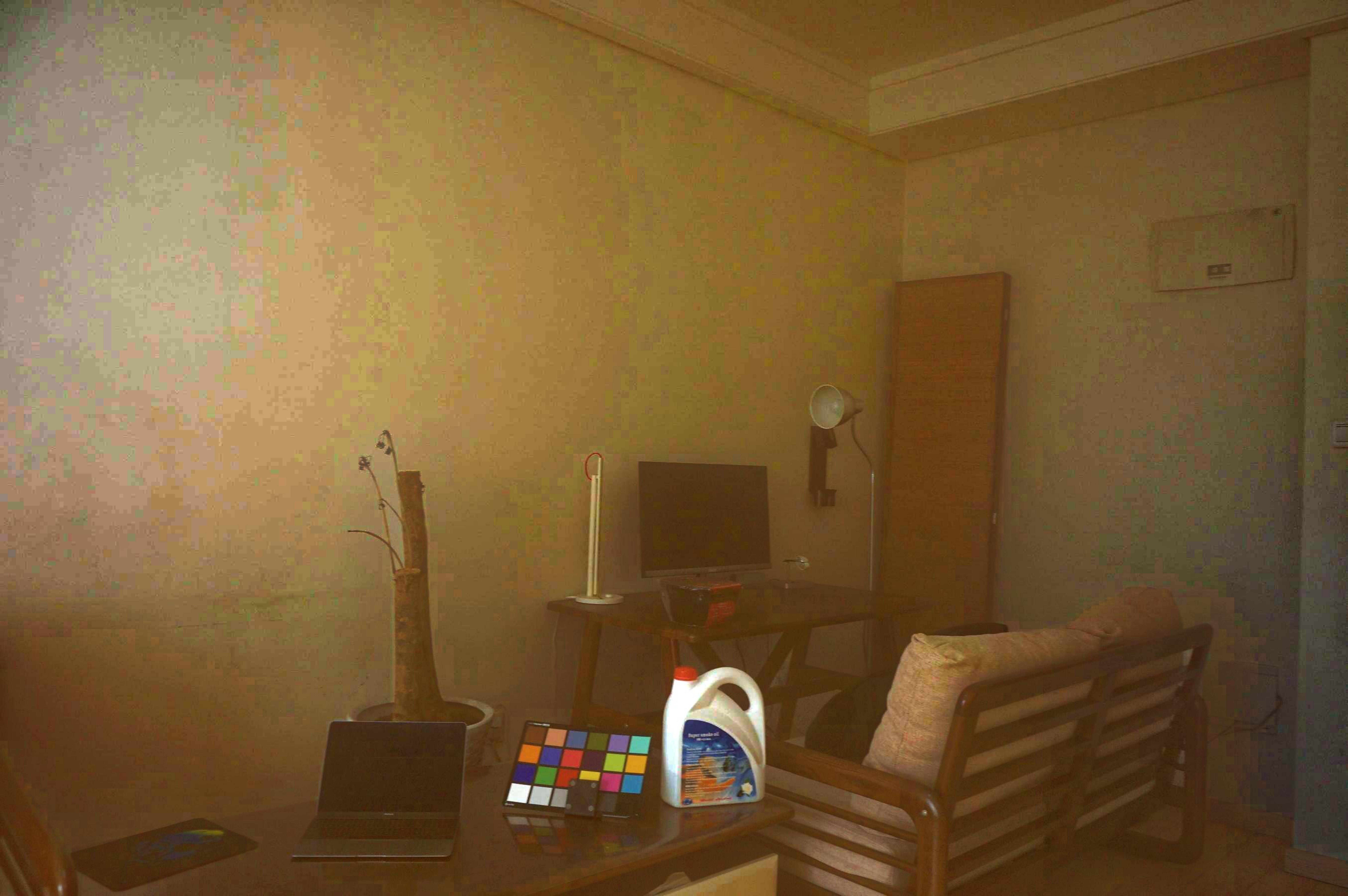} & \hspace{-0.46cm}
			\includegraphics[width = 0.12\linewidth, height = 0.1\linewidth]{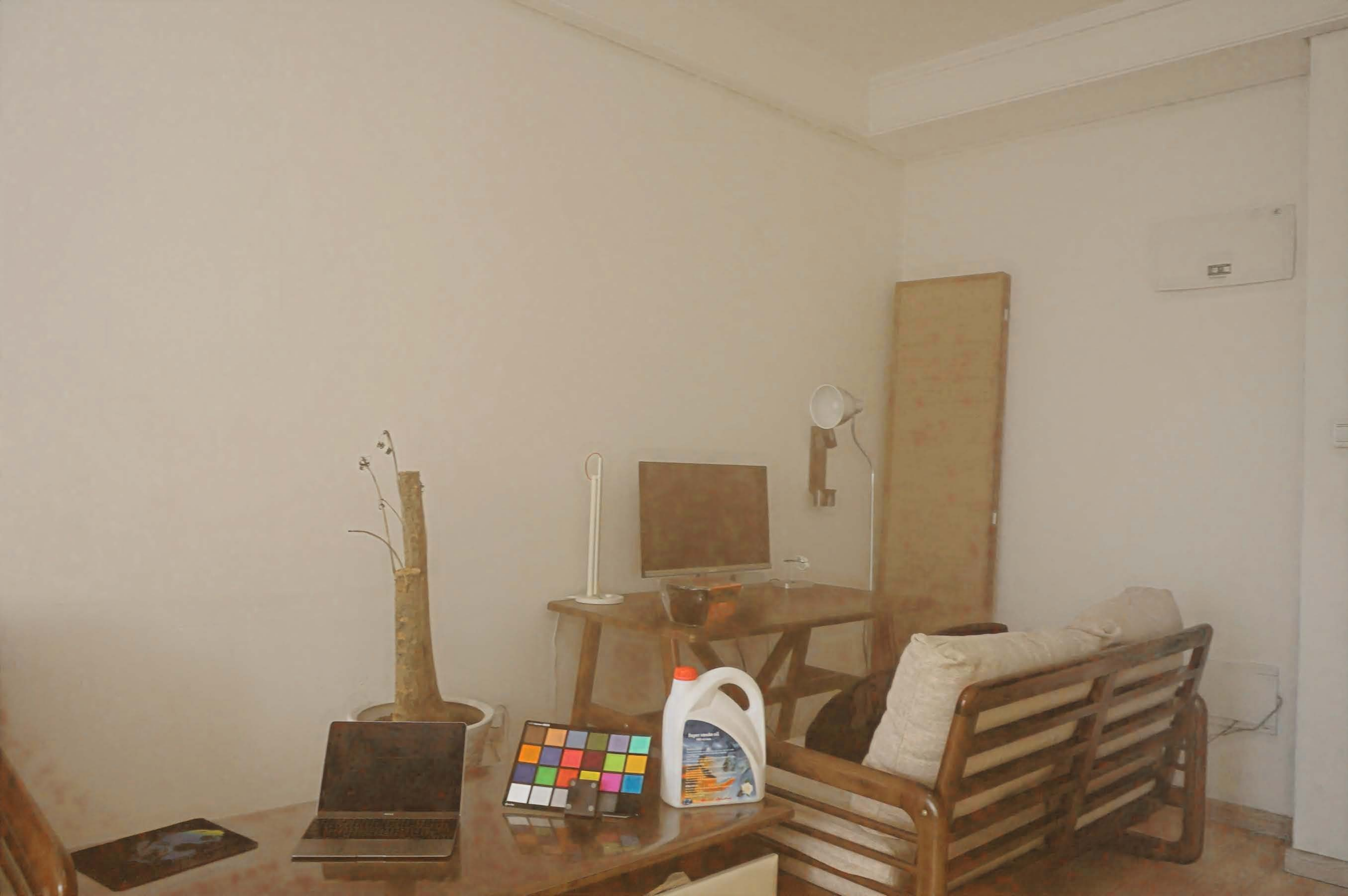} & \hspace{-0.46cm}
			\includegraphics[width = 0.12\linewidth, height = 0.1\linewidth]{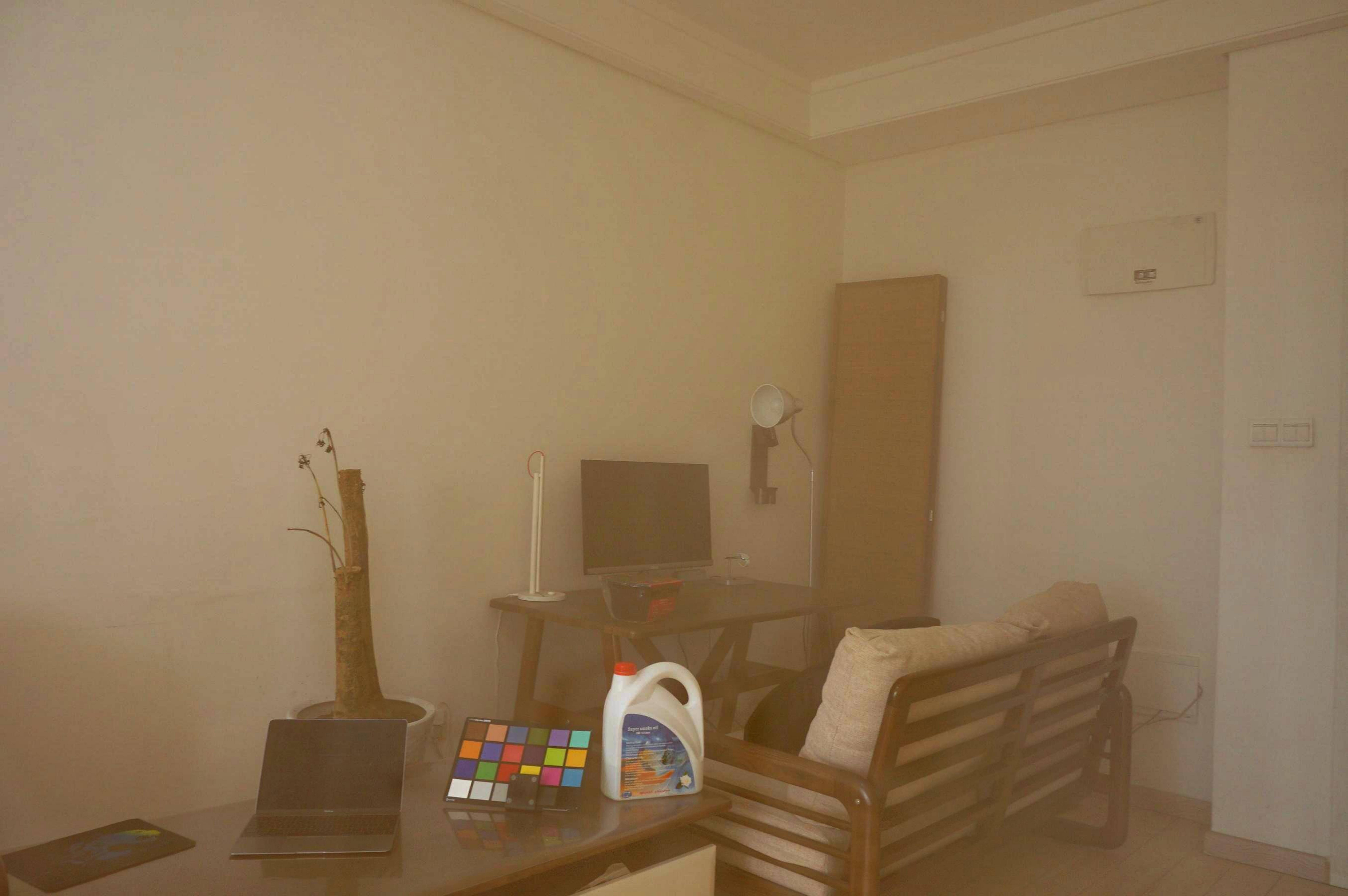} & \hspace{-0.46cm}
			\includegraphics[width = 0.12\linewidth, height = 0.1\linewidth]{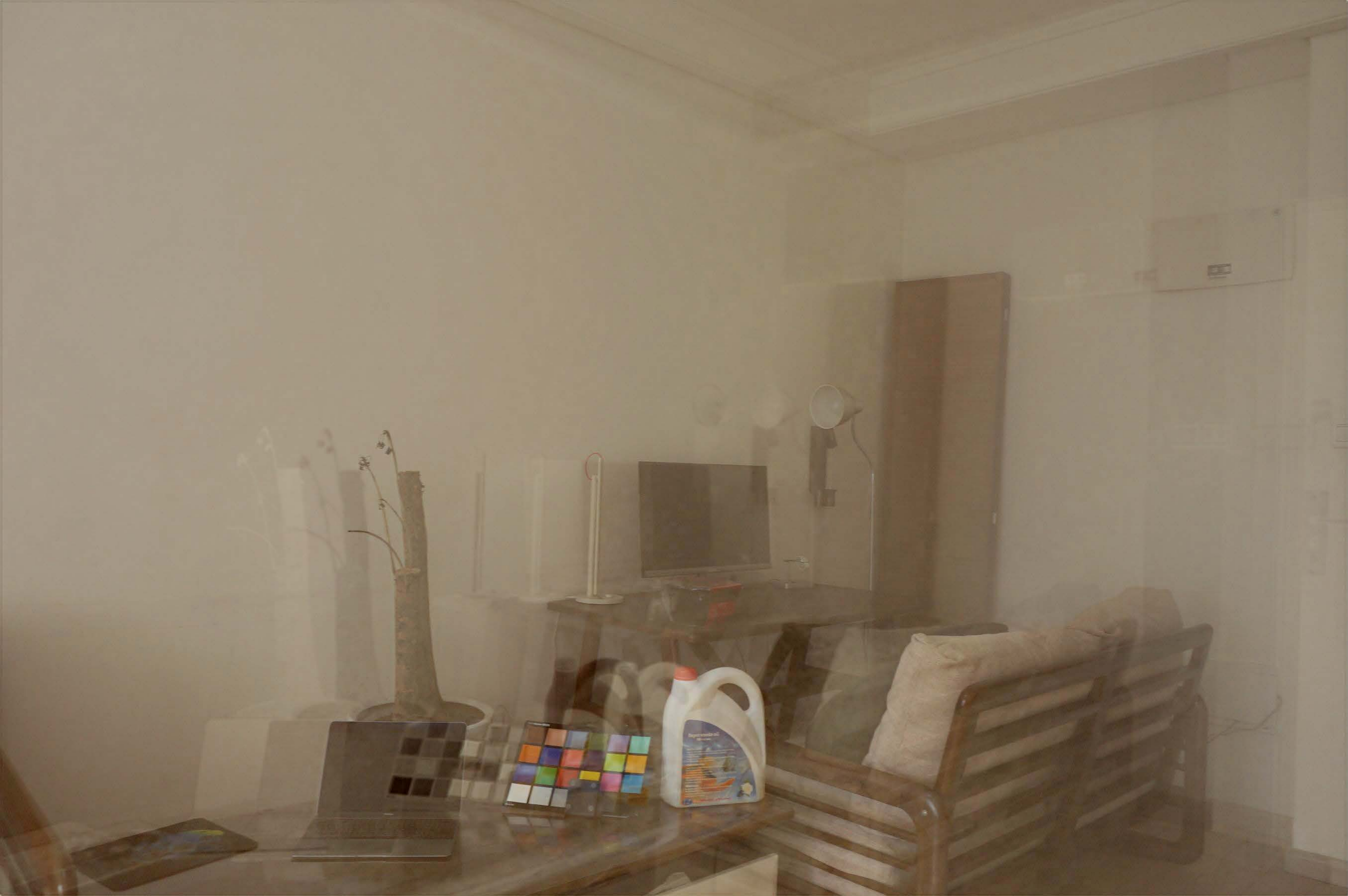} & \hspace{-0.46cm}
			\includegraphics[width = 0.12\linewidth, height = 0.1\linewidth]{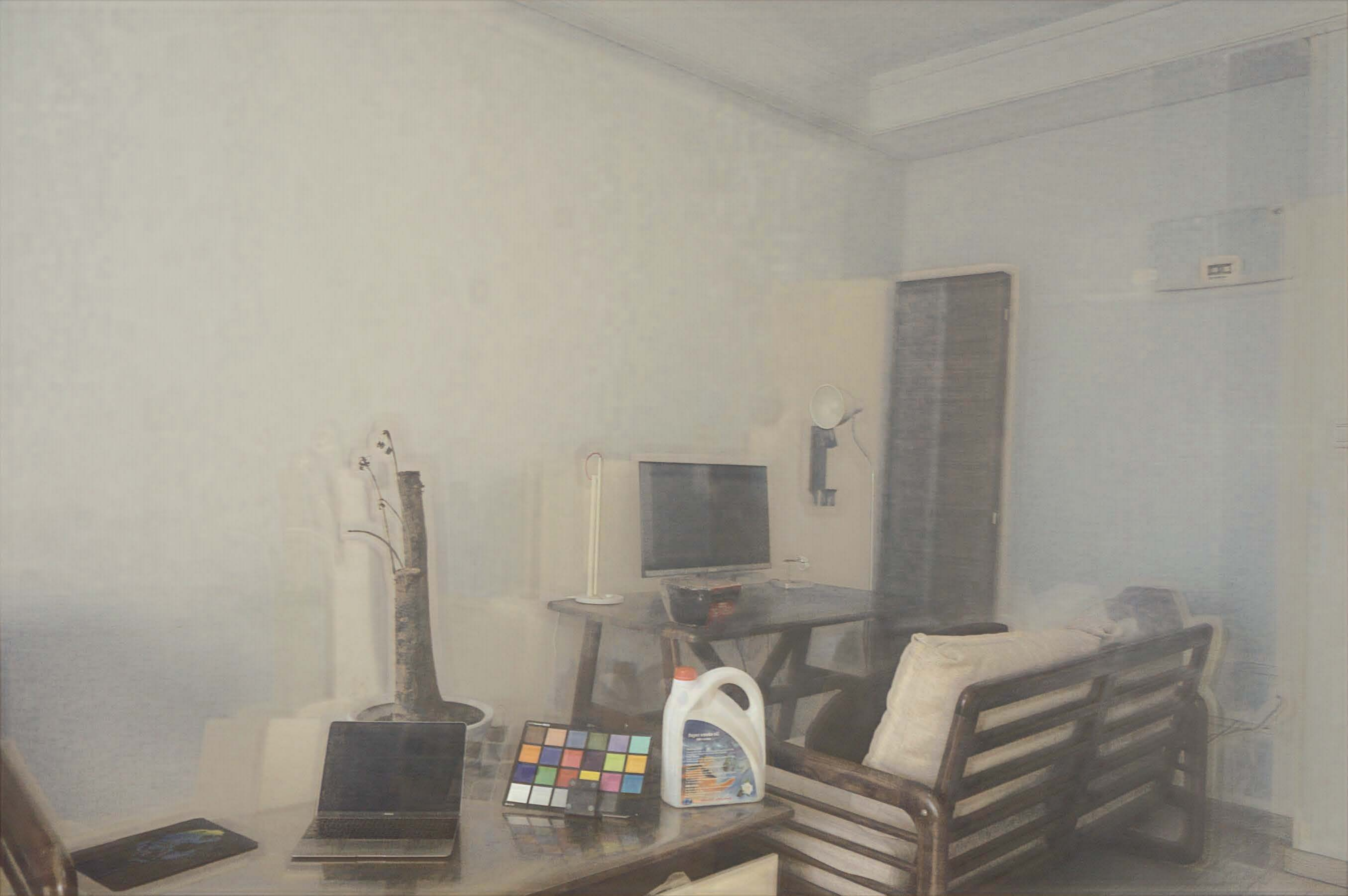} & \hspace{-0.46cm}
			\includegraphics[width = 0.12\linewidth, height = 0.1\linewidth]{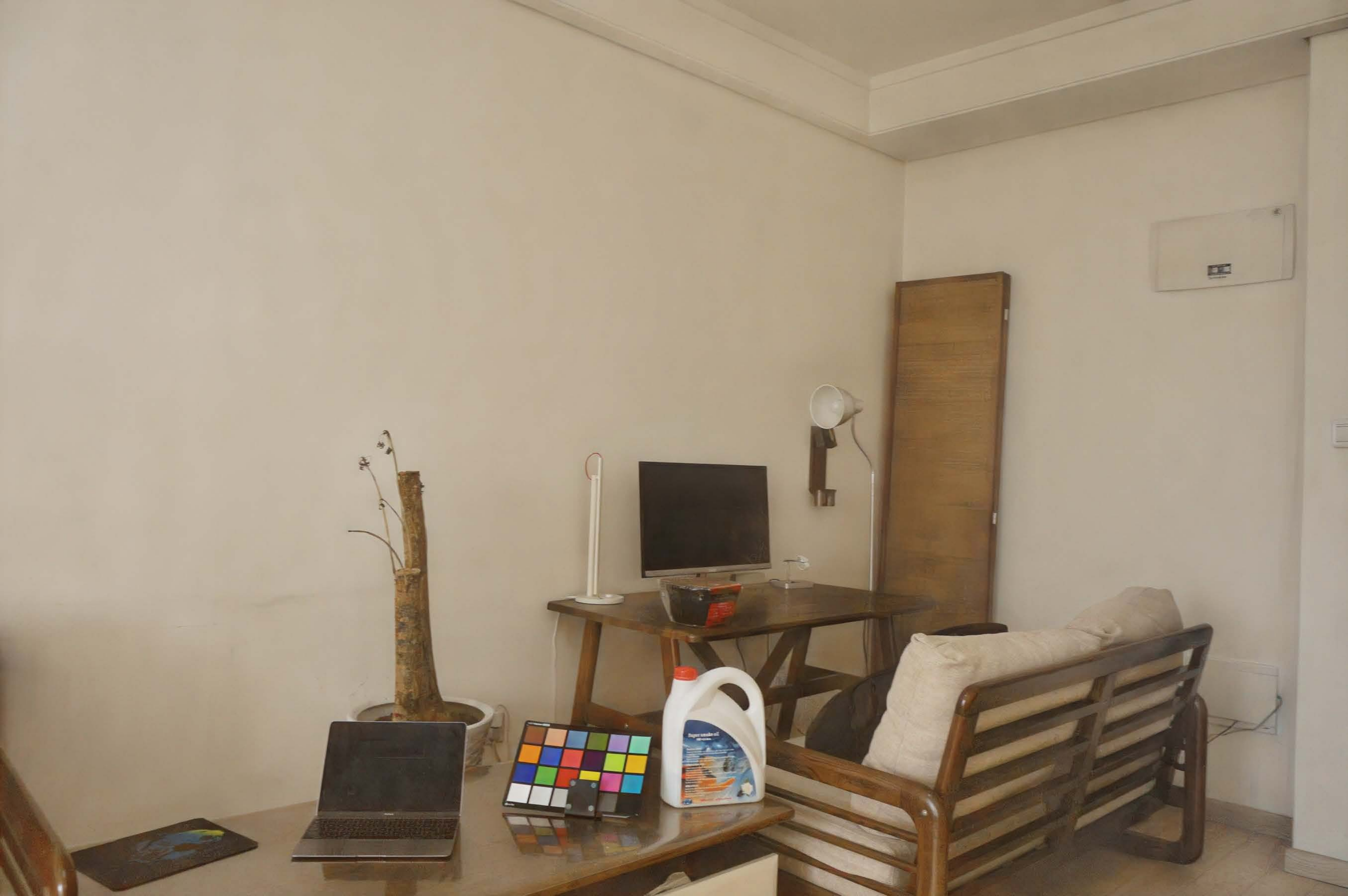} & \hspace{-0.46cm}
			\includegraphics[width = 0.12\linewidth, height = 0.1\linewidth]{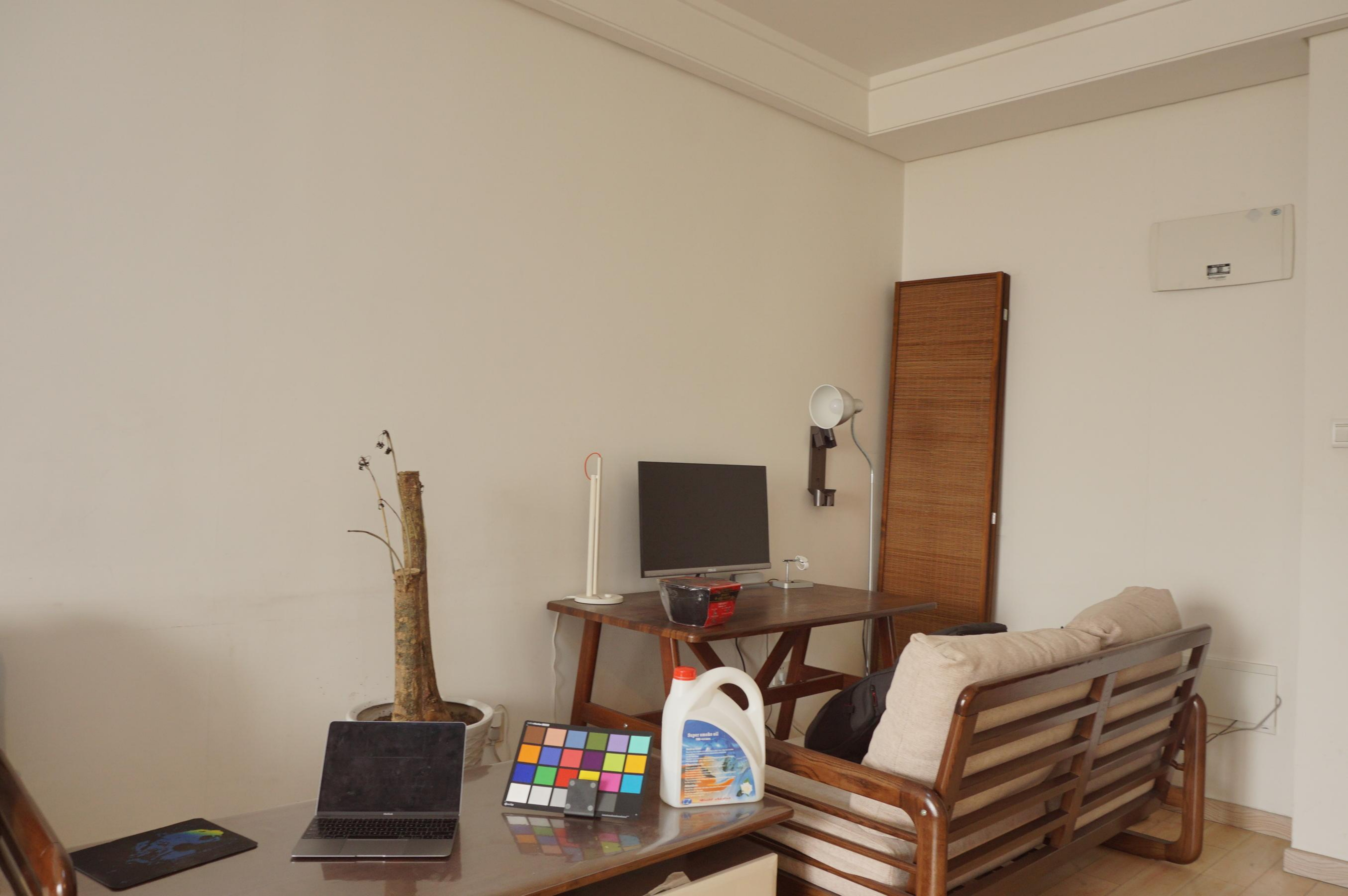}\\
			18.99/0.911  & \hspace{-0.46cm} 11.57/0.663 & \hspace{-0.46cm} 23.32/0.940 & \hspace{-0.46cm} 14.90/0.786& \hspace{-0.46cm} 17.73/0.911& \hspace{-0.46cm}  20.67/0.716 & \hspace{-0.46cm}  24.39/0.956& \hspace{-0.46cm}  $+\infty$/1 \\			
            \includegraphics[width = 0.12\linewidth, height = 0.1\linewidth]{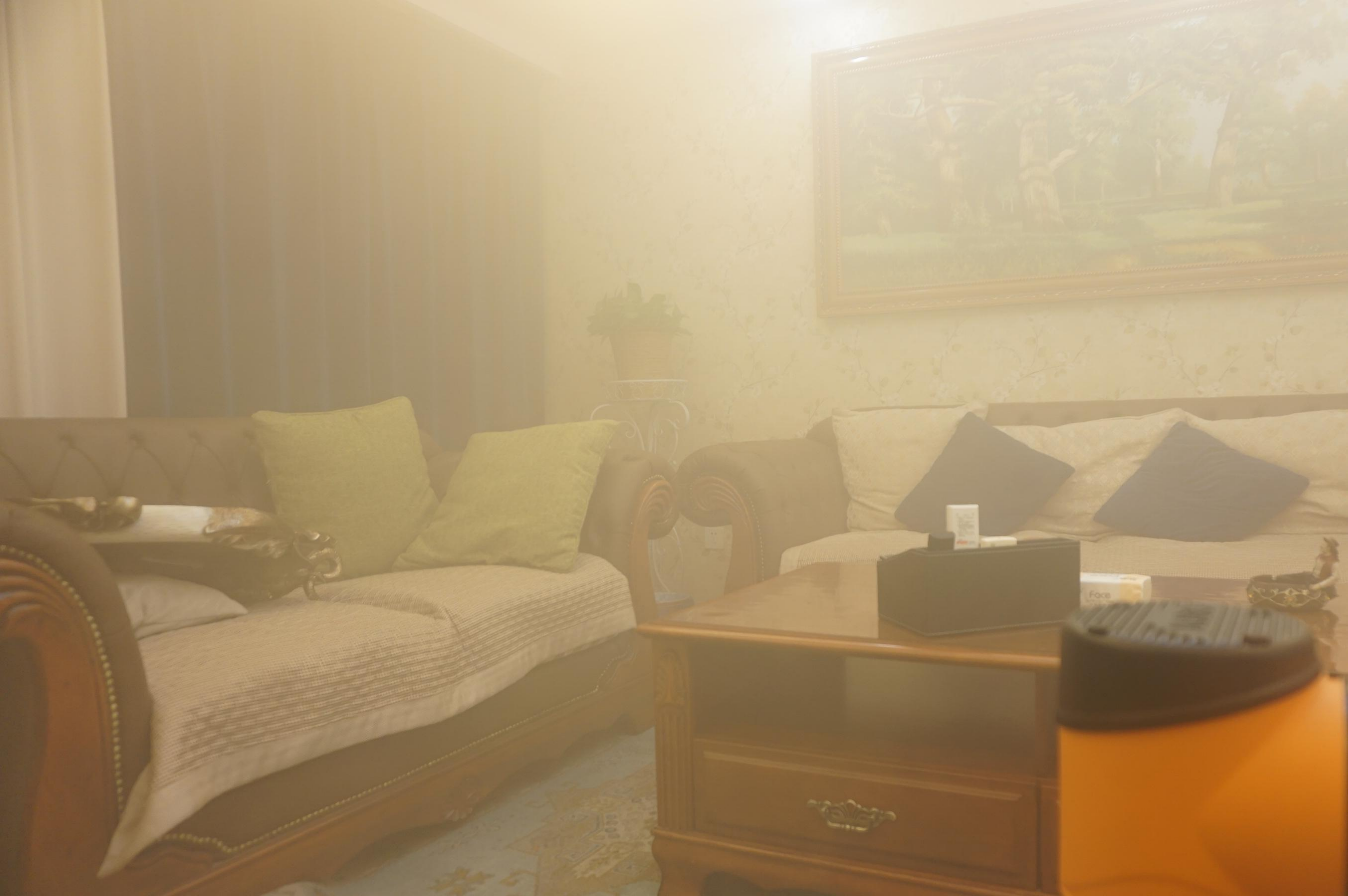}& \hspace{-0.46cm}
            \includegraphics[width = 0.12\linewidth, height = 0.1\linewidth]{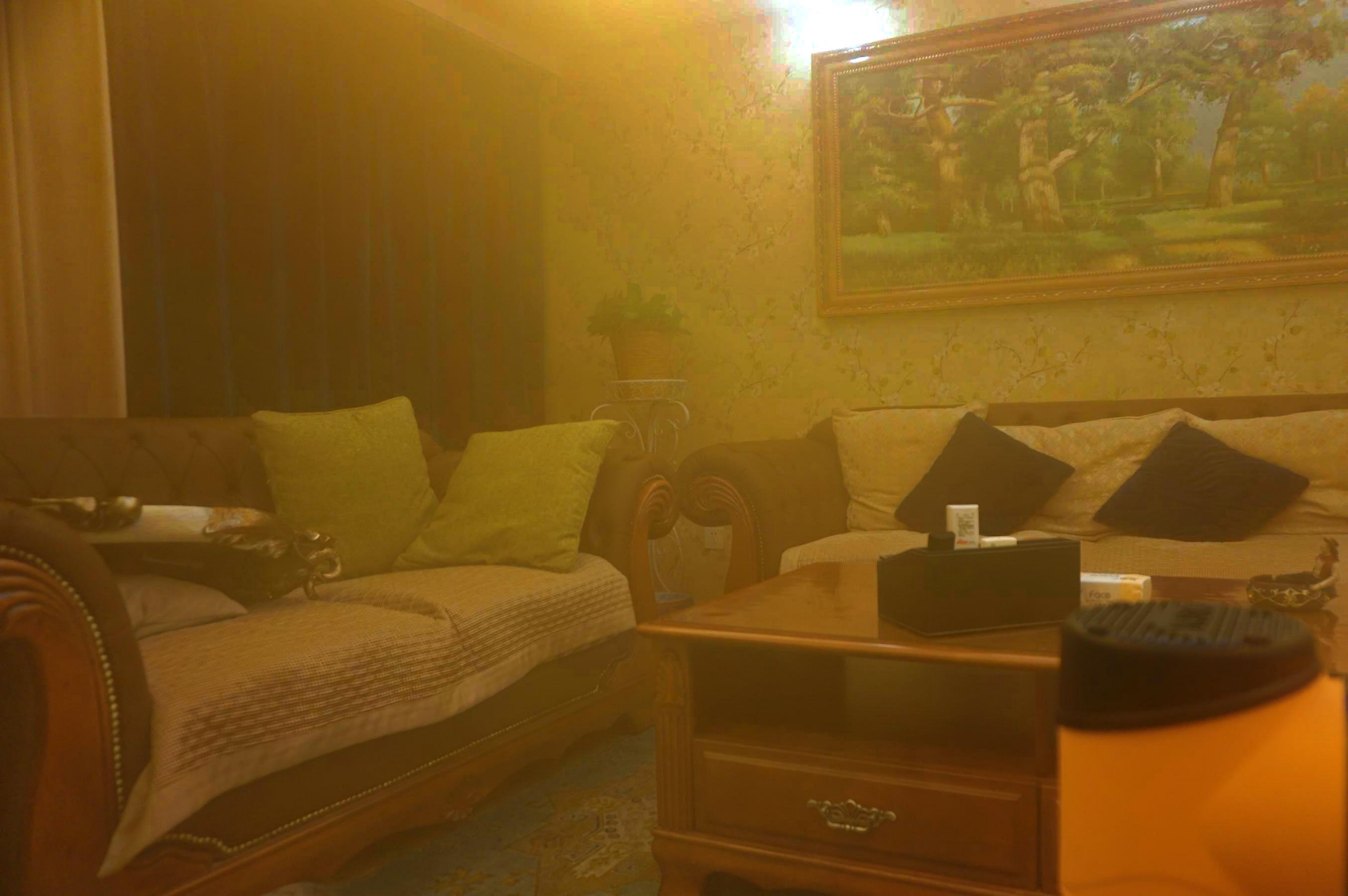} & \hspace{-0.46cm}
            \includegraphics[width = 0.12\linewidth, height = 0.1\linewidth]{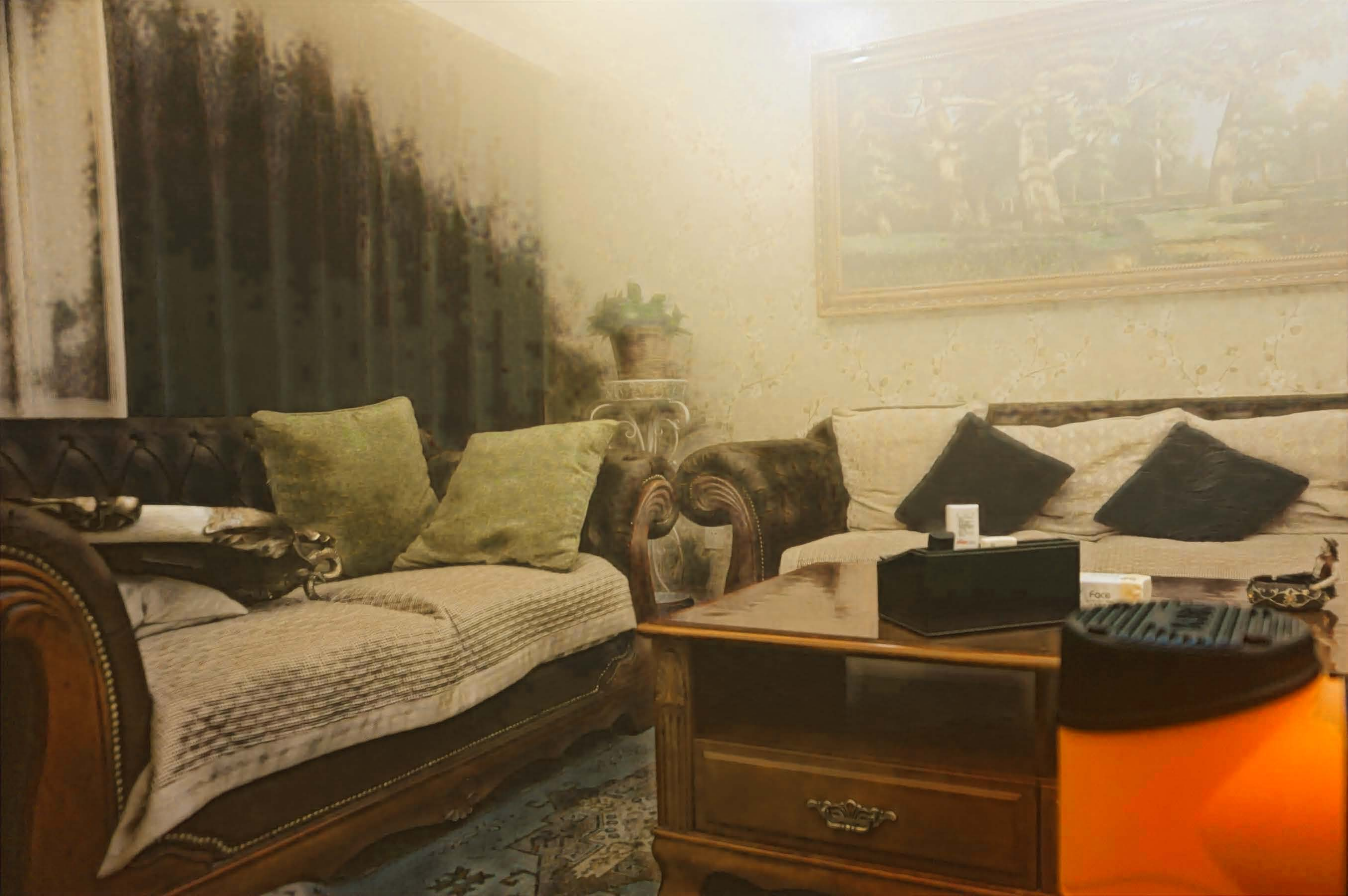} & \hspace{-0.46cm}
            \includegraphics[width = 0.12\linewidth, height = 0.1\linewidth]{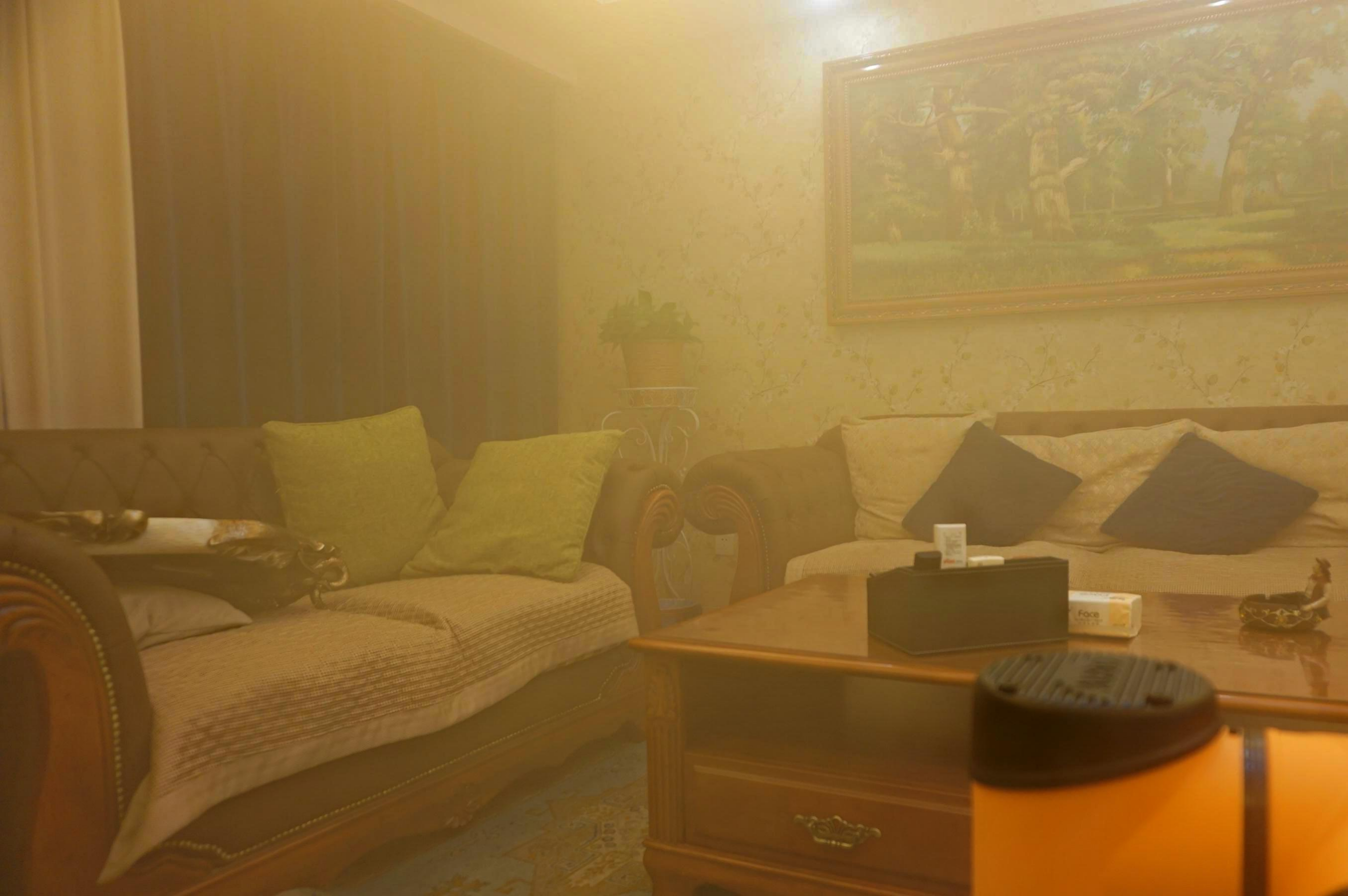} & \hspace{-0.46cm}
            \includegraphics[width = 0.12\linewidth, height = 0.1\linewidth]{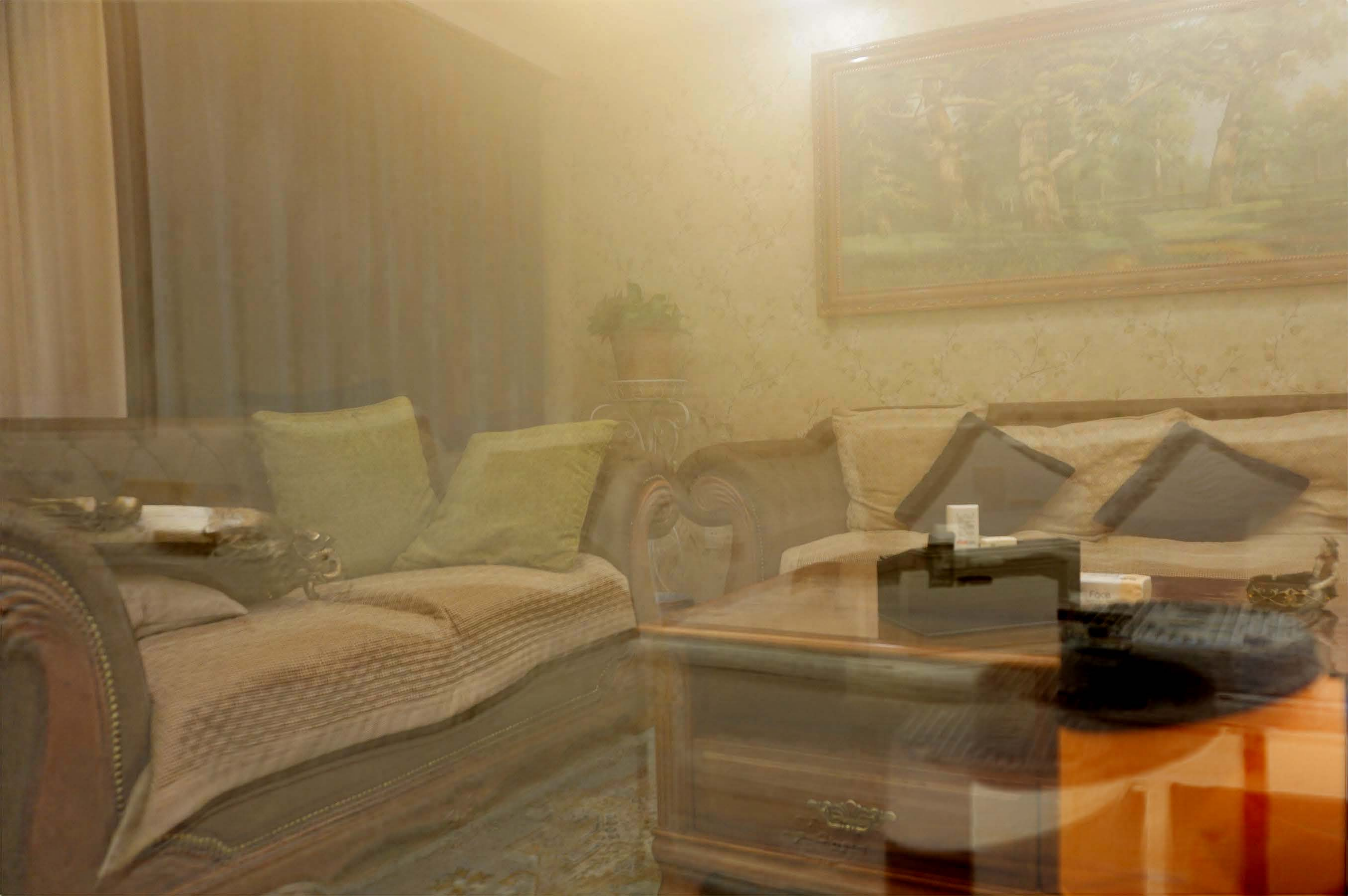} & \hspace{-0.46cm}
            \includegraphics[width = 0.12\linewidth, height = 0.1\linewidth]{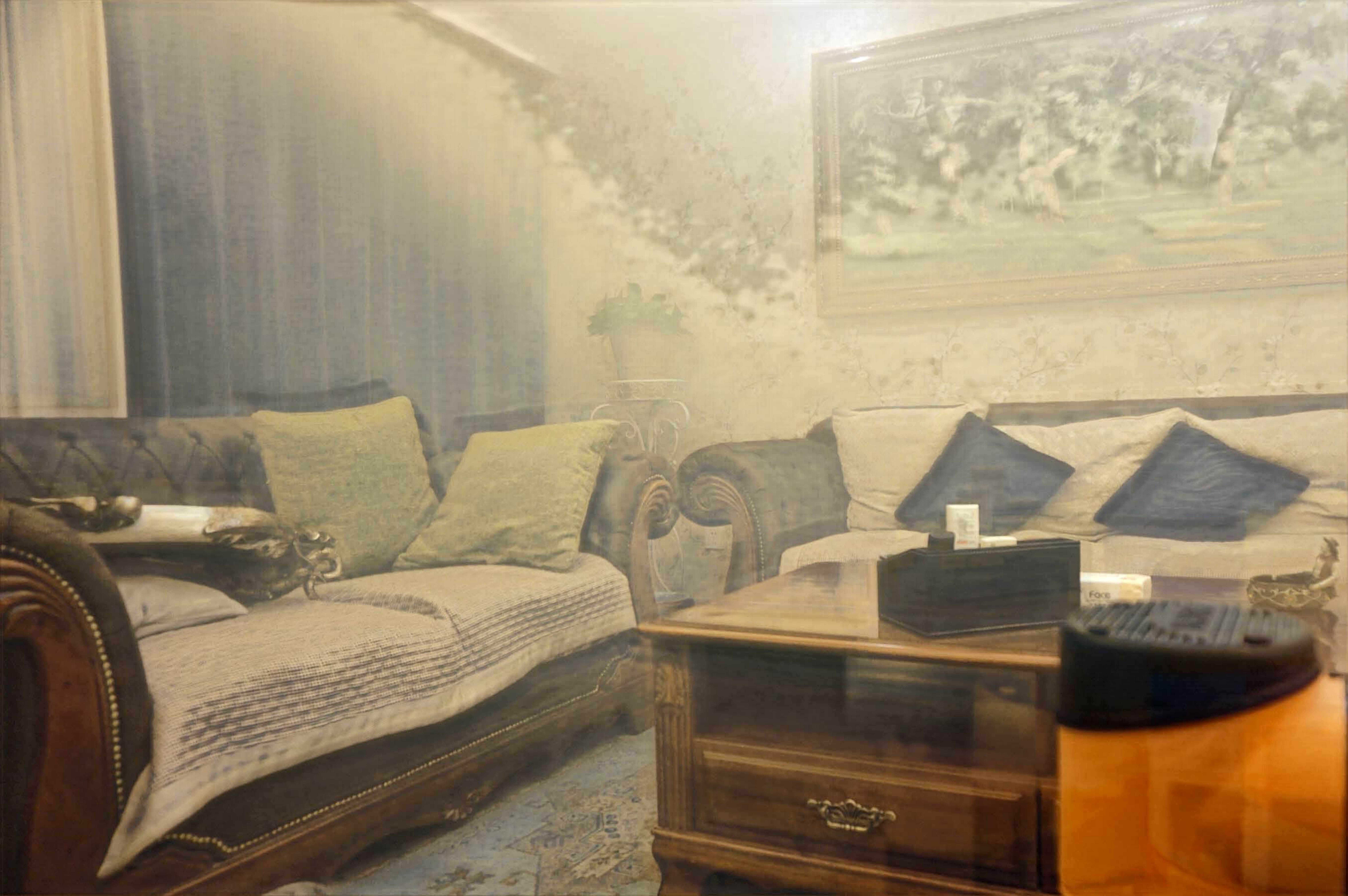} & \hspace{-0.46cm}
            \includegraphics[width = 0.12\linewidth, height = 0.1\linewidth]{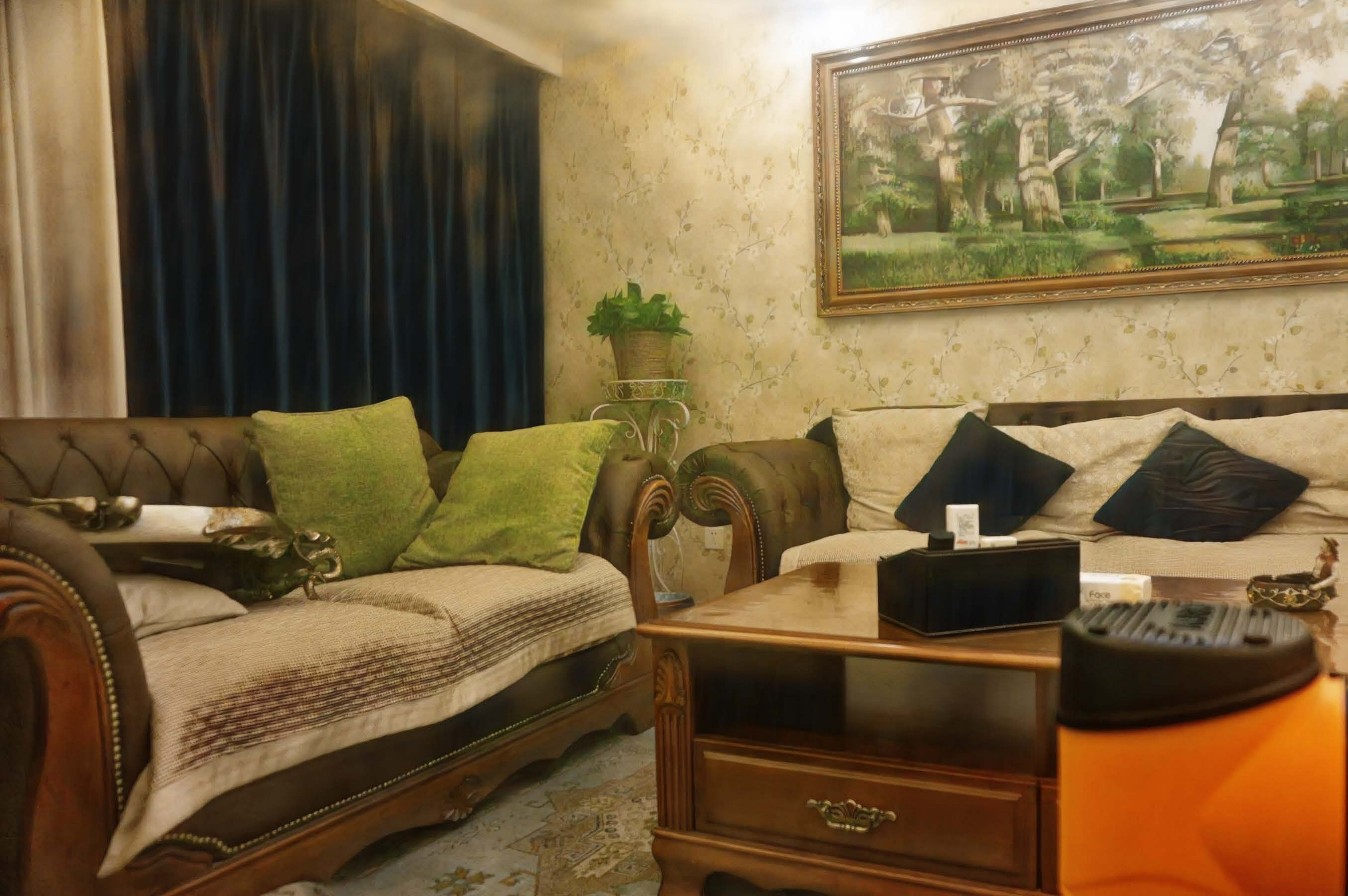} & \hspace{-0.46cm}
            \includegraphics[width = 0.12\linewidth, height = 0.1\linewidth]{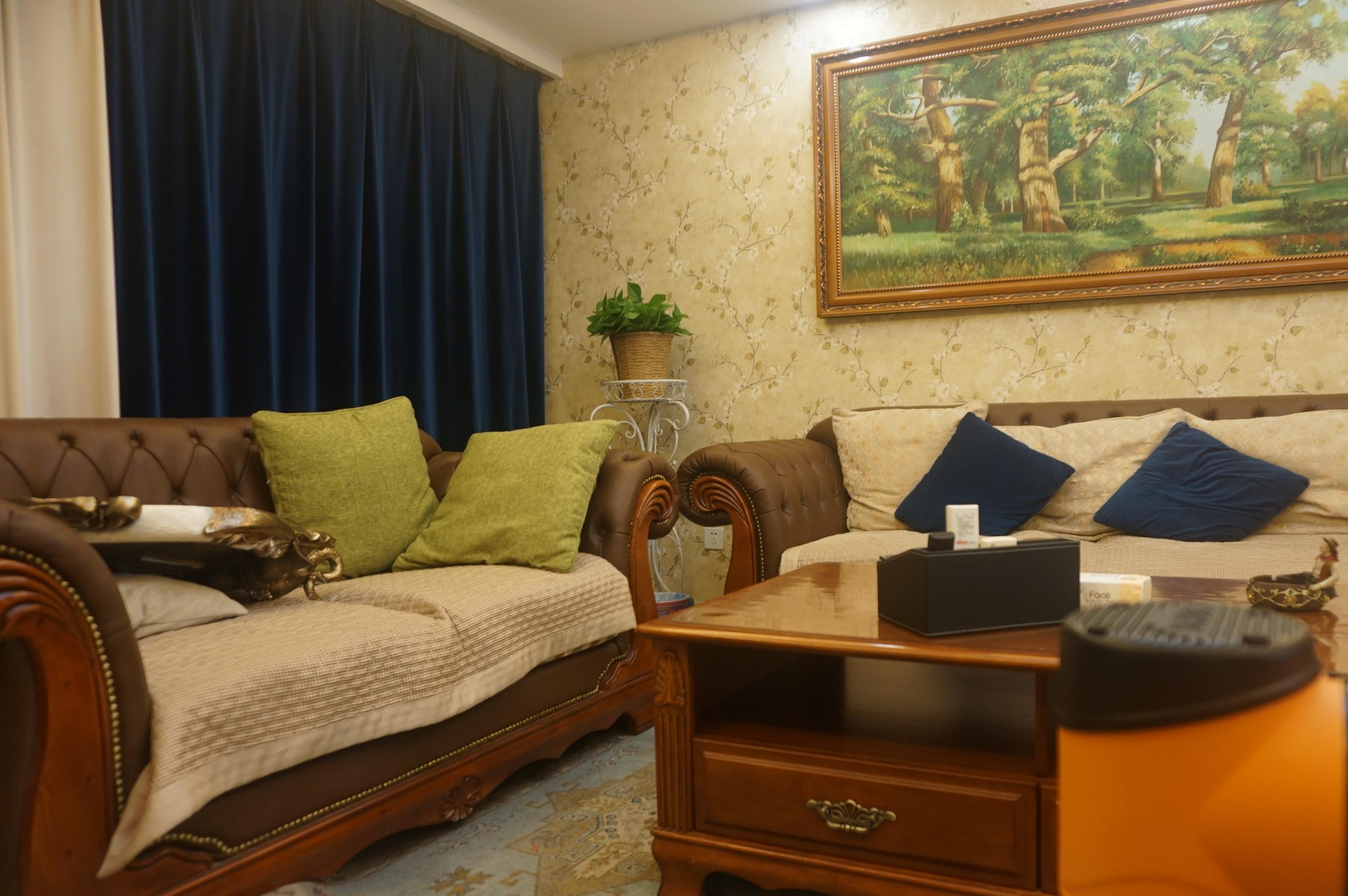}\\
            11.42/0.648  & \hspace{-0.46cm} 14.91/0.657 & \hspace{-0.46cm} 15.46/0.719 & \hspace{-0.46cm} 14.42/0.664& \hspace{-0.46cm} 14.54/0.688& \hspace{-0.46cm}  12.48/0.678 & \hspace{-0.46cm}  21.35/0.885& \hspace{-0.46cm}  $+\infty$/1 \\
            \includegraphics[width = 0.12\linewidth, height = 0.1\linewidth]{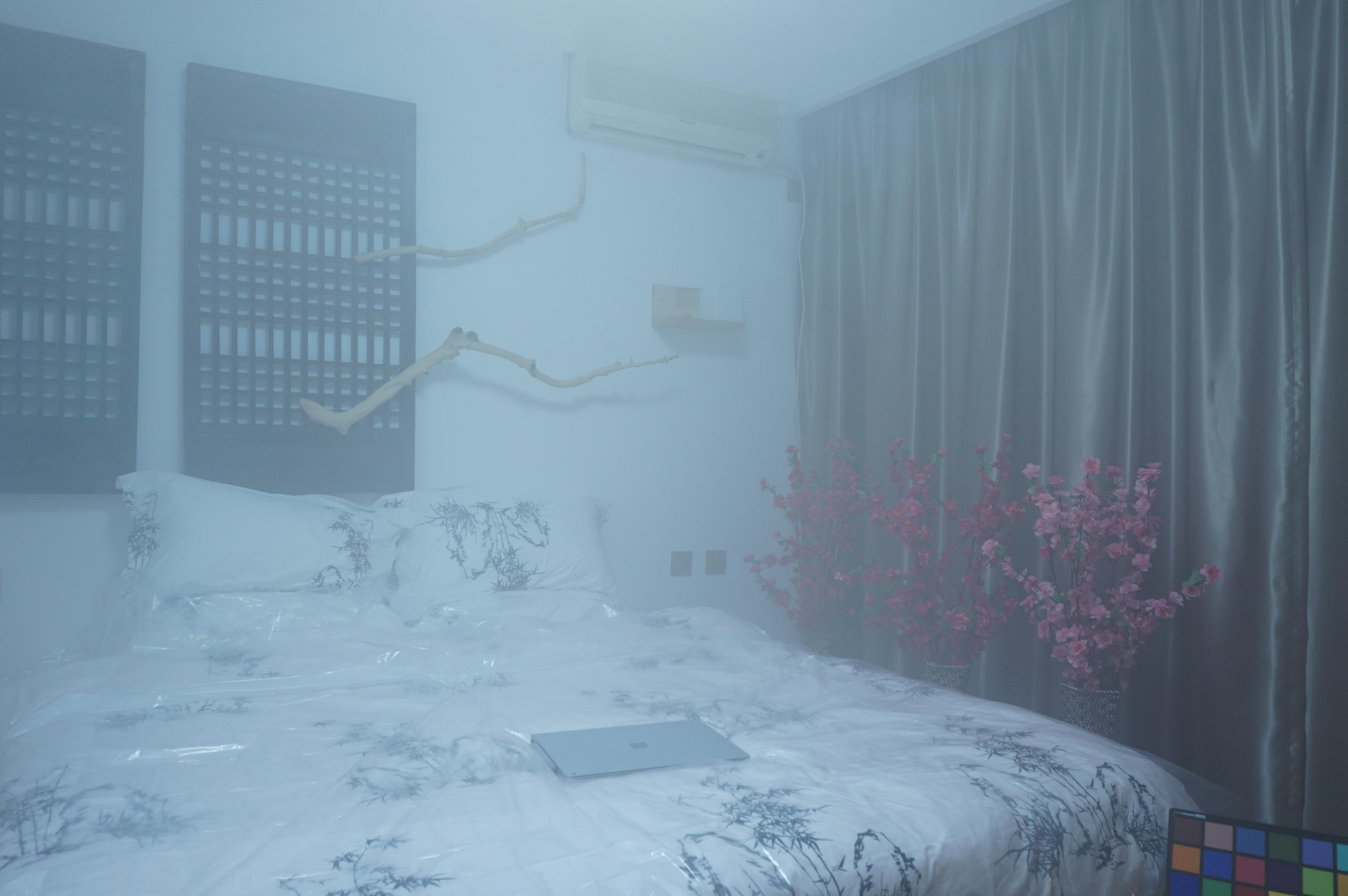}& \hspace{-0.46cm}
            \includegraphics[width = 0.12\linewidth, height = 0.1\linewidth]{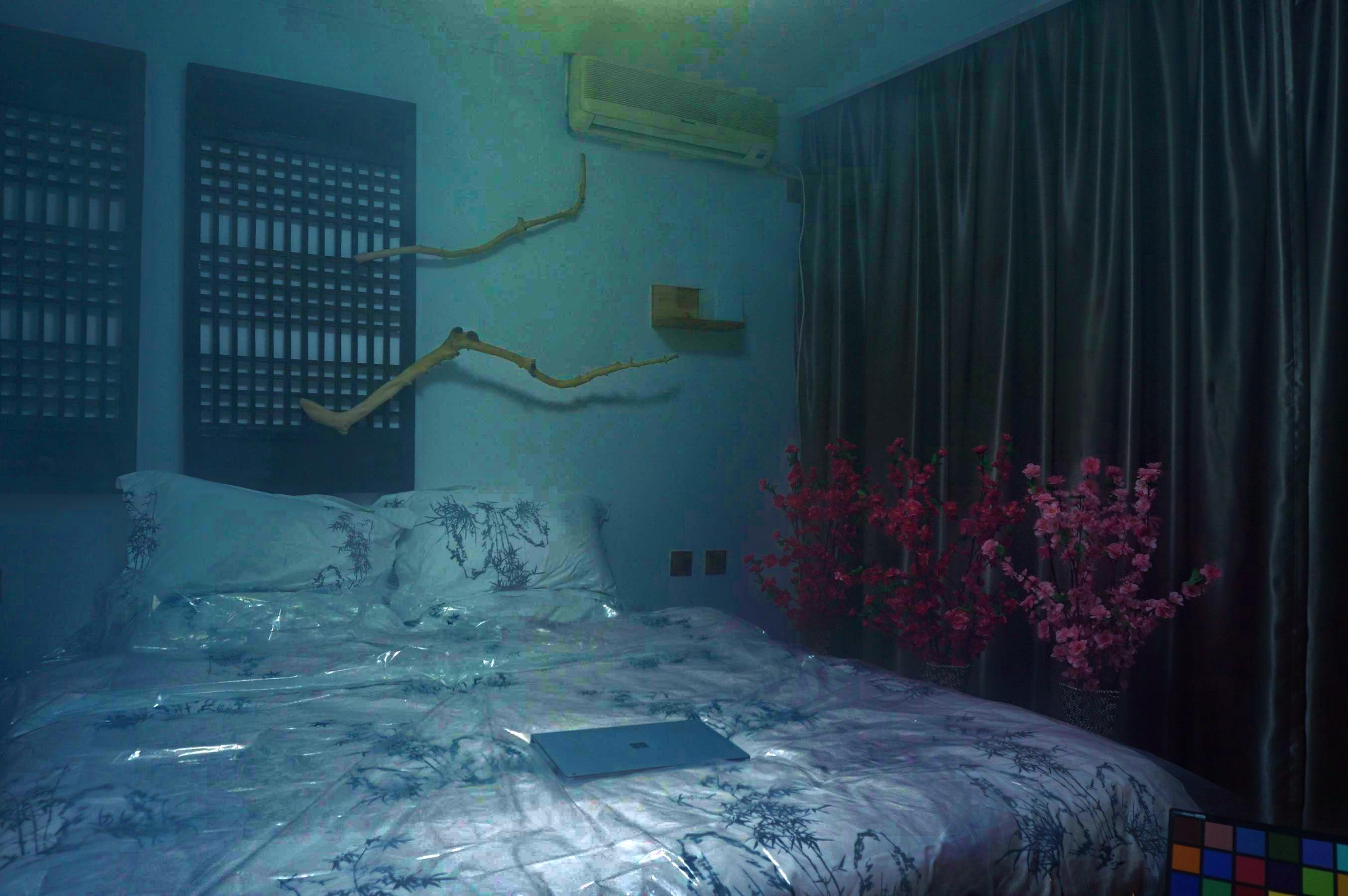} & \hspace{-0.46cm}
            \includegraphics[width = 0.12\linewidth, height = 0.1\linewidth]{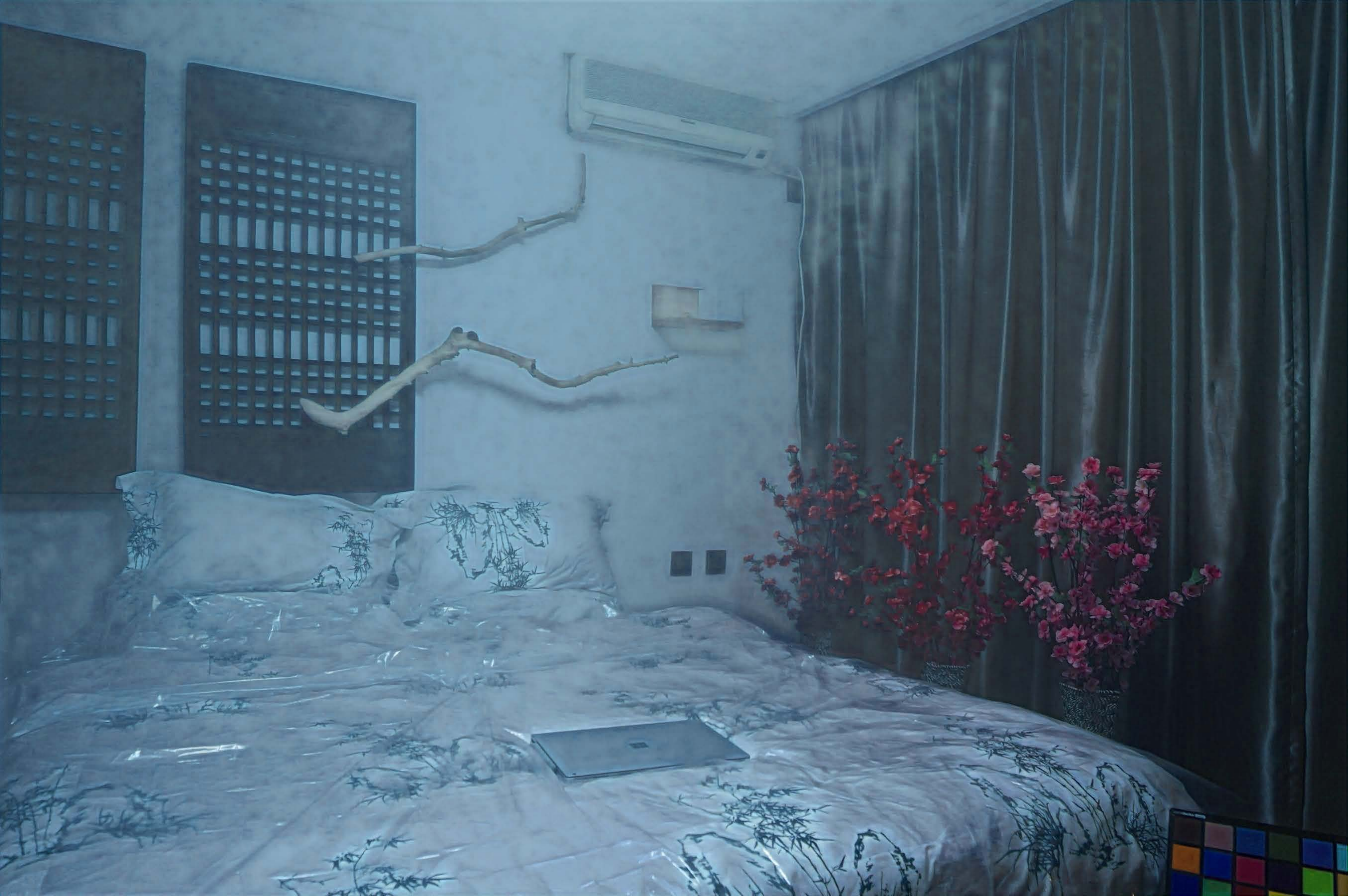} & \hspace{-0.46cm}
            \includegraphics[width = 0.12\linewidth, height = 0.1\linewidth]{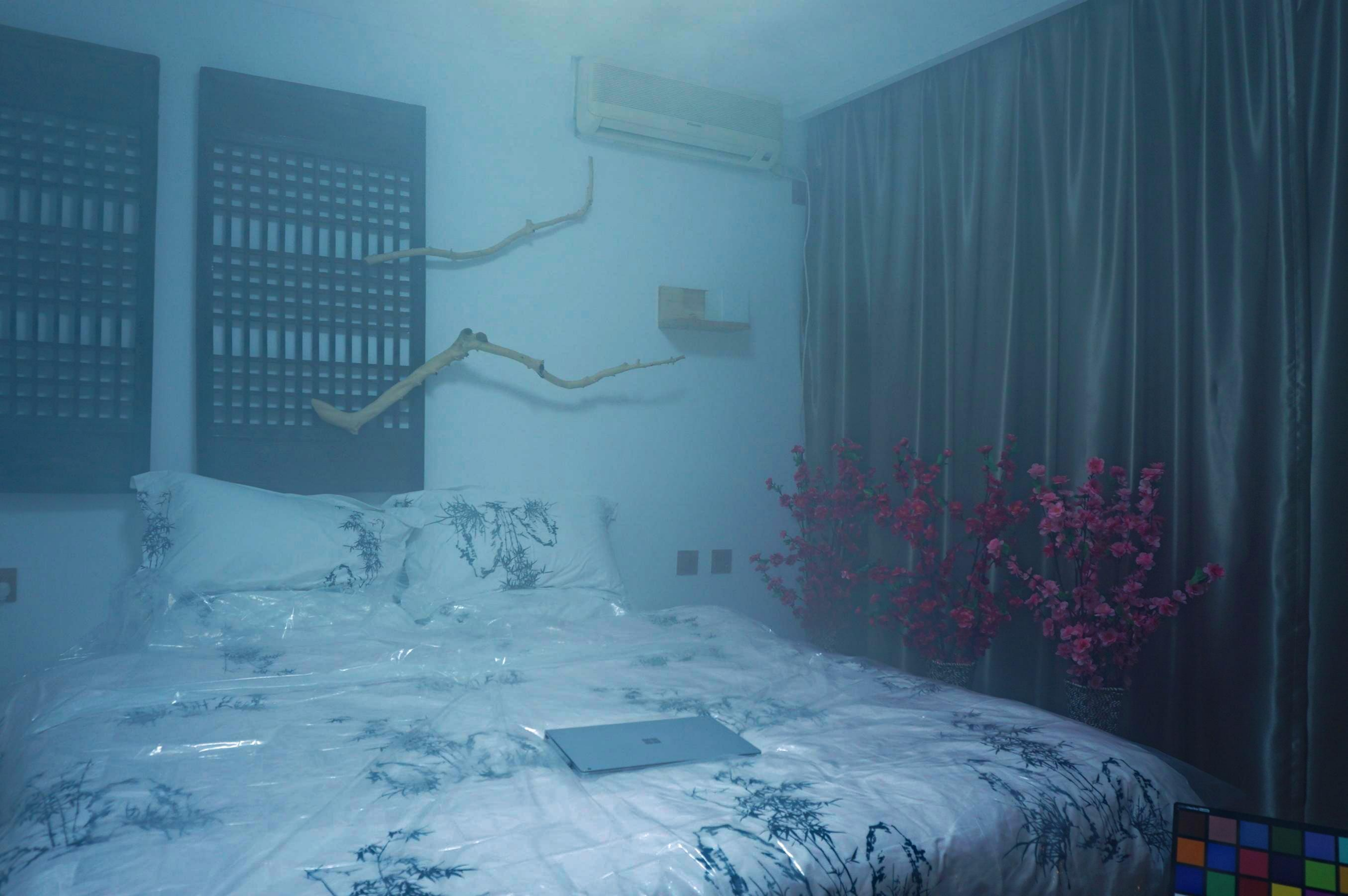} & \hspace{-0.46cm}
            \includegraphics[width = 0.12\linewidth, height = 0.1\linewidth]{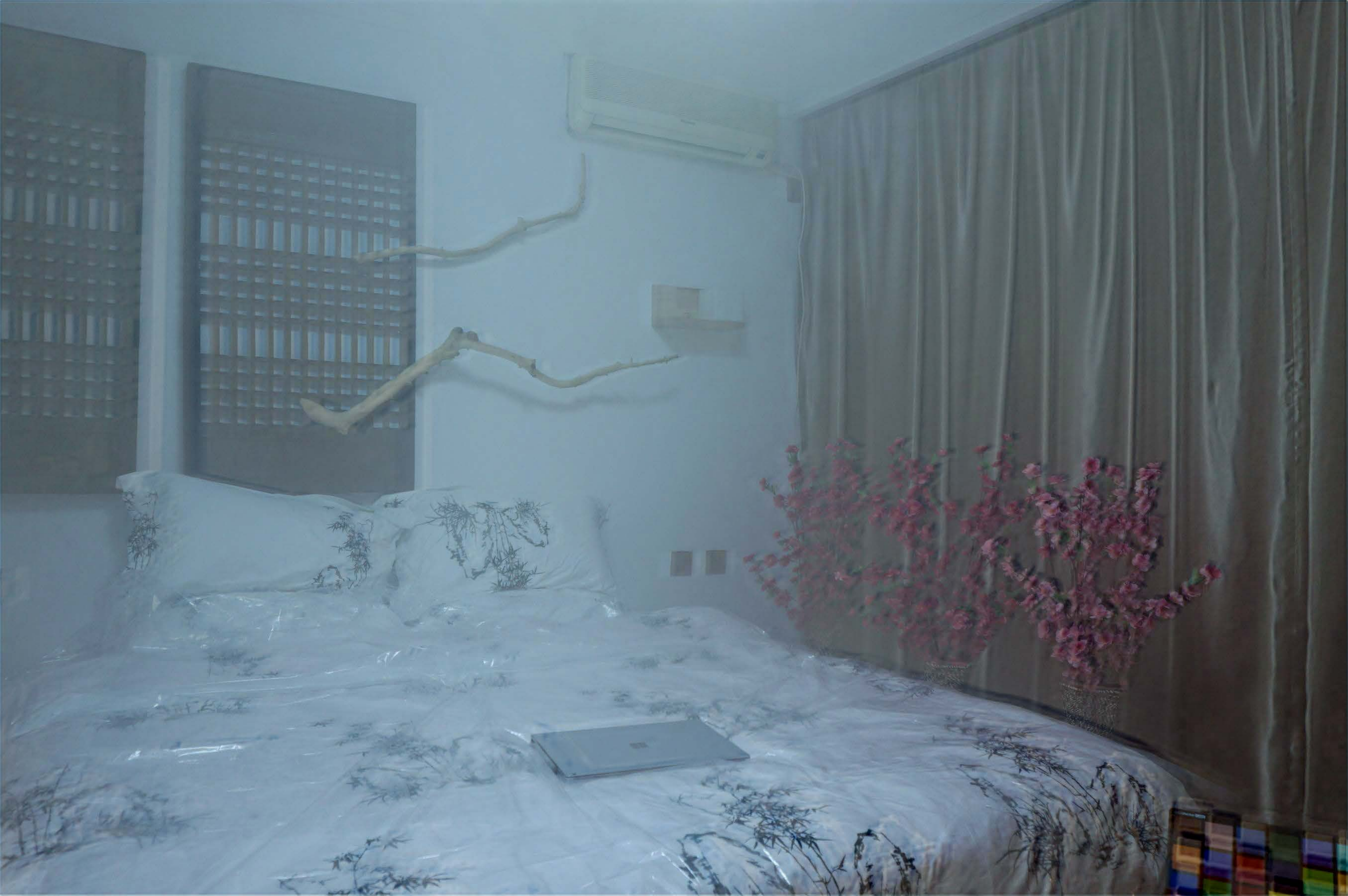} & \hspace{-0.46cm}
            \includegraphics[width = 0.12\linewidth, height = 0.1\linewidth]{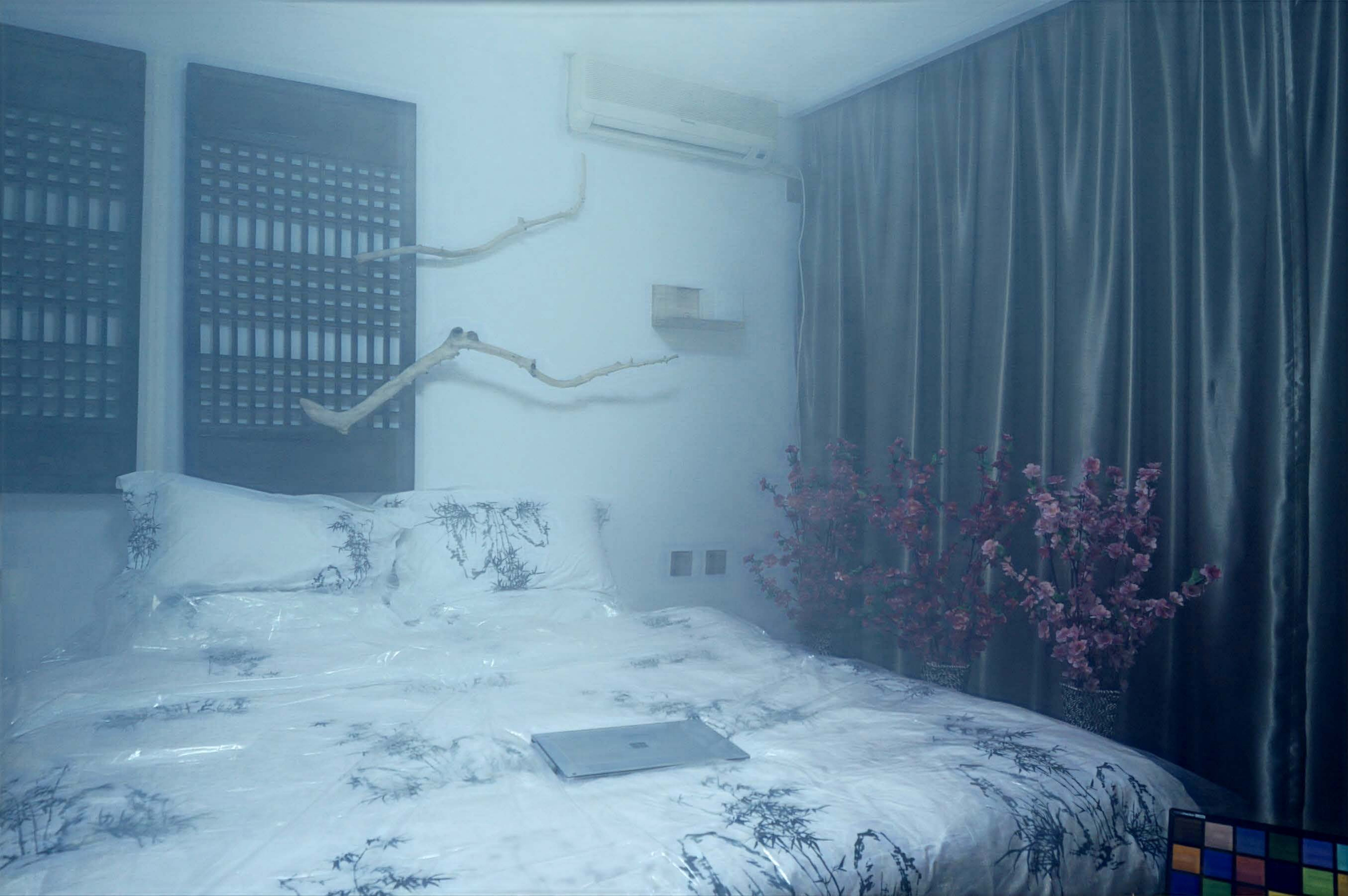} & \hspace{-0.46cm}
            \includegraphics[width = 0.12\linewidth, height = 0.1\linewidth]{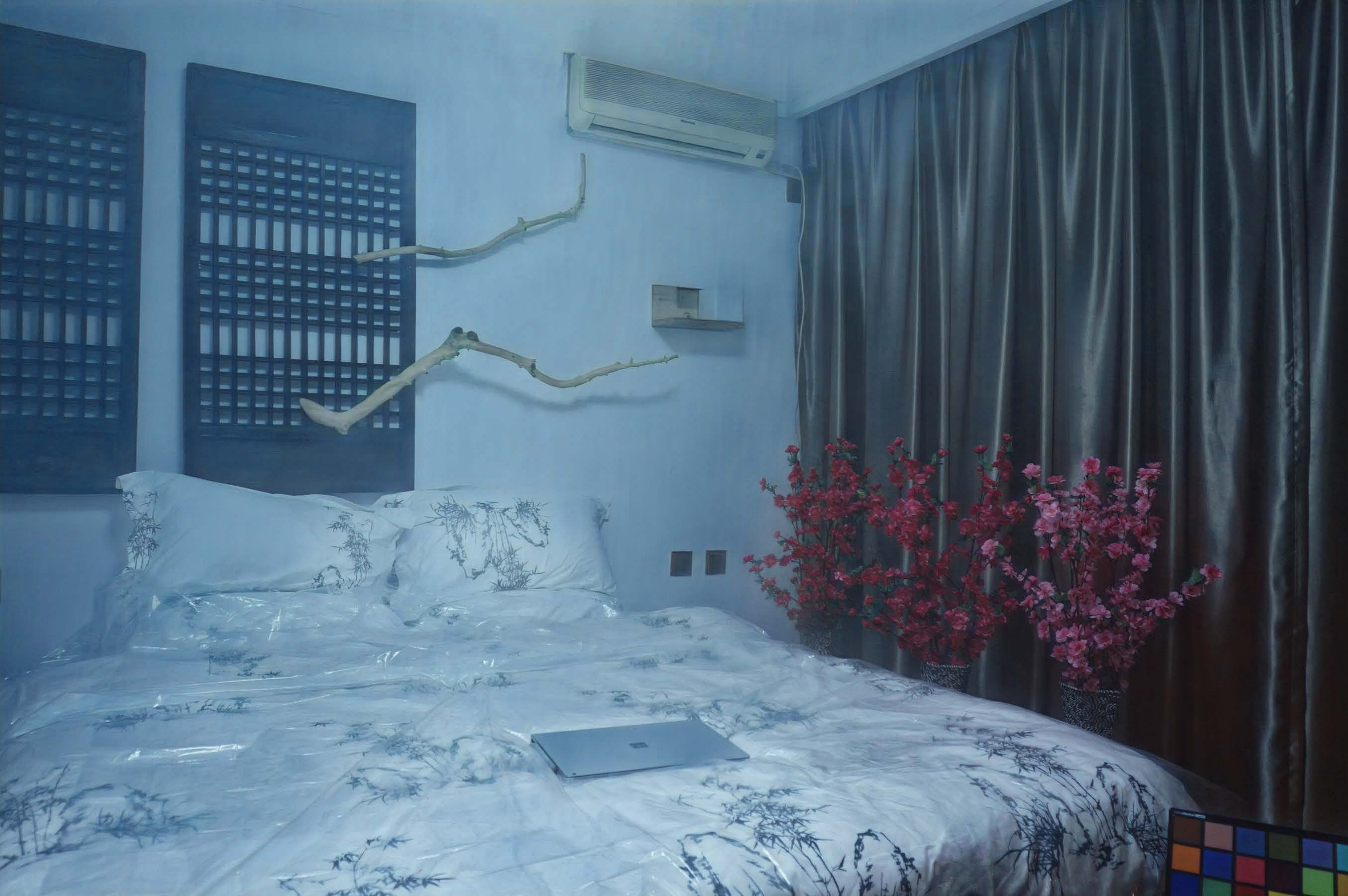} & \hspace{-0.46cm}
            \includegraphics[width = 0.12\linewidth, height = 0.1\linewidth]{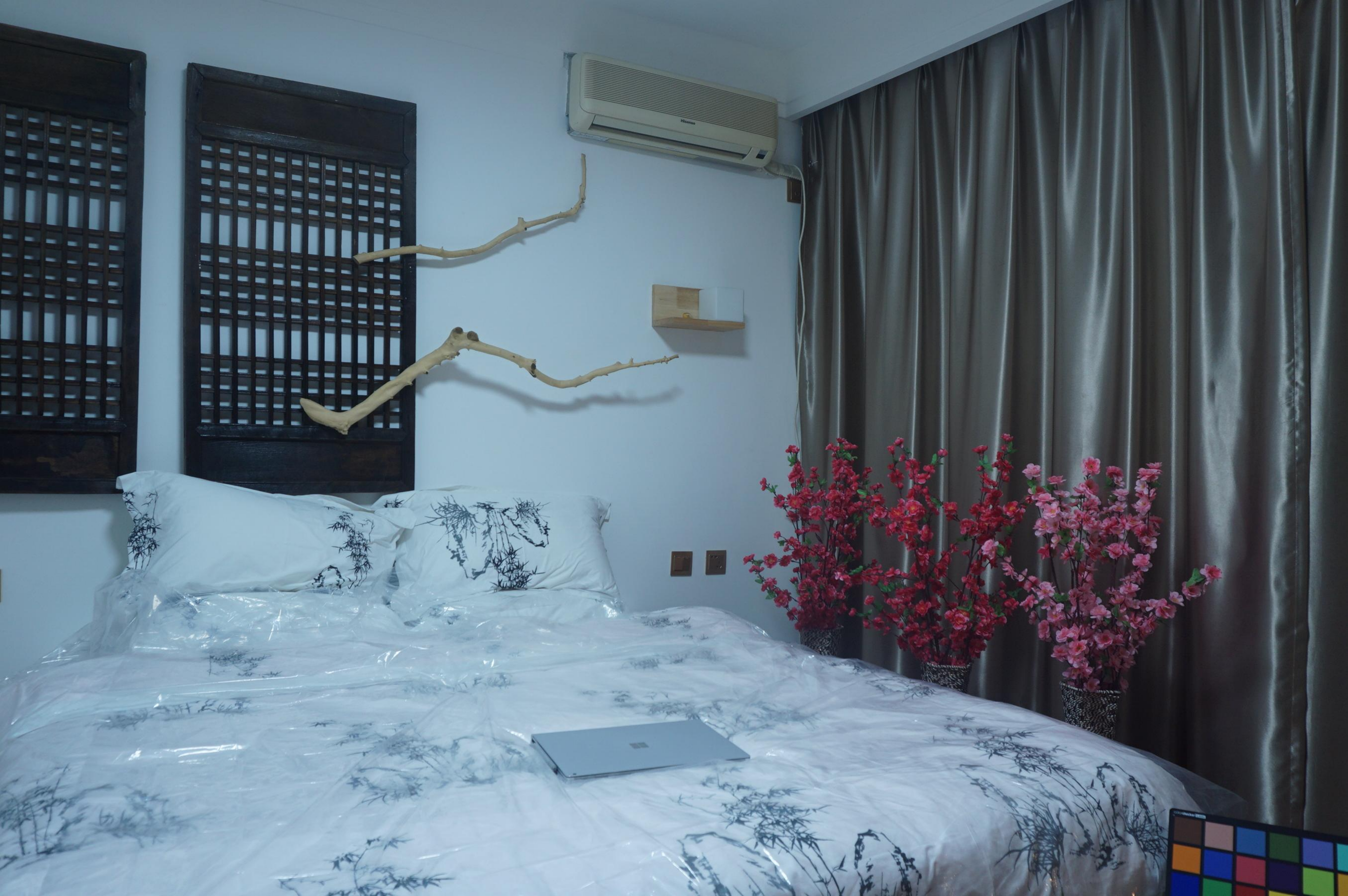}\\
            14.66/0.700  & \hspace{-0.46cm} 13.30/0.666 & \hspace{-0.46cm} 17.90/0.781& \hspace{-0.46cm} 16.58/0.609& \hspace{-0.46cm} 17.48/0.782& \hspace{-0.46cm}  17.01/0.664 & \hspace{-0.46cm}  22.23/0.832& \hspace{-0.46cm}  $+\infty$/1 \\
            \includegraphics[width = 0.12\linewidth, height = 0.1\linewidth]{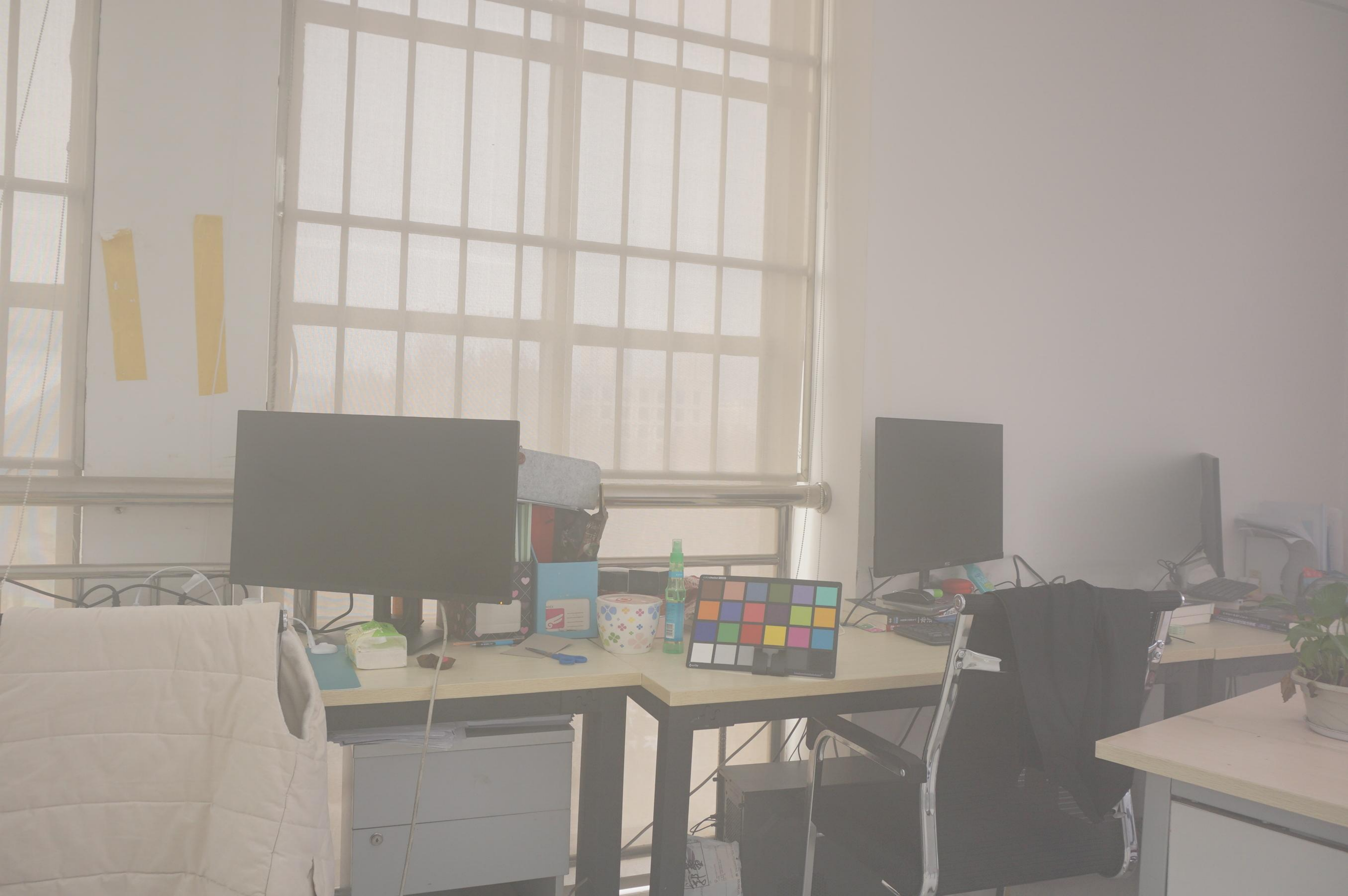}& \hspace{-0.46cm}
            \includegraphics[width = 0.12\linewidth, height = 0.1\linewidth]{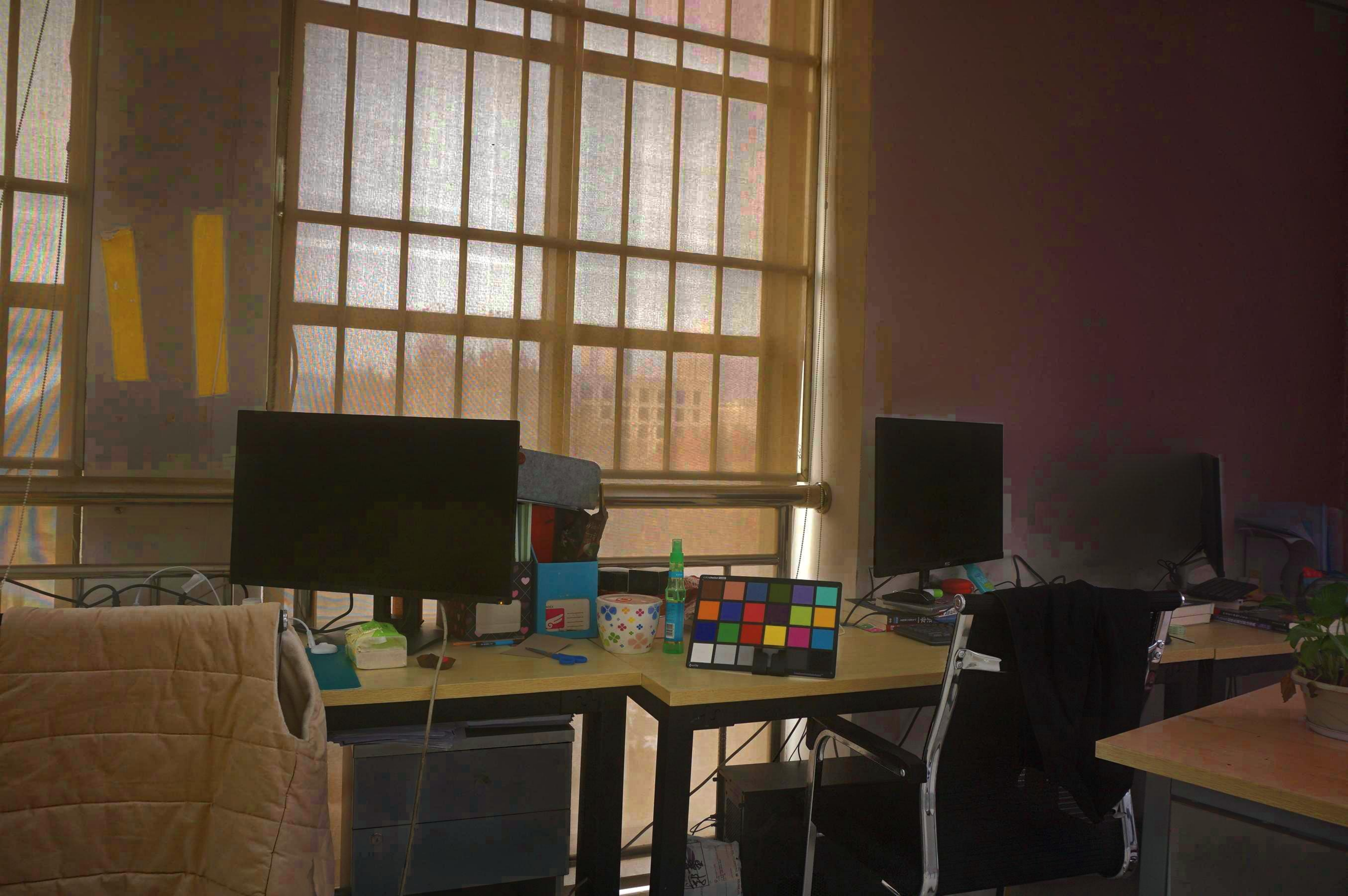} & \hspace{-0.46cm}
            \includegraphics[width = 0.12\linewidth, height = 0.1\linewidth]{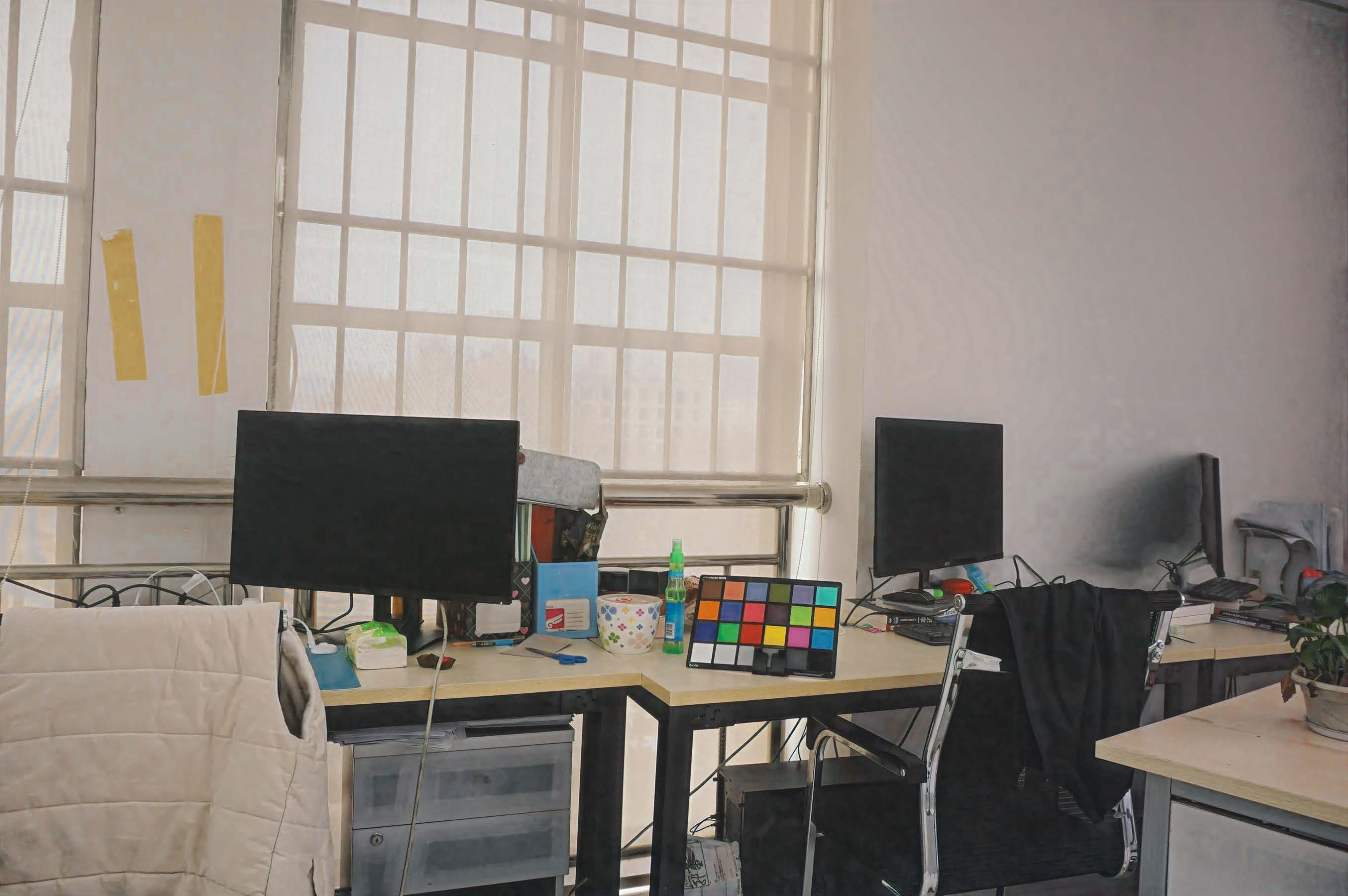} & \hspace{-0.46cm}
            \includegraphics[width = 0.12\linewidth, height = 0.1\linewidth]{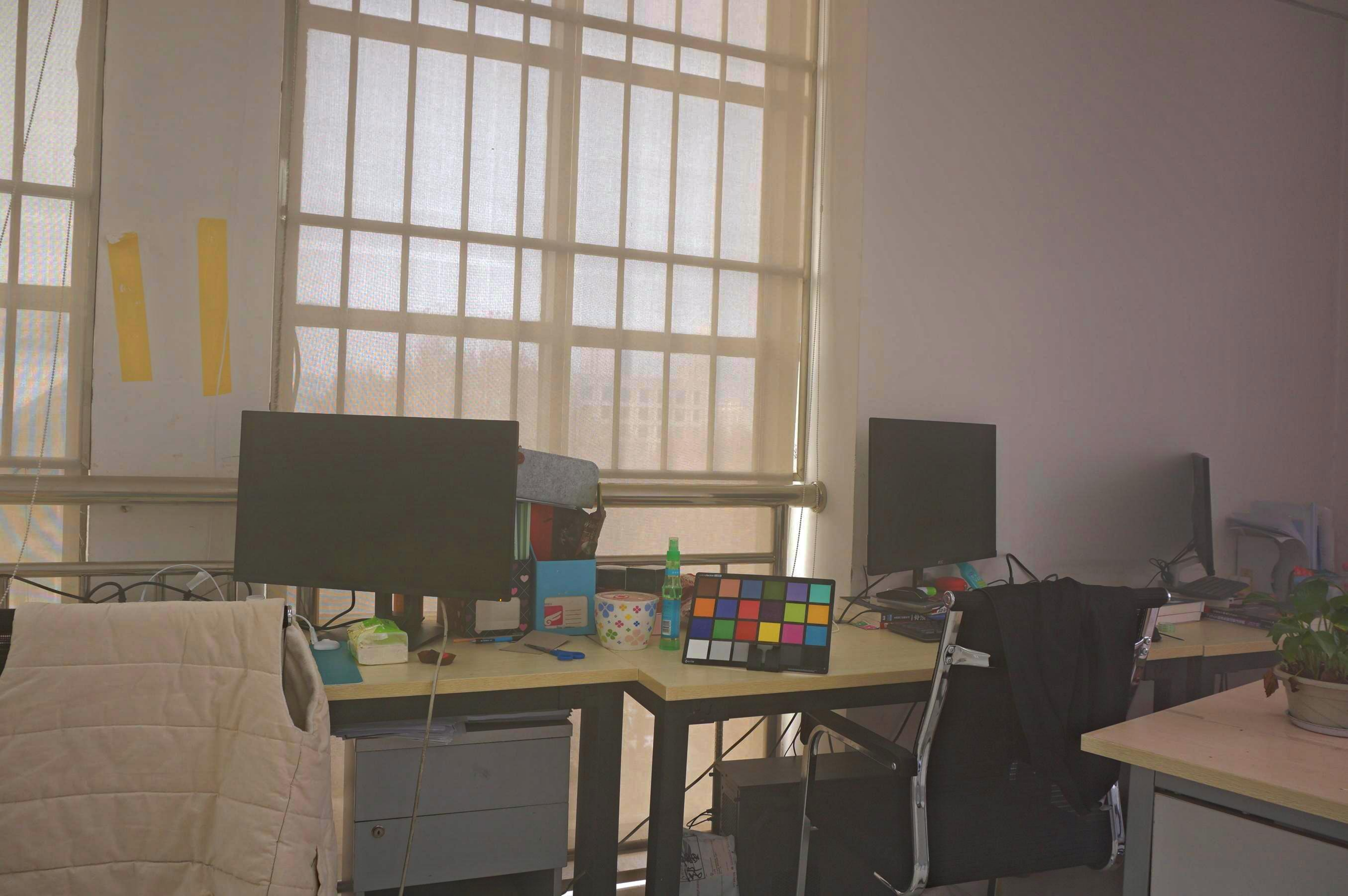} & \hspace{-0.46cm}
            \includegraphics[width = 0.12\linewidth, height = 0.1\linewidth]{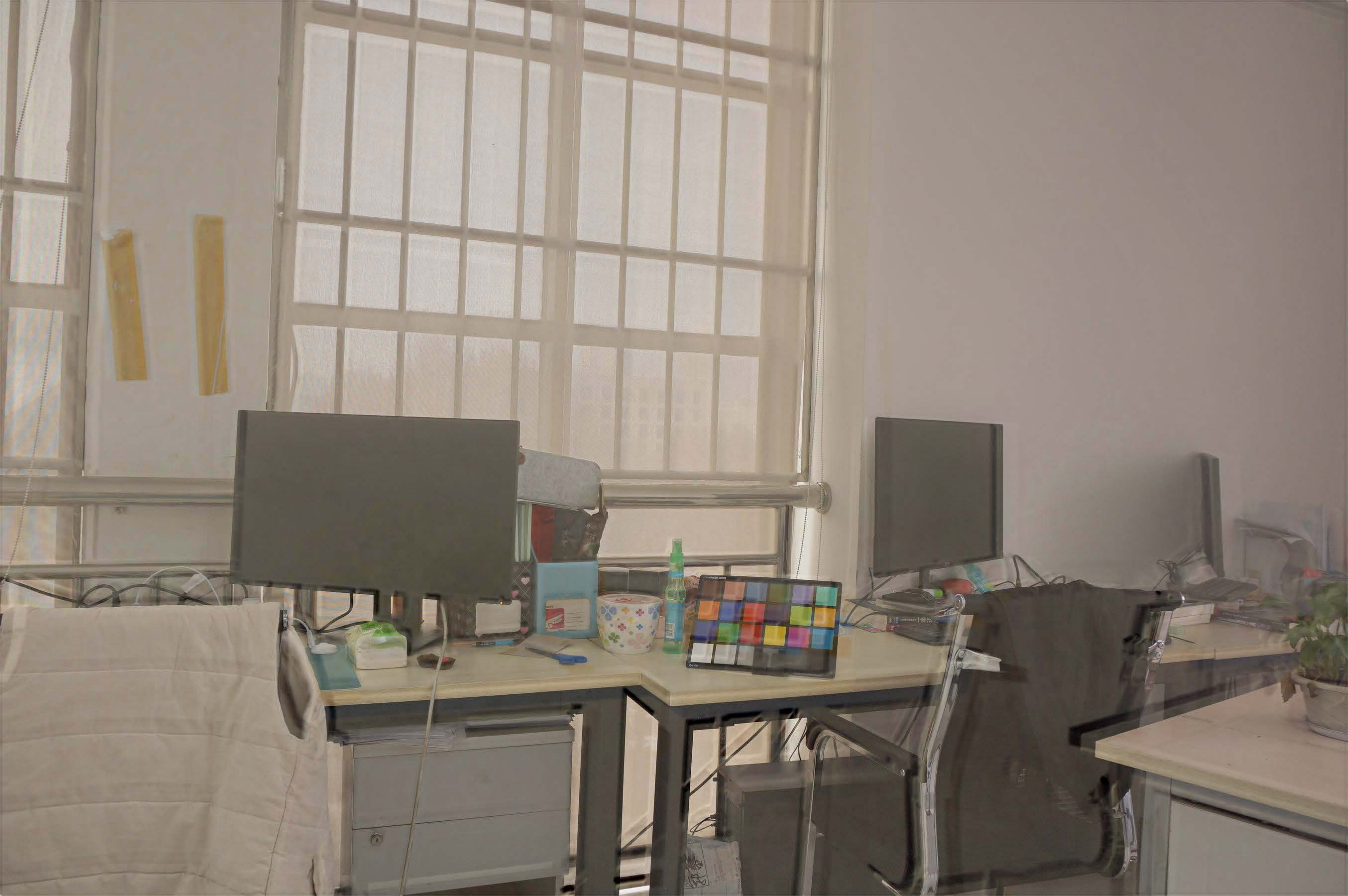} & \hspace{-0.46cm}
            \includegraphics[width = 0.12\linewidth, height = 0.1\linewidth]{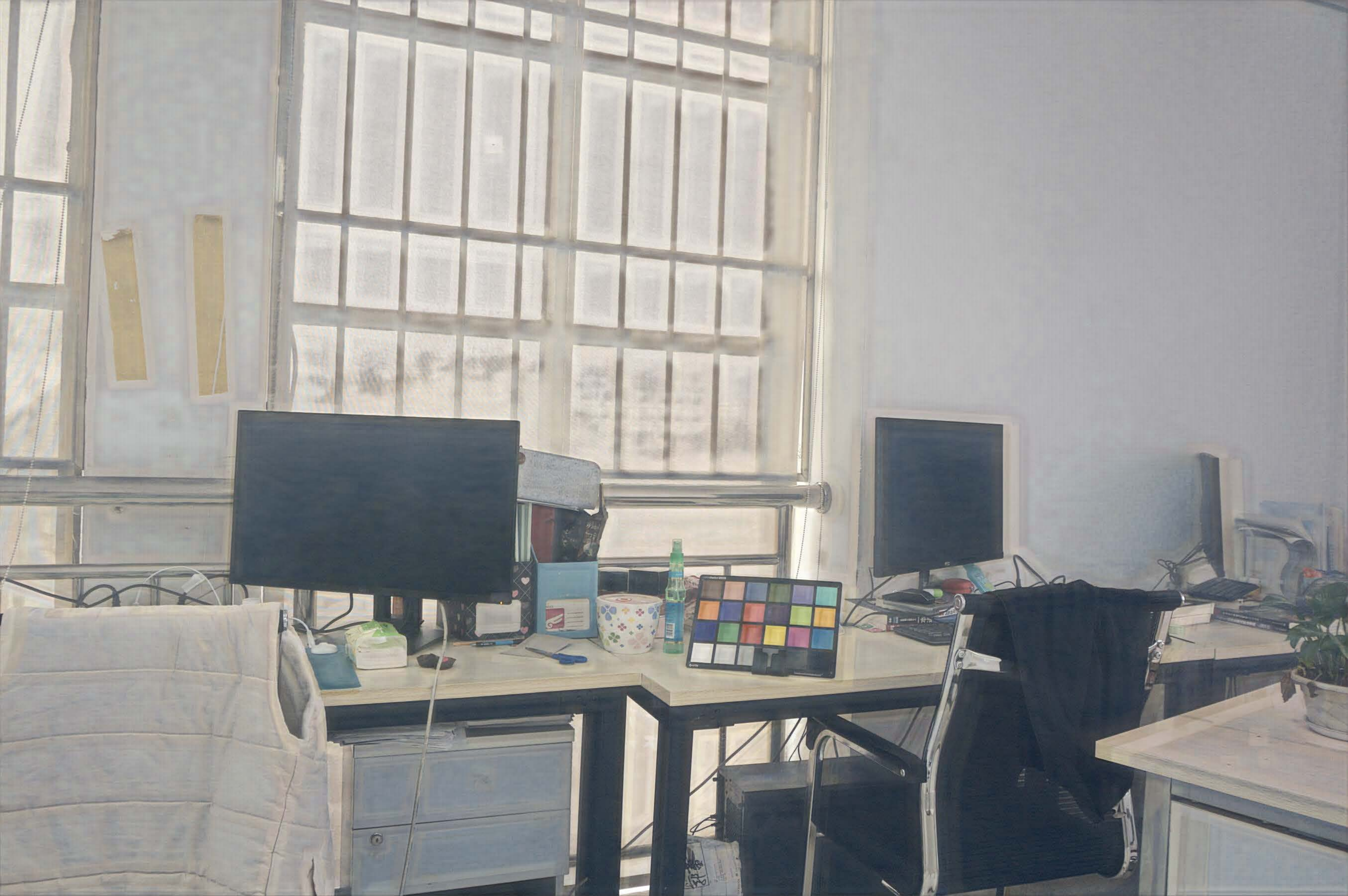} & \hspace{-0.46cm}
            \includegraphics[width = 0.12\linewidth, height = 0.1\linewidth]{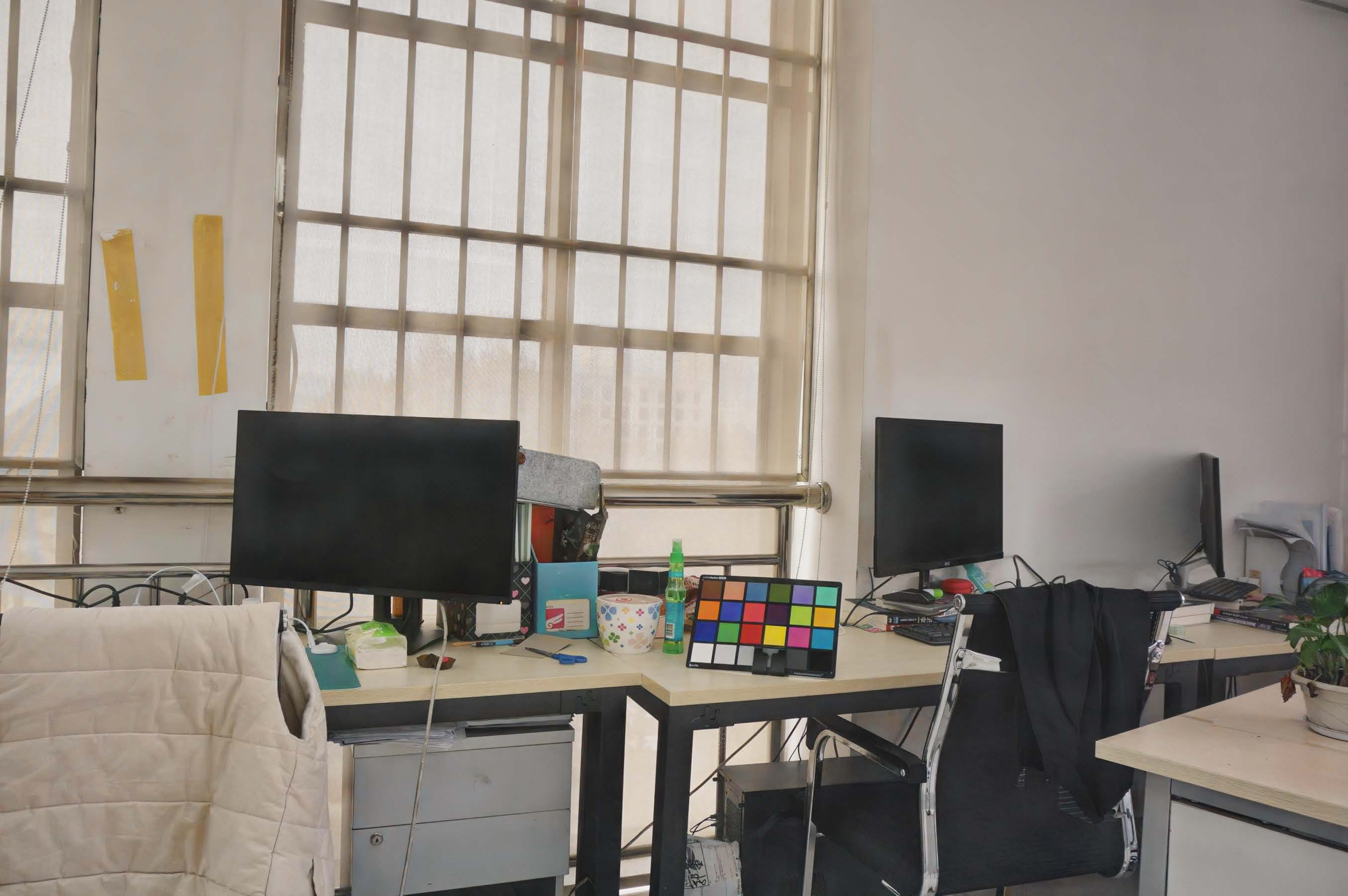} & \hspace{-0.46cm}
            \includegraphics[width = 0.12\linewidth, height = 0.1\linewidth]{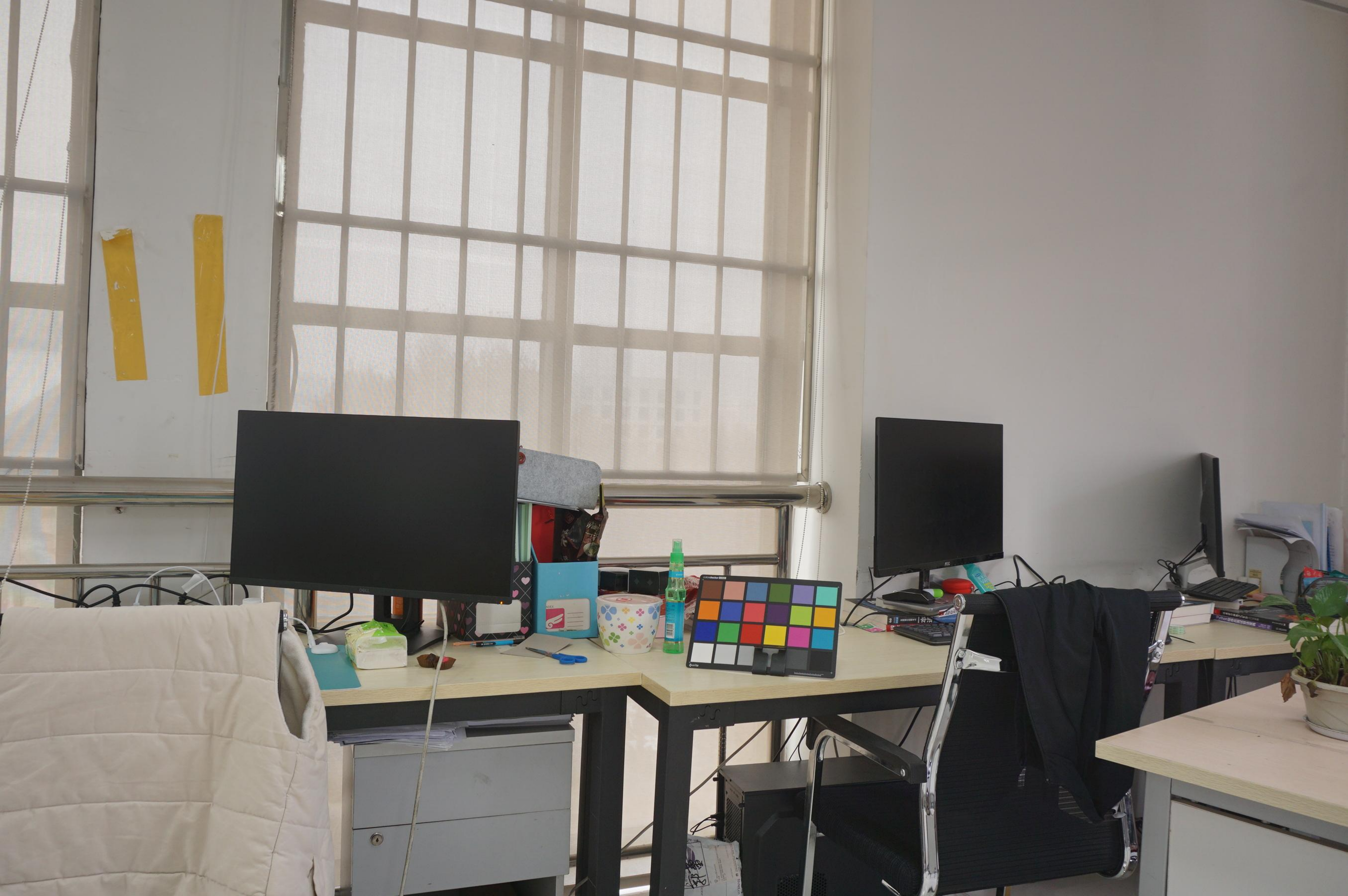}\\
			17.00/0.825  & \hspace{-0.46cm} 10.97/0.533 & \hspace{-0.46cm} 25.39/0.906& \hspace{-0.46cm} 17.07/0.654& \hspace{-0.46cm} 21.67/0.775& \hspace{-0.46cm}  19.86/0.681 & \hspace{-0.46cm}  27.09/0.924& \hspace{-0.46cm}  $+\infty$/1 \\
			(a) Original  & \hspace{-0.46cm} (b) DCP \cite{DBLP:journals/pami/He0T11} & \hspace{-0.46cm} (c) FFA \cite{DBLP:conf/aaai/QinWBXJ20} & \hspace{-0.46cm} (d) STMRF \cite{DBLP:conf/pcm/CaiXT16}& \hspace{-0.46cm} (e) EDVNet \cite{DBLP:conf/aaai/LiPWXF18}& \hspace{-0.46cm}  (f) FastDVD \cite{DBLP:conf/cvpr/TassanoDV20}& \hspace{-0.46cm}  (g) Ours& \hspace{-0.46cm}  (h) Ground Truth \\
		\end{tabular}
	\end{center}
	%\vspace{-0.5cm}
	\caption{ Examples from the REVIDE test dataset and corresponding dehazed frames by several state-of-the-art methods. PSNR/SSIM values are below each frame. The dehazed frames by our method contain fewer haze residuals and artifacts.}
	\label{fig:frame1}
\end{figure*}
%%%%%%%%%%%%%%%%%%
In training process, we randomly select five consecutive neighboring frames as input and crop them to patches with $512\times512$ pixel.
The learning rate is set to be 0.0001 and decreases 0.1 times at the 200-th eopch. The weights $\alpha$ and $\beta$ are empirically set to be 10 and 1, respectively.
The ``Adam" optimizer~\cite{DBLP:journals/corr/KingmaB14} is used to optimize the training process. The proposed algorithm is deployed on the ``Pytorch" platform.

\begin{figure*}[!htp]
	\footnotesize
	\begin{center}
		\begin{tabular}{ccccccc}
			%\vspace{-0.2cm}
			\includegraphics[width = 0.14\linewidth, height = 0.13\linewidth]{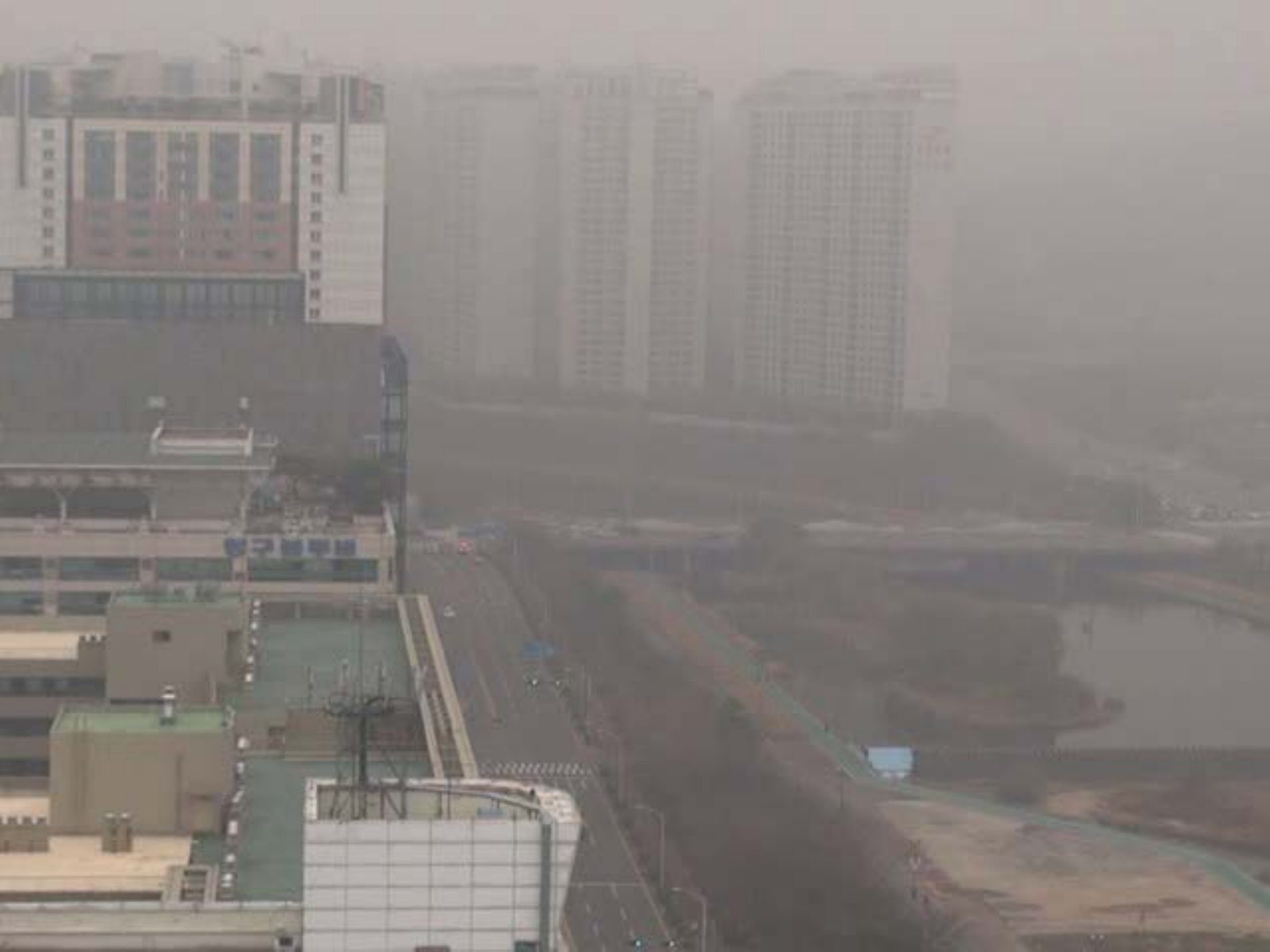}& \hspace{-0.46cm}
			\includegraphics[width = 0.14\linewidth, height = 0.13\linewidth]{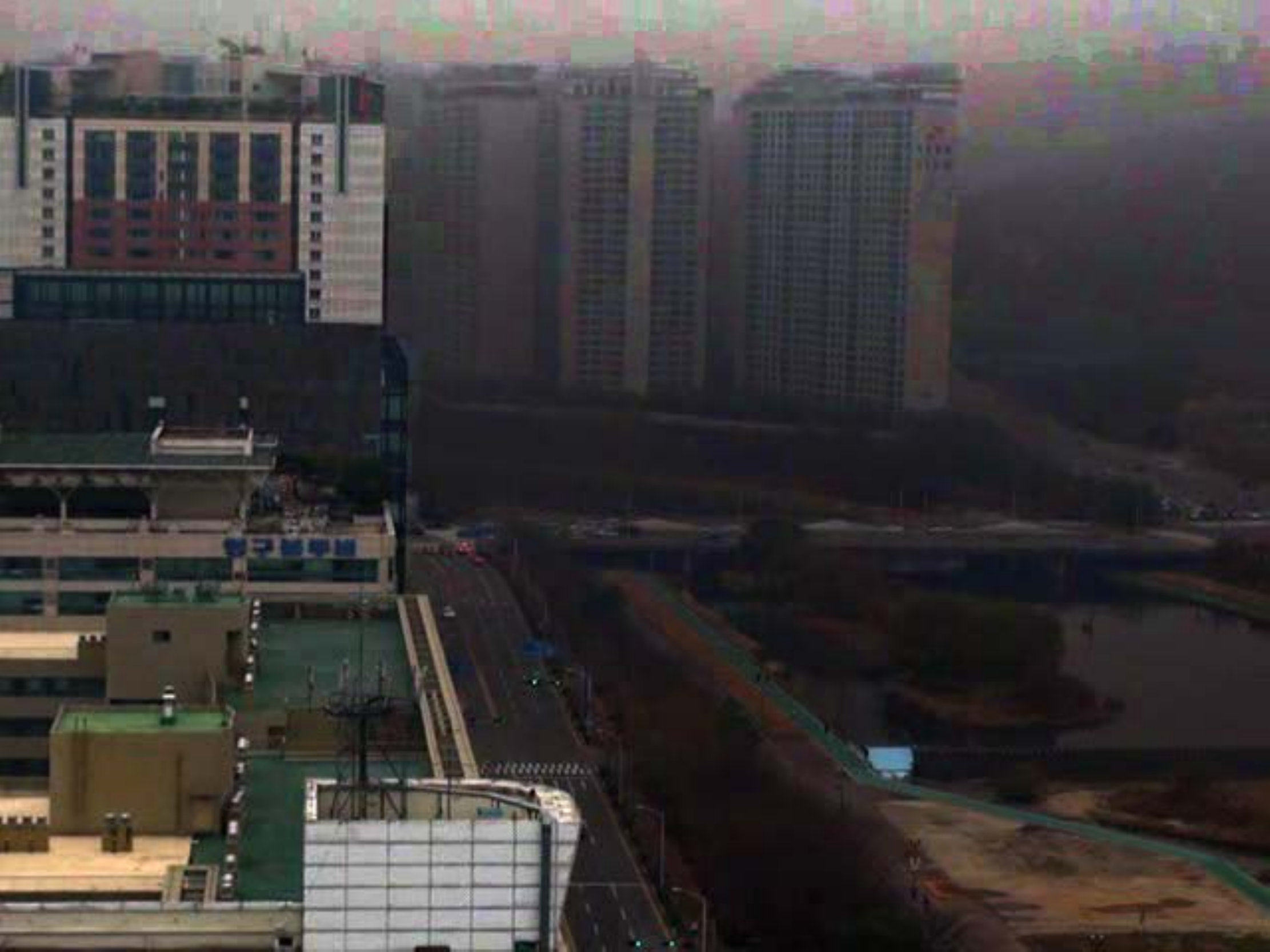} & \hspace{-0.46cm}
			\includegraphics[width = 0.14\linewidth, height = 0.13\linewidth]{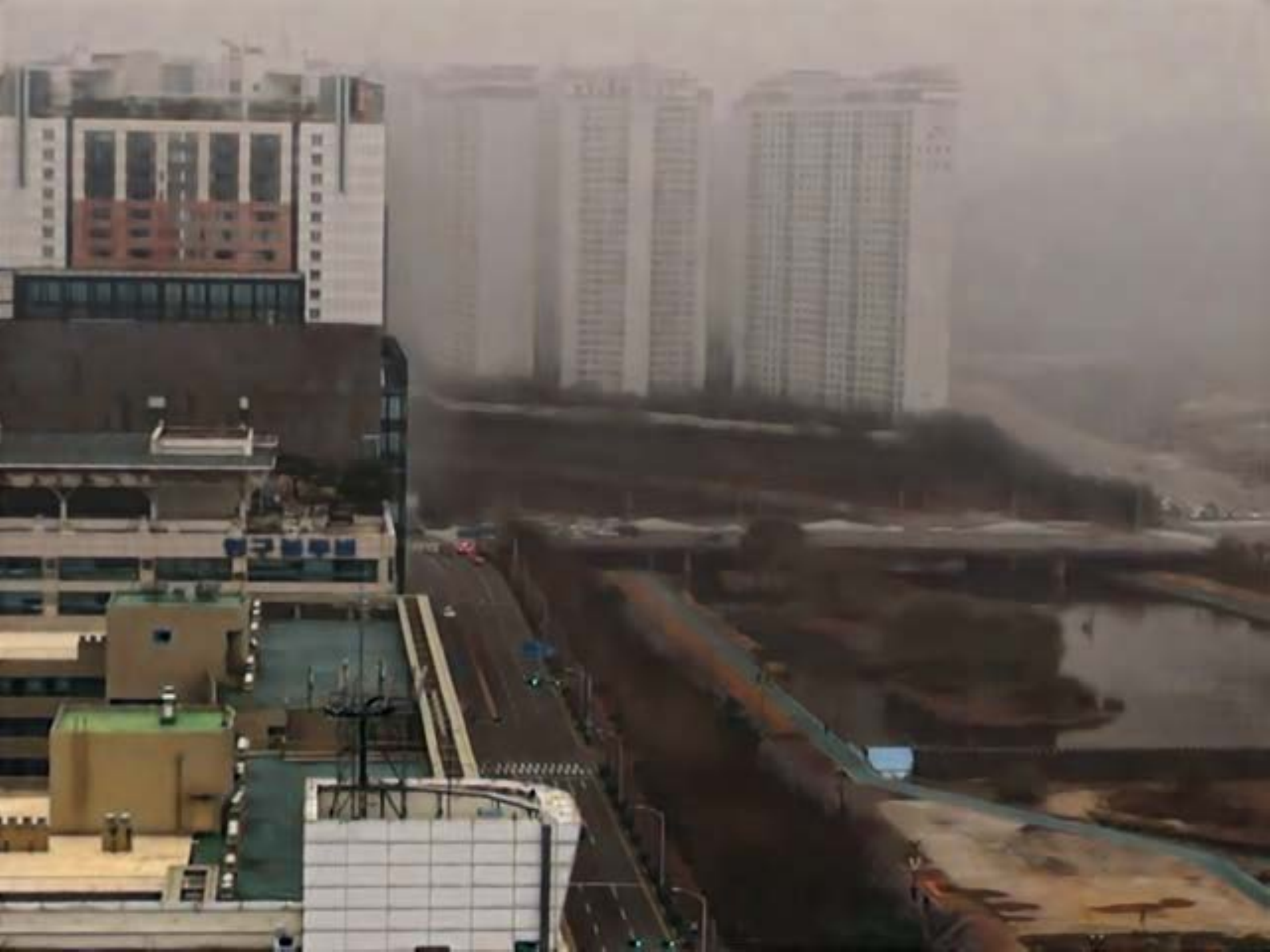} & \hspace{-0.46cm}
			\includegraphics[width = 0.14\linewidth, height = 0.13\linewidth]{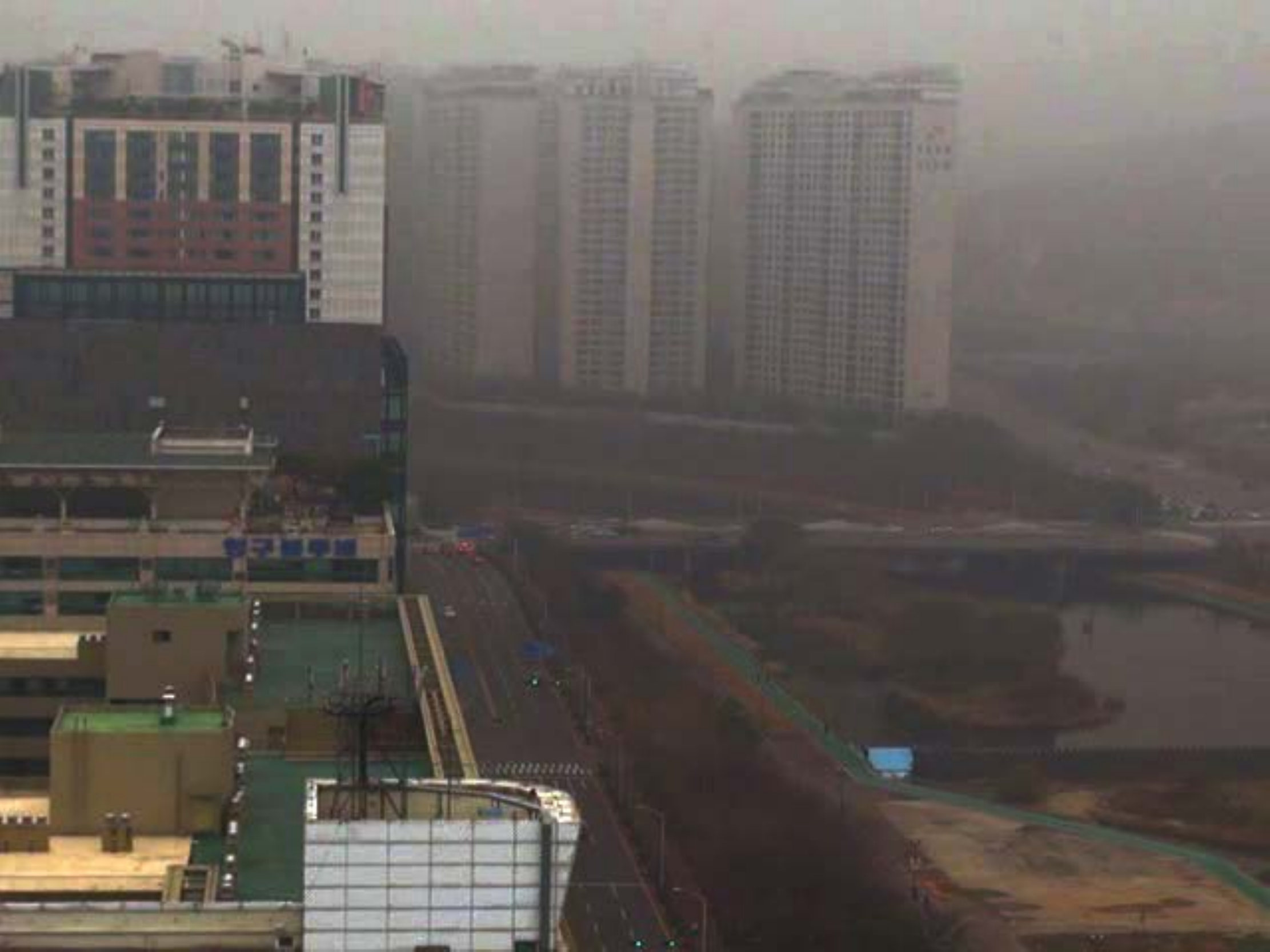} & \hspace{-0.46cm}
			\includegraphics[width = 0.14\linewidth, height = 0.13\linewidth]{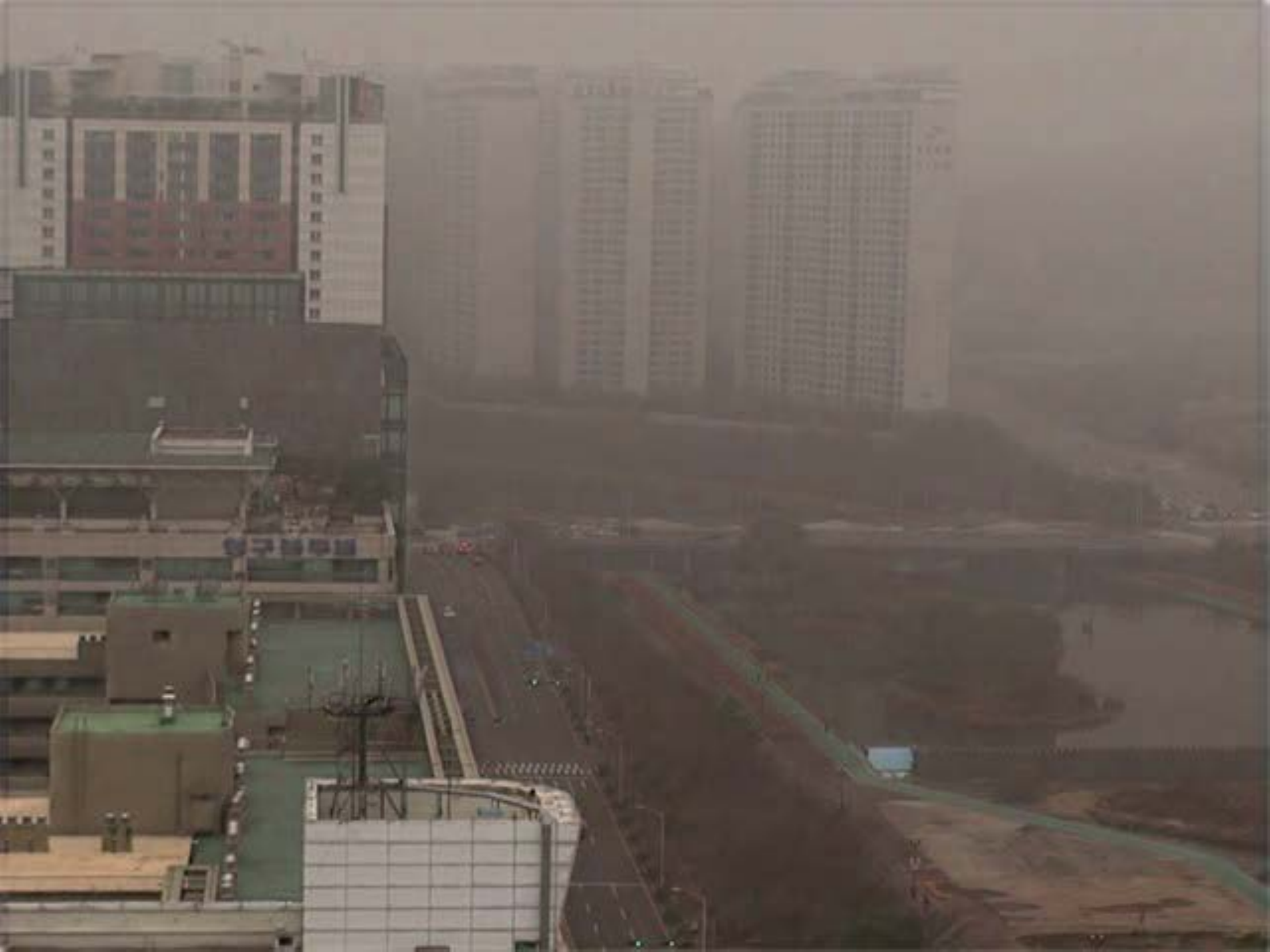} & \hspace{-0.46cm}
			\includegraphics[width = 0.14\linewidth, height = 0.13\linewidth]{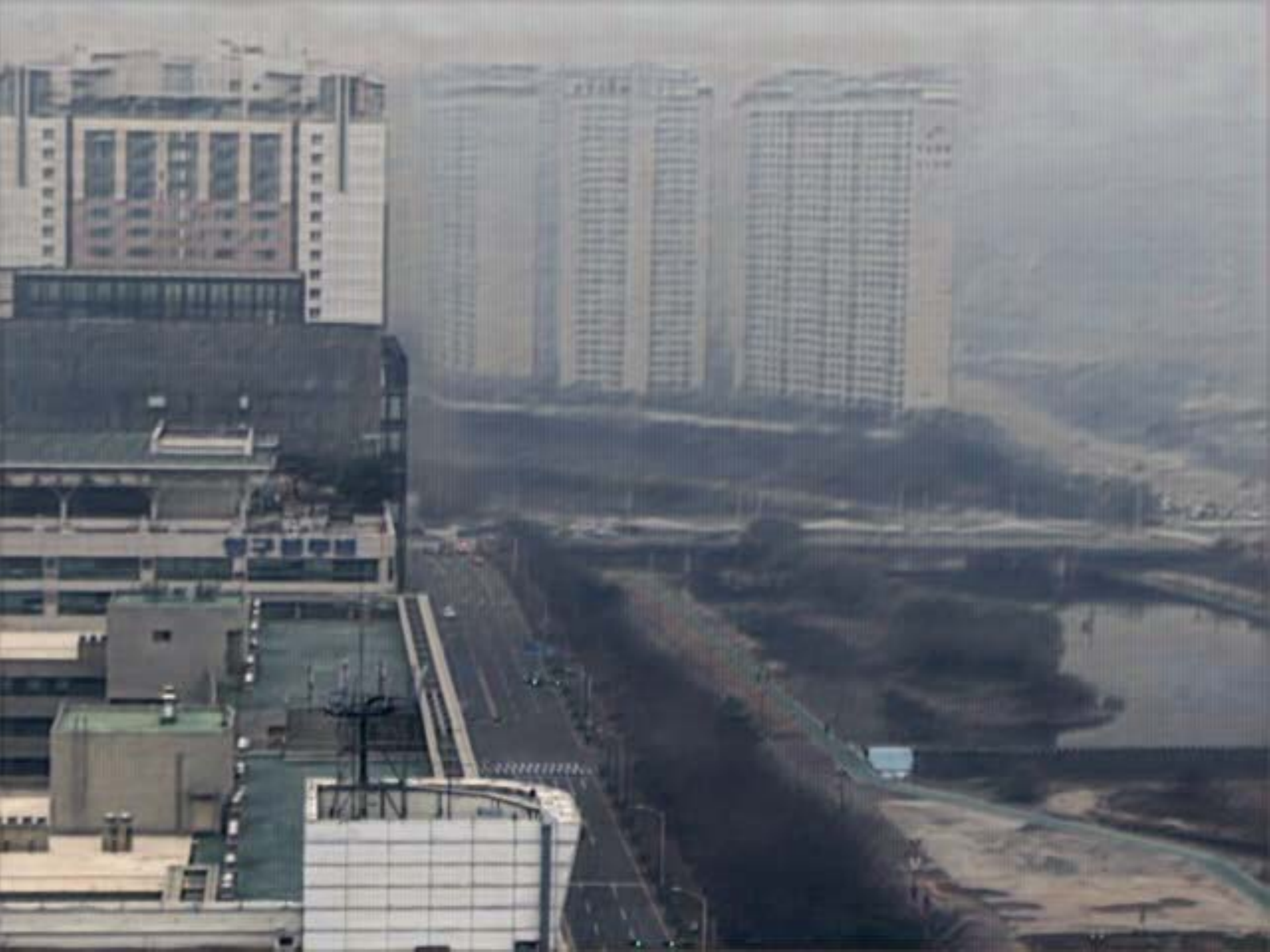} & \hspace{-0.46cm}
			\includegraphics[width = 0.14\linewidth, height = 0.13\linewidth]{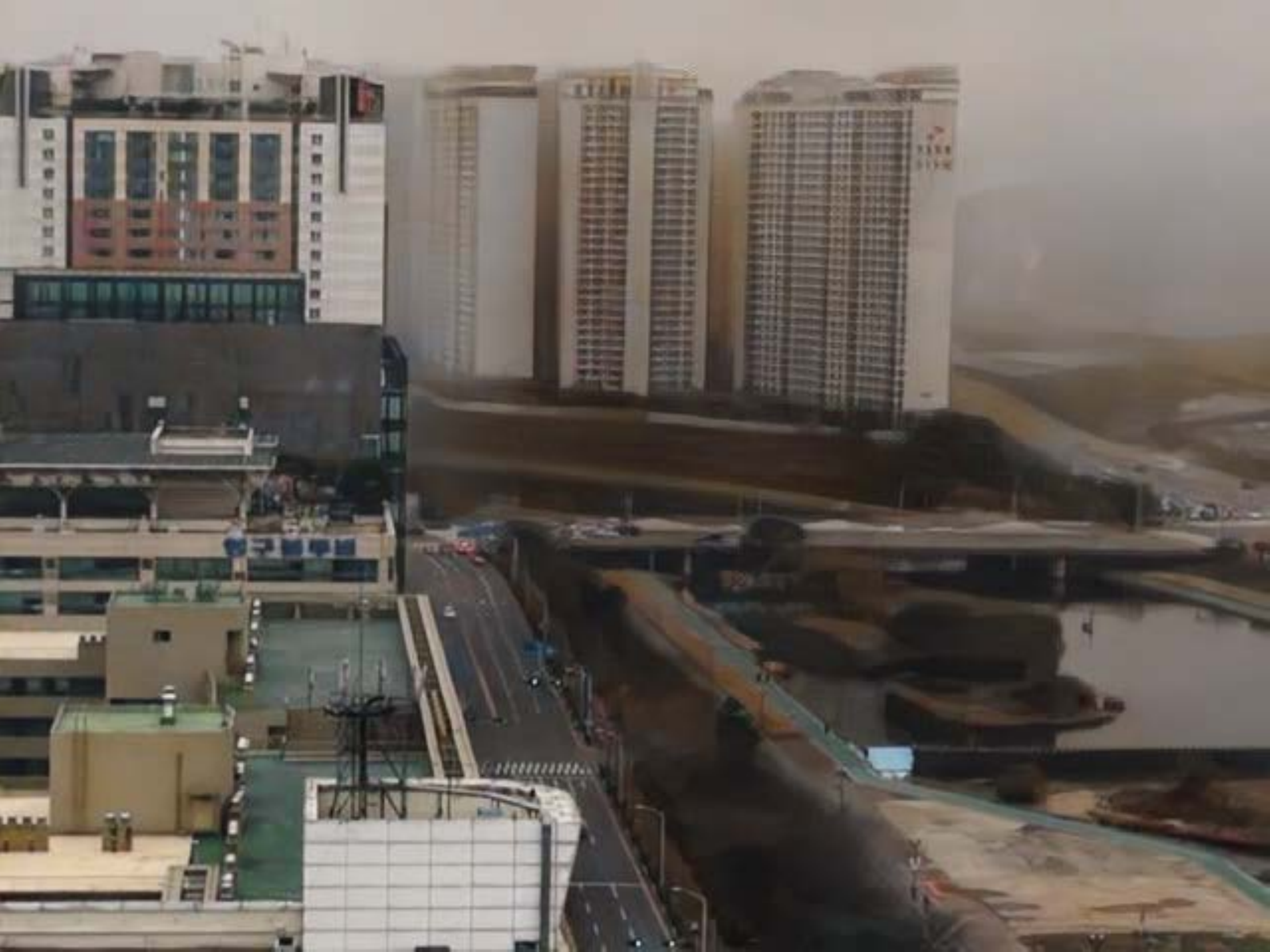}\\
		    \includegraphics[width = 0.14\linewidth, height = 0.13\linewidth]{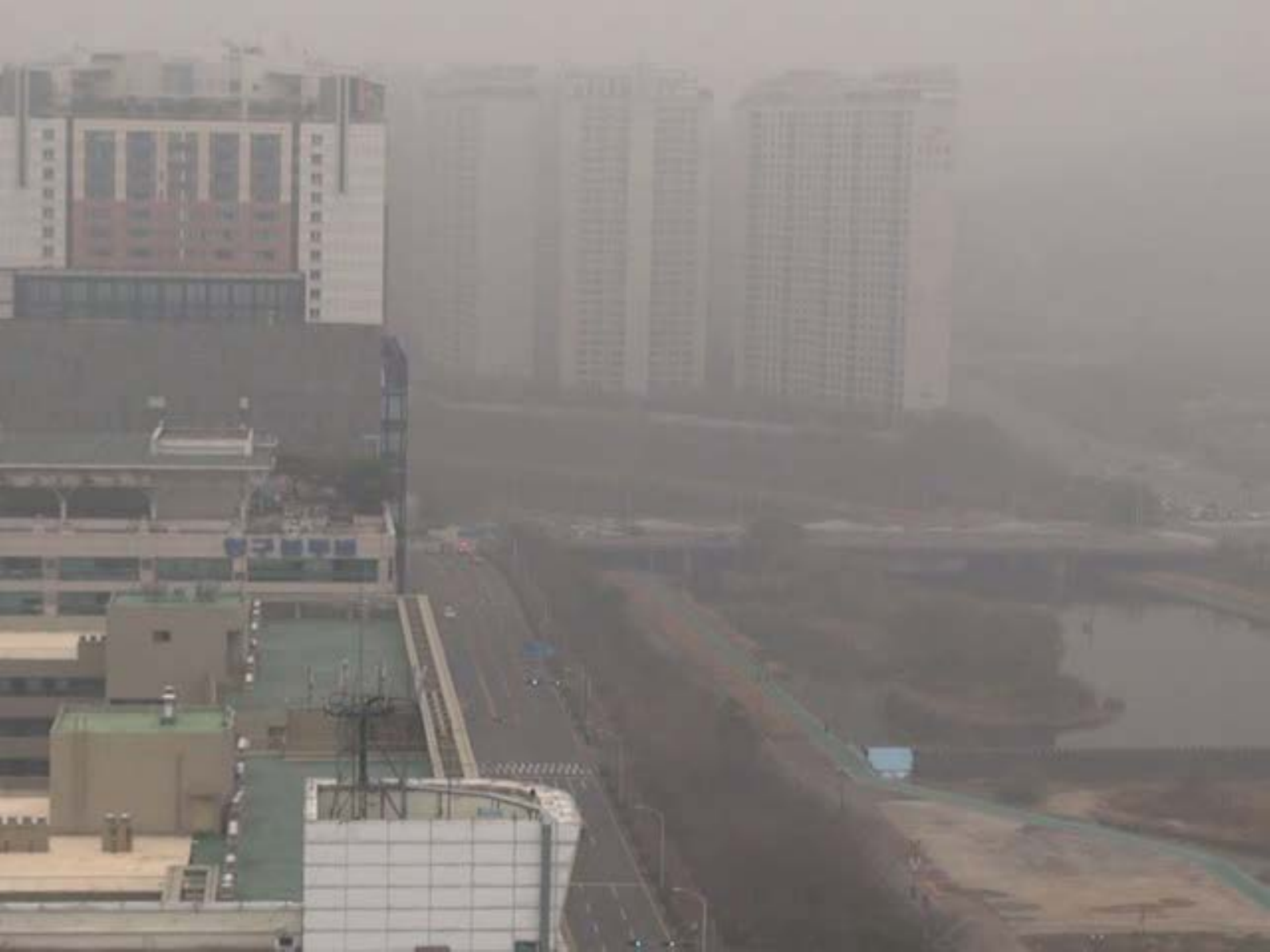}& \hspace{-0.46cm}
			\includegraphics[width = 0.14\linewidth, height = 0.13\linewidth]{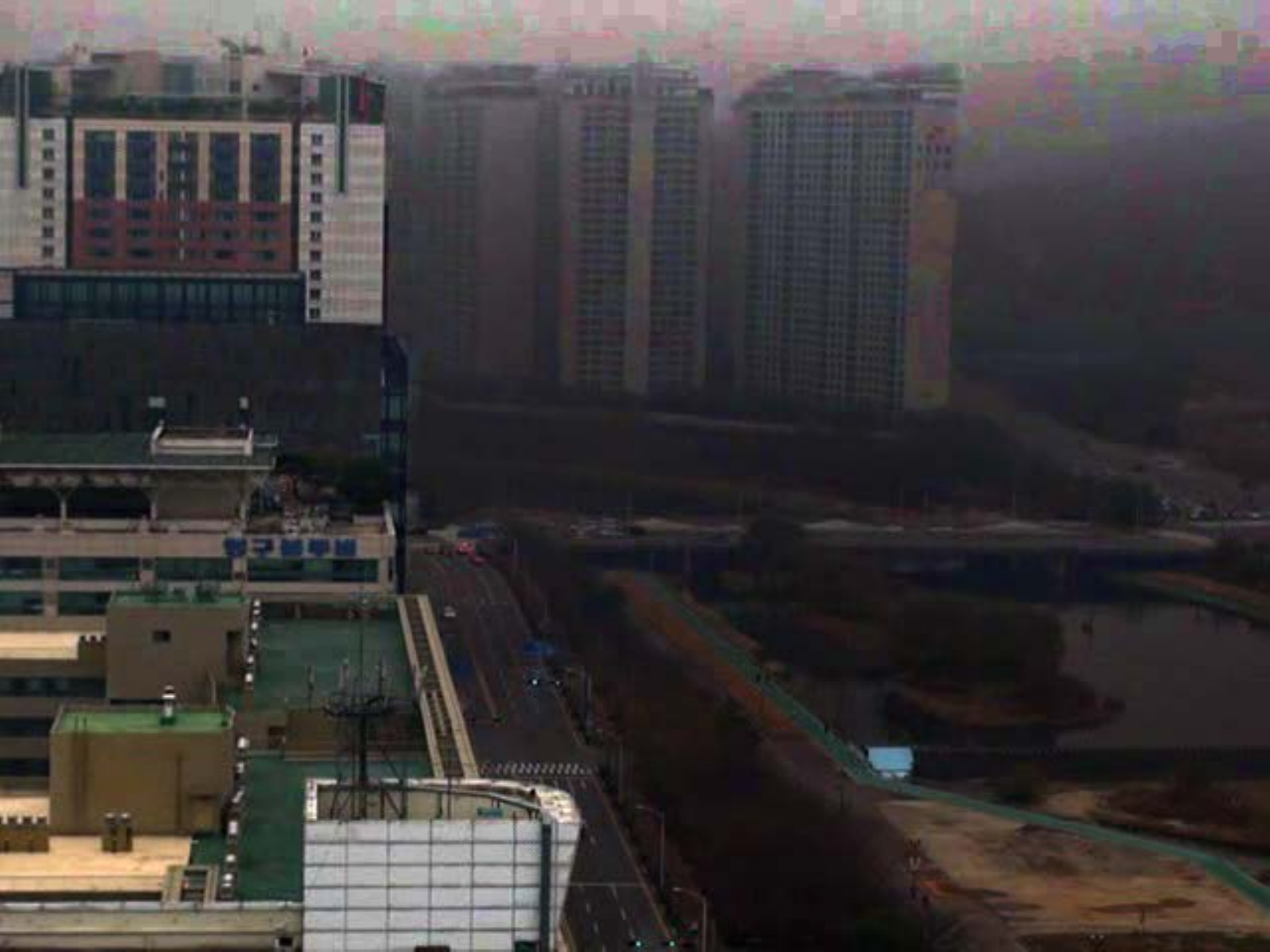} & \hspace{-0.46cm}
			\includegraphics[width = 0.14\linewidth, height = 0.13\linewidth]{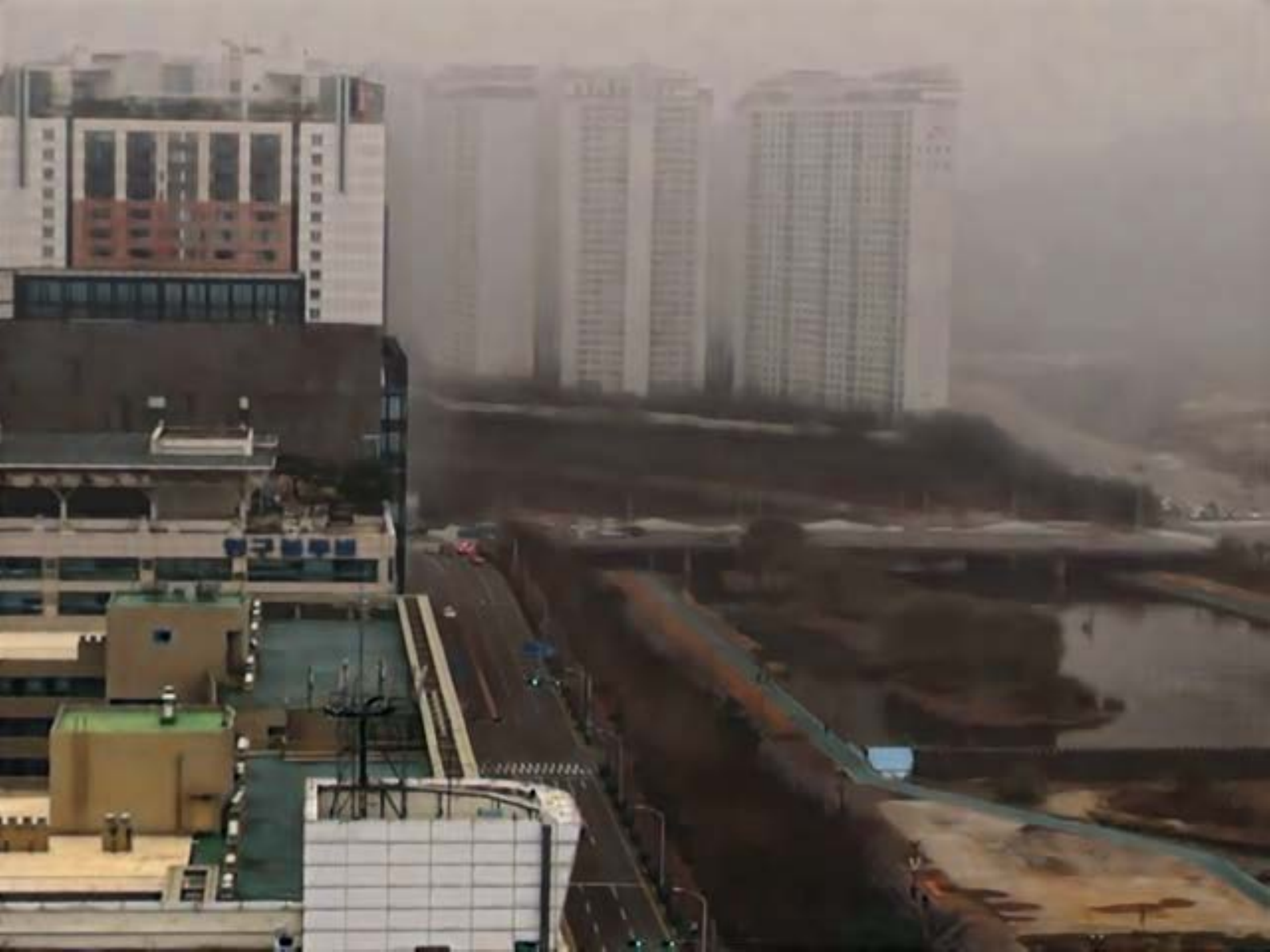} & \hspace{-0.46cm}
			\includegraphics[width = 0.14\linewidth, height = 0.13\linewidth]{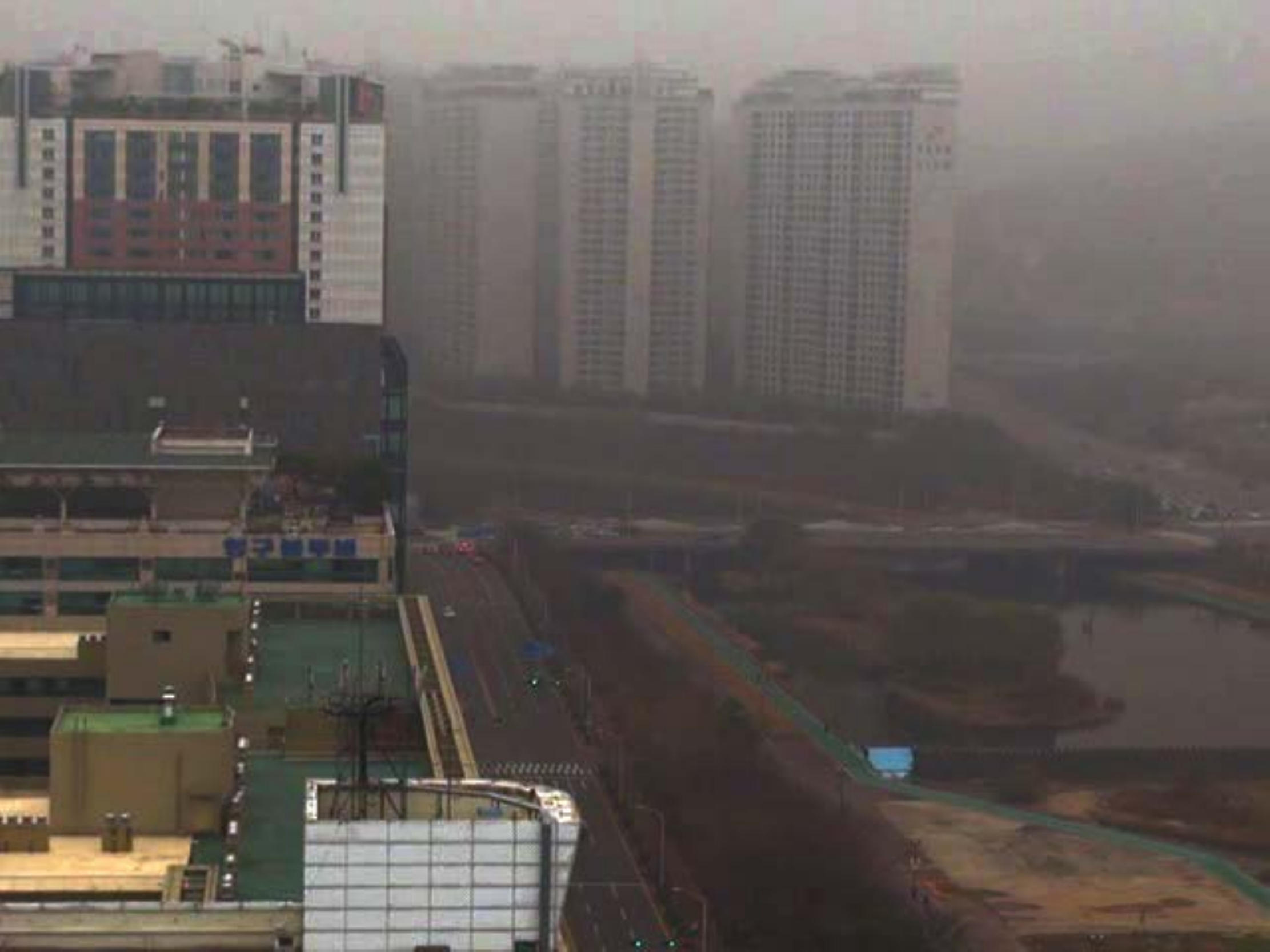} & \hspace{-0.46cm}
			\includegraphics[width = 0.14\linewidth, height = 0.13\linewidth]{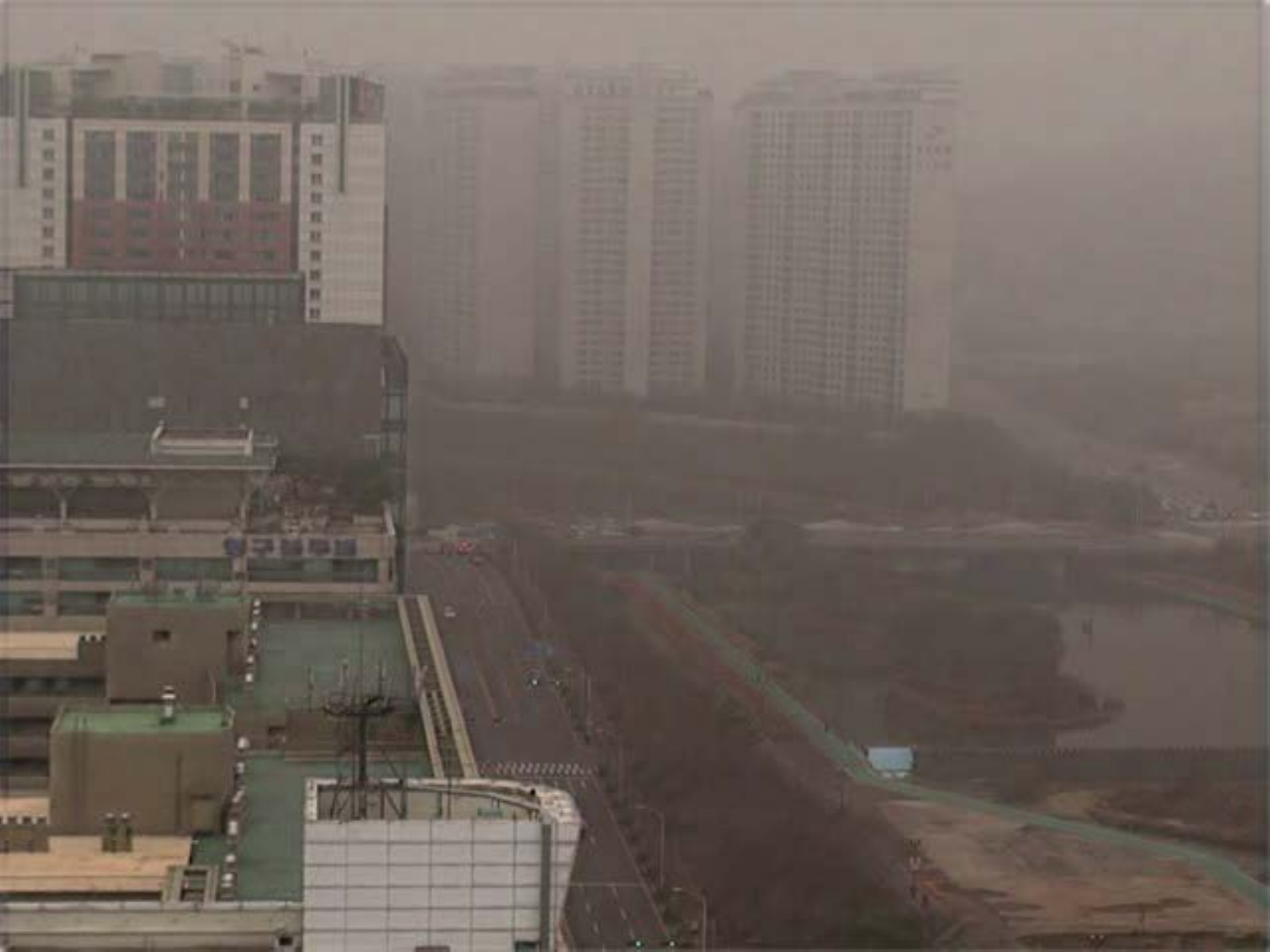} & \hspace{-0.46cm}
			\includegraphics[width = 0.14\linewidth, height = 0.13\linewidth]{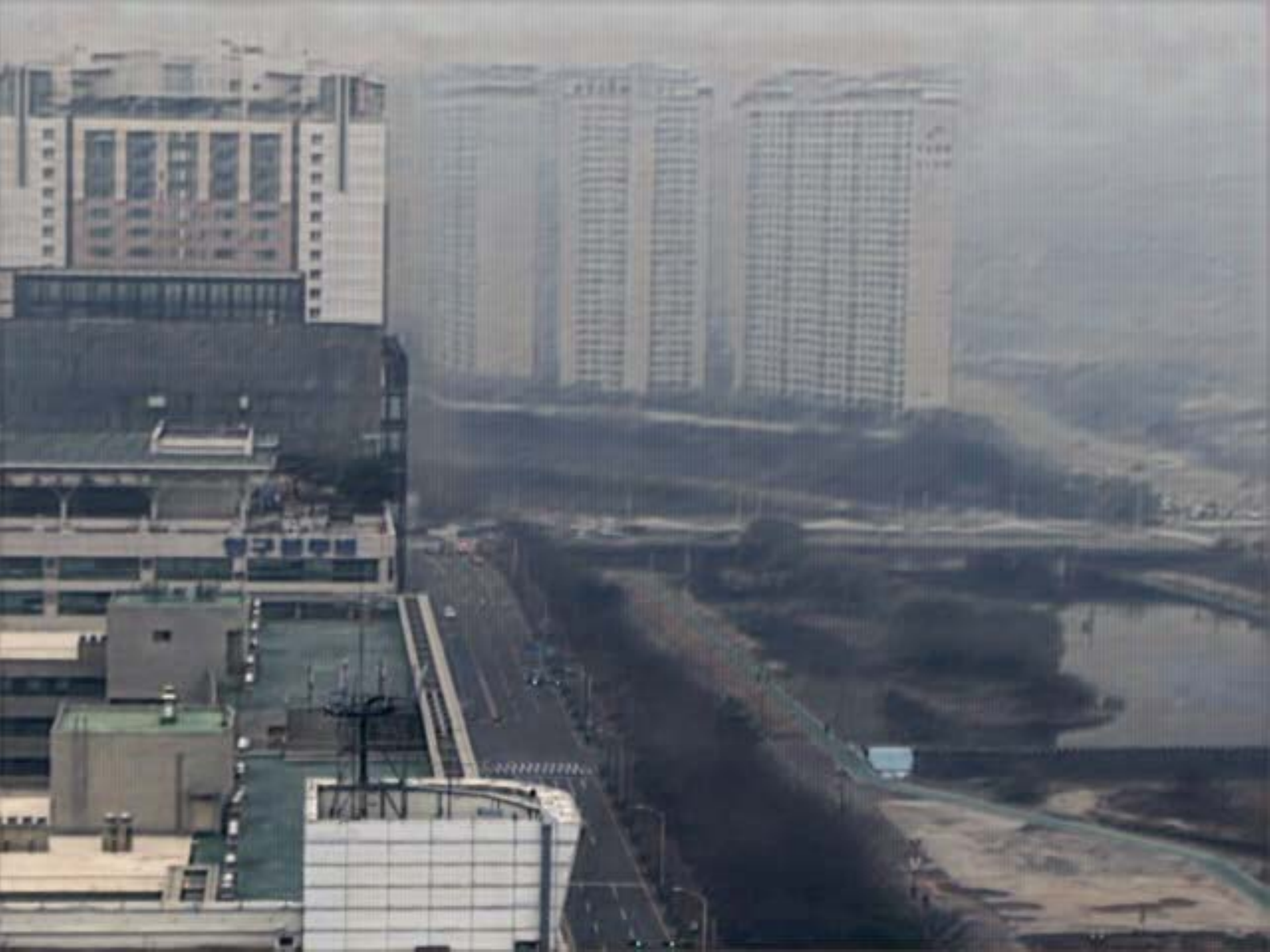} & \hspace{-0.46cm}
			\includegraphics[width = 0.14\linewidth, height = 0.13\linewidth]{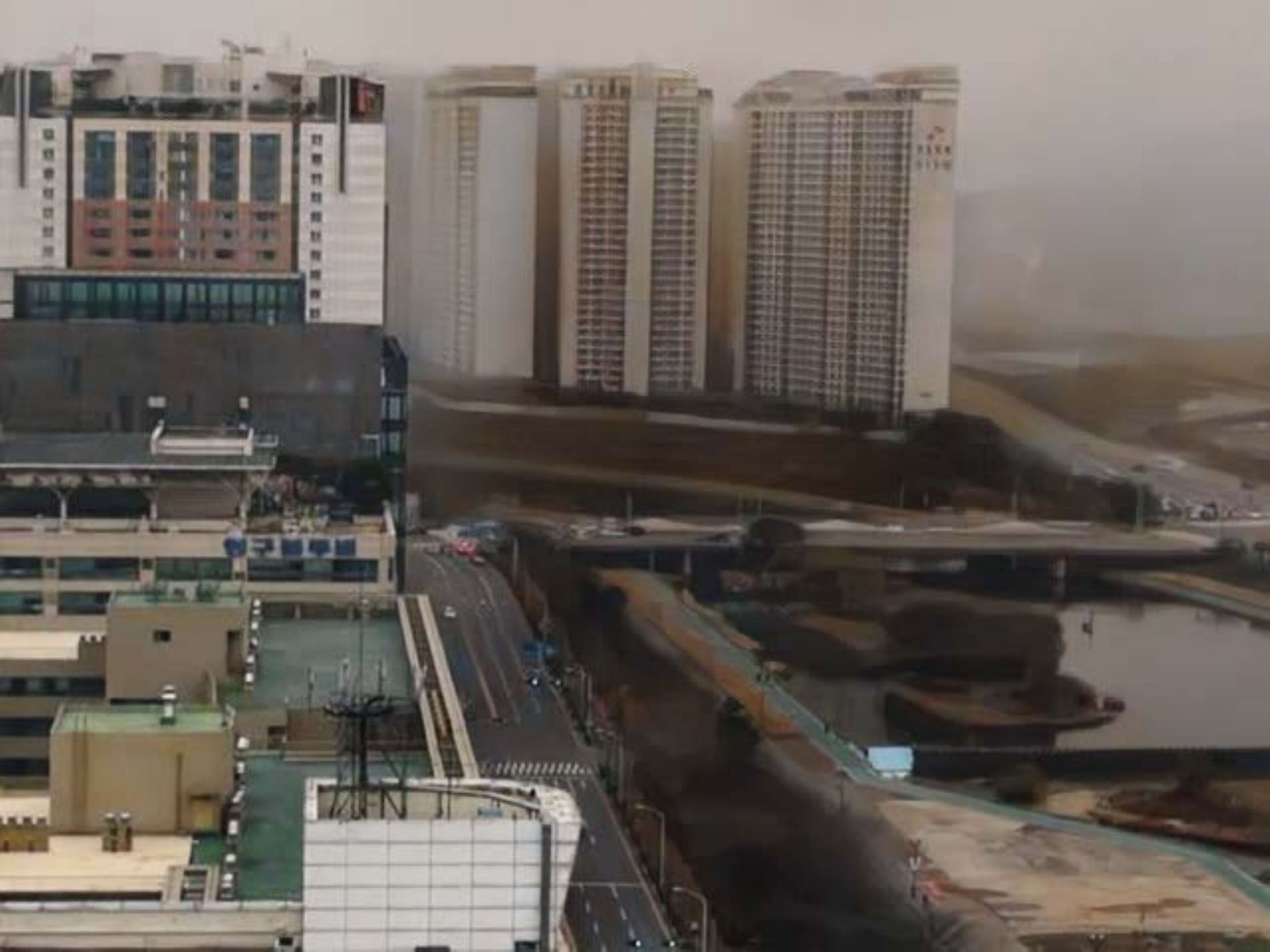}\\
			\includegraphics[width = 0.14\linewidth, height = 0.13\linewidth]{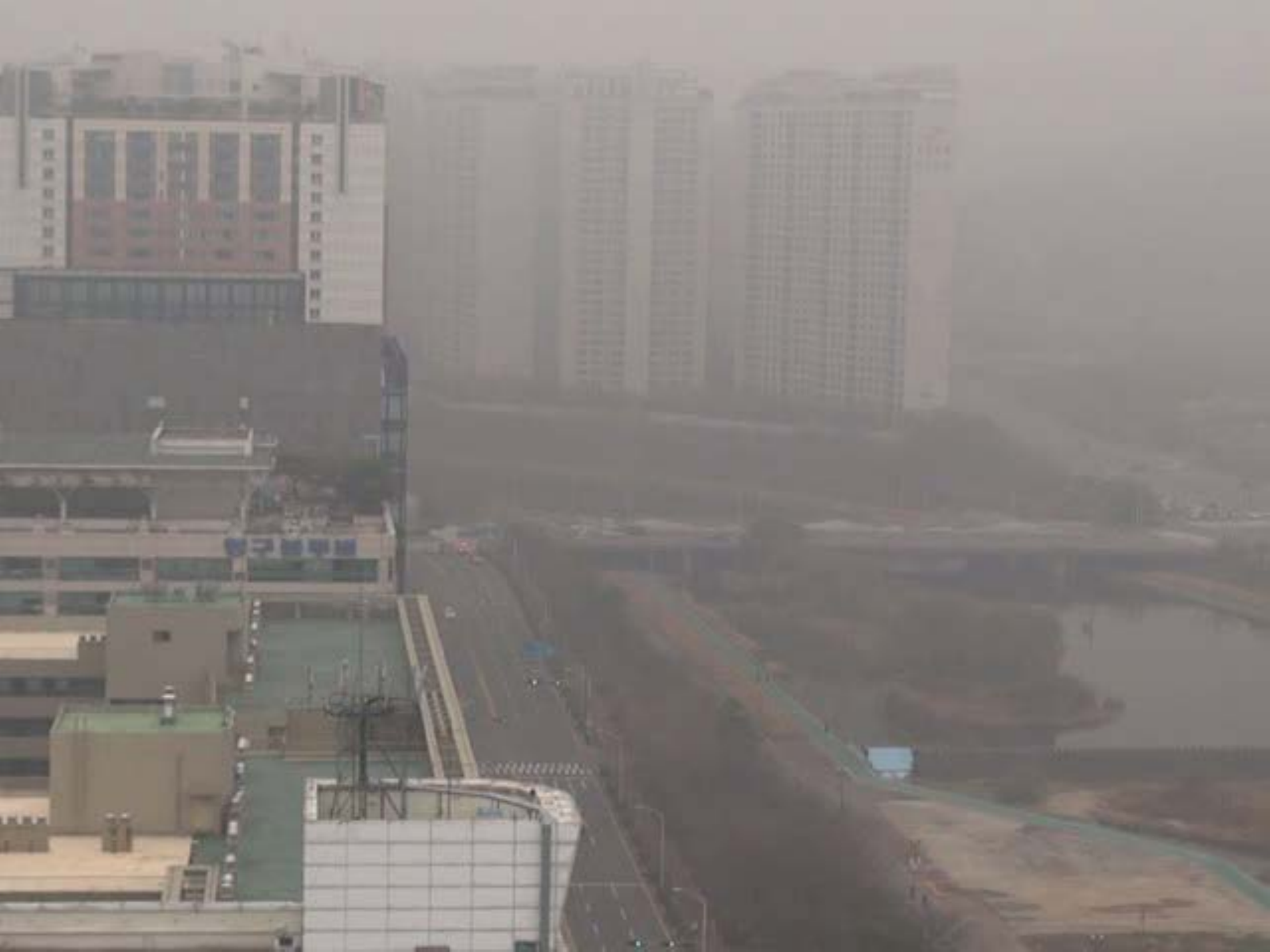}& \hspace{-0.46cm}
			\includegraphics[width = 0.14\linewidth, height = 0.13\linewidth]{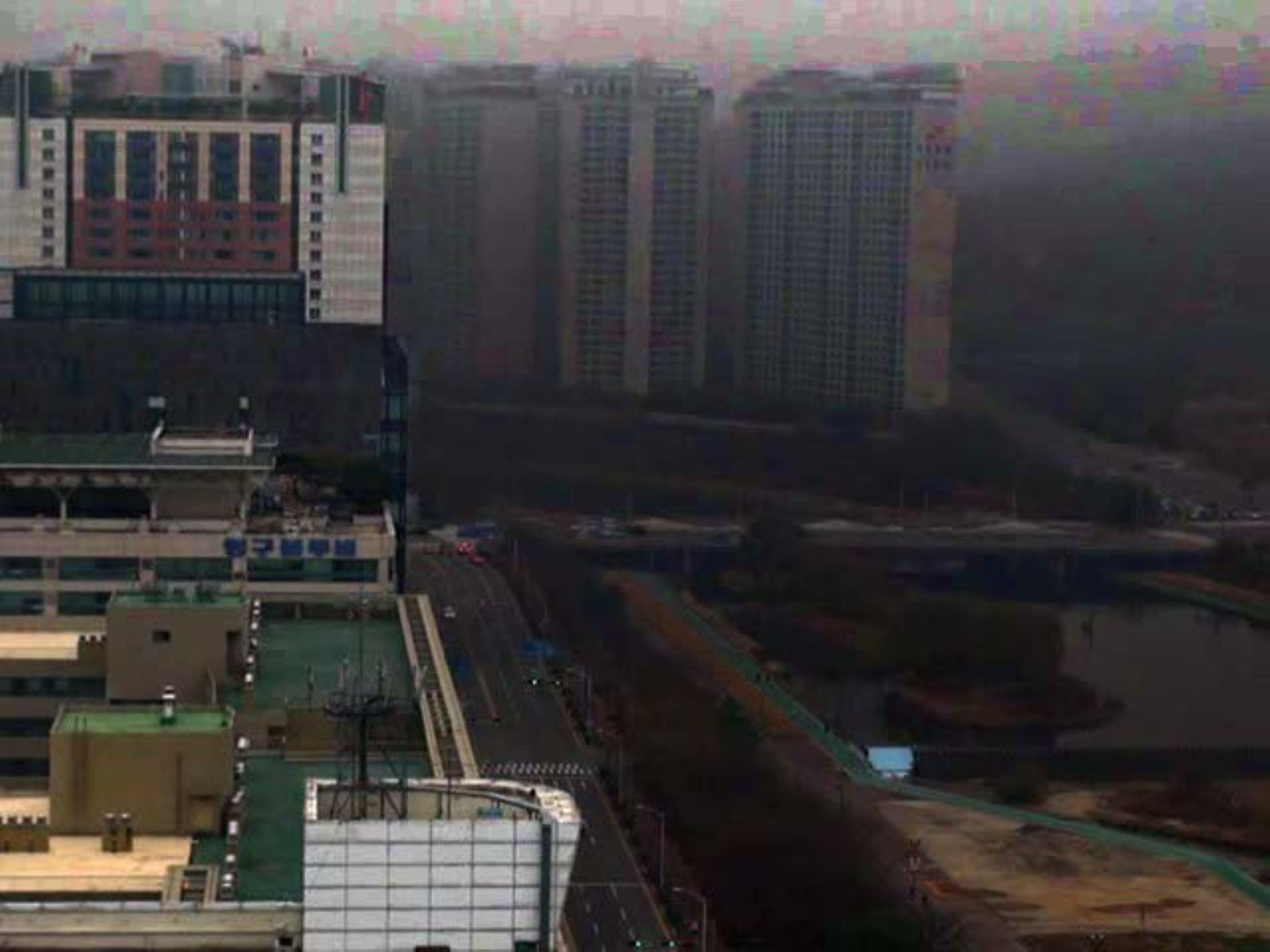} & \hspace{-0.46cm}
			\includegraphics[width = 0.14\linewidth, height = 0.13\linewidth]{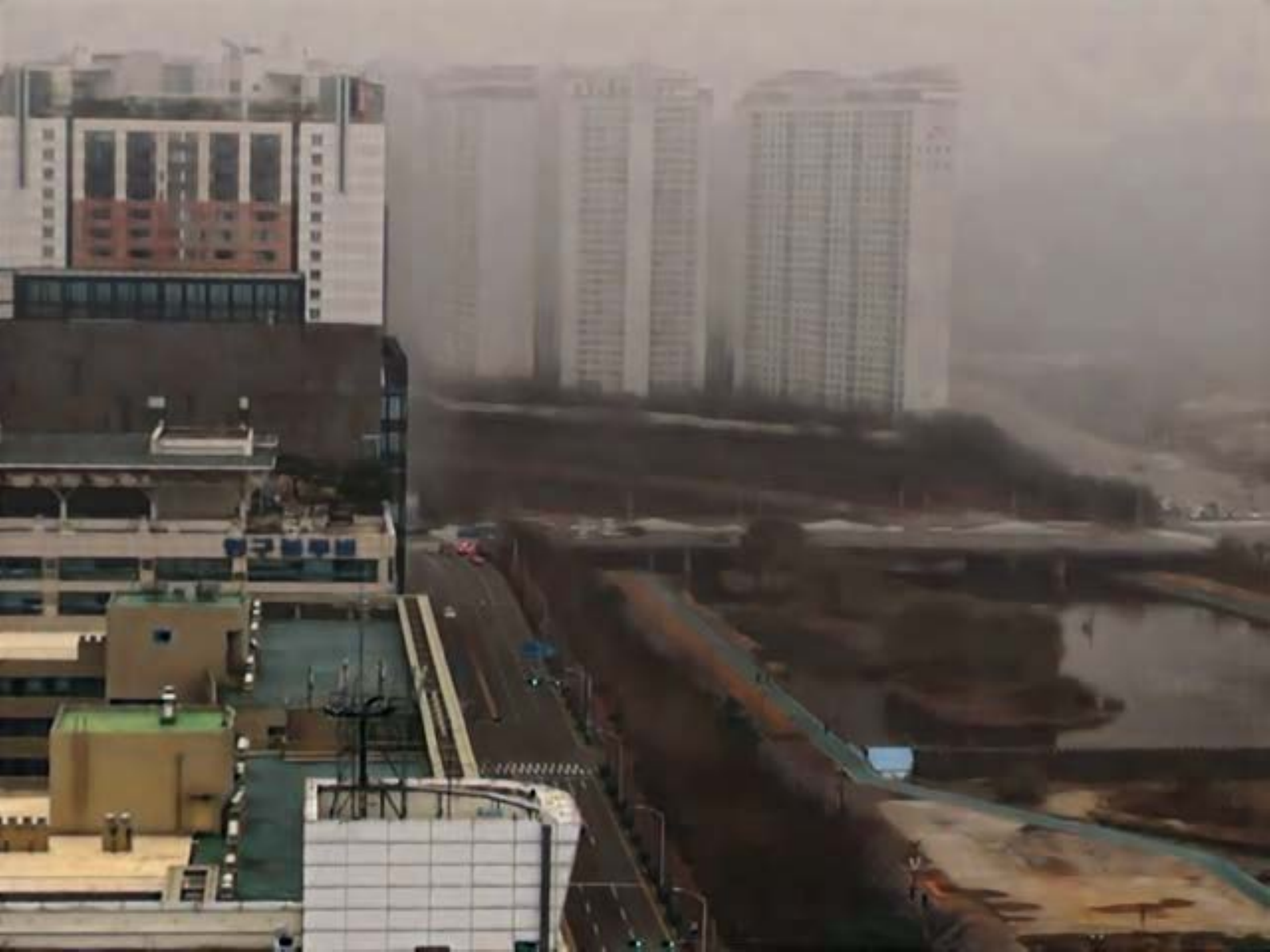} & \hspace{-0.46cm}
			\includegraphics[width = 0.14\linewidth, height = 0.13\linewidth]{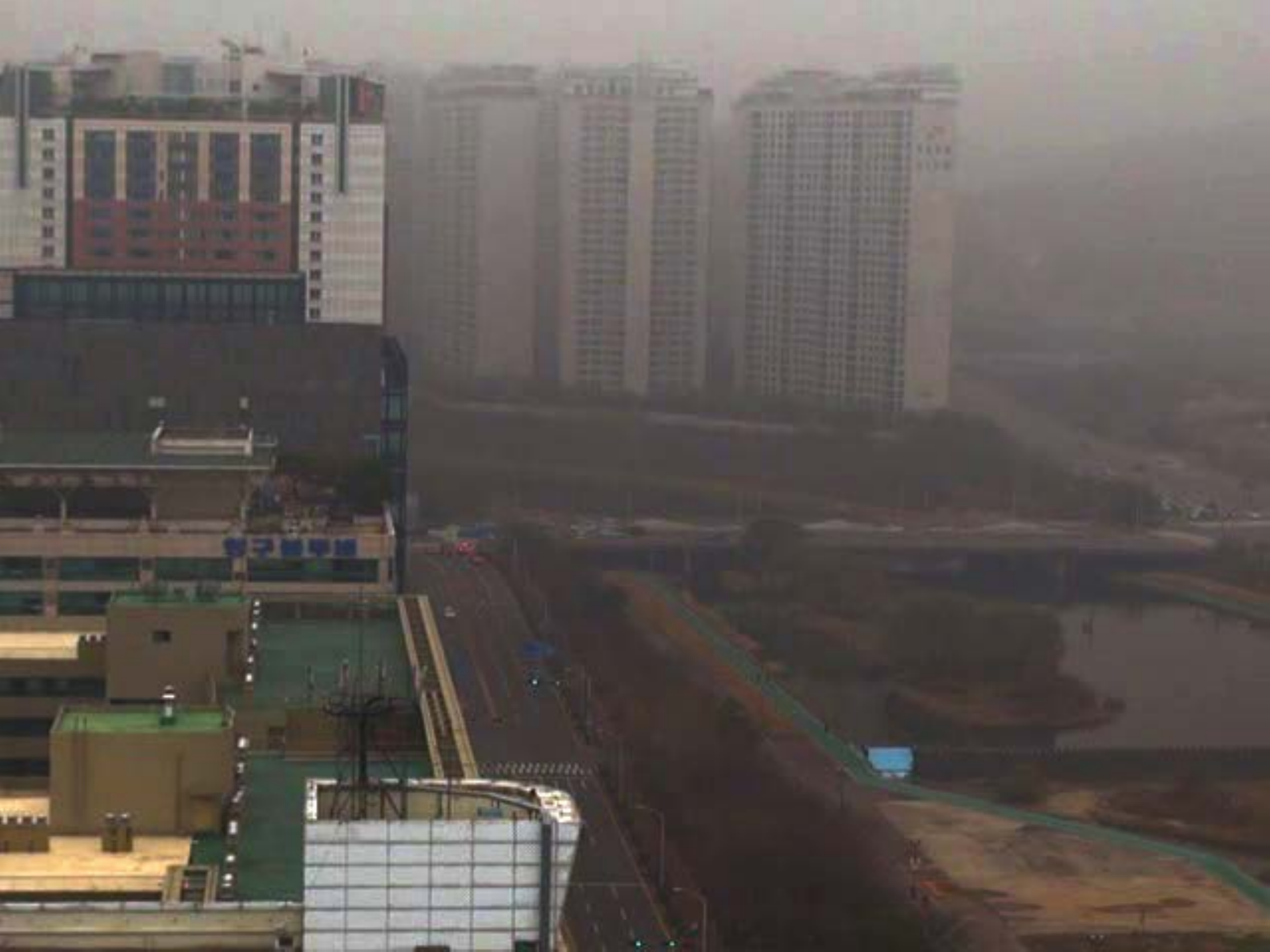} & \hspace{-0.46cm}
			\includegraphics[width = 0.14\linewidth, height = 0.13\linewidth]{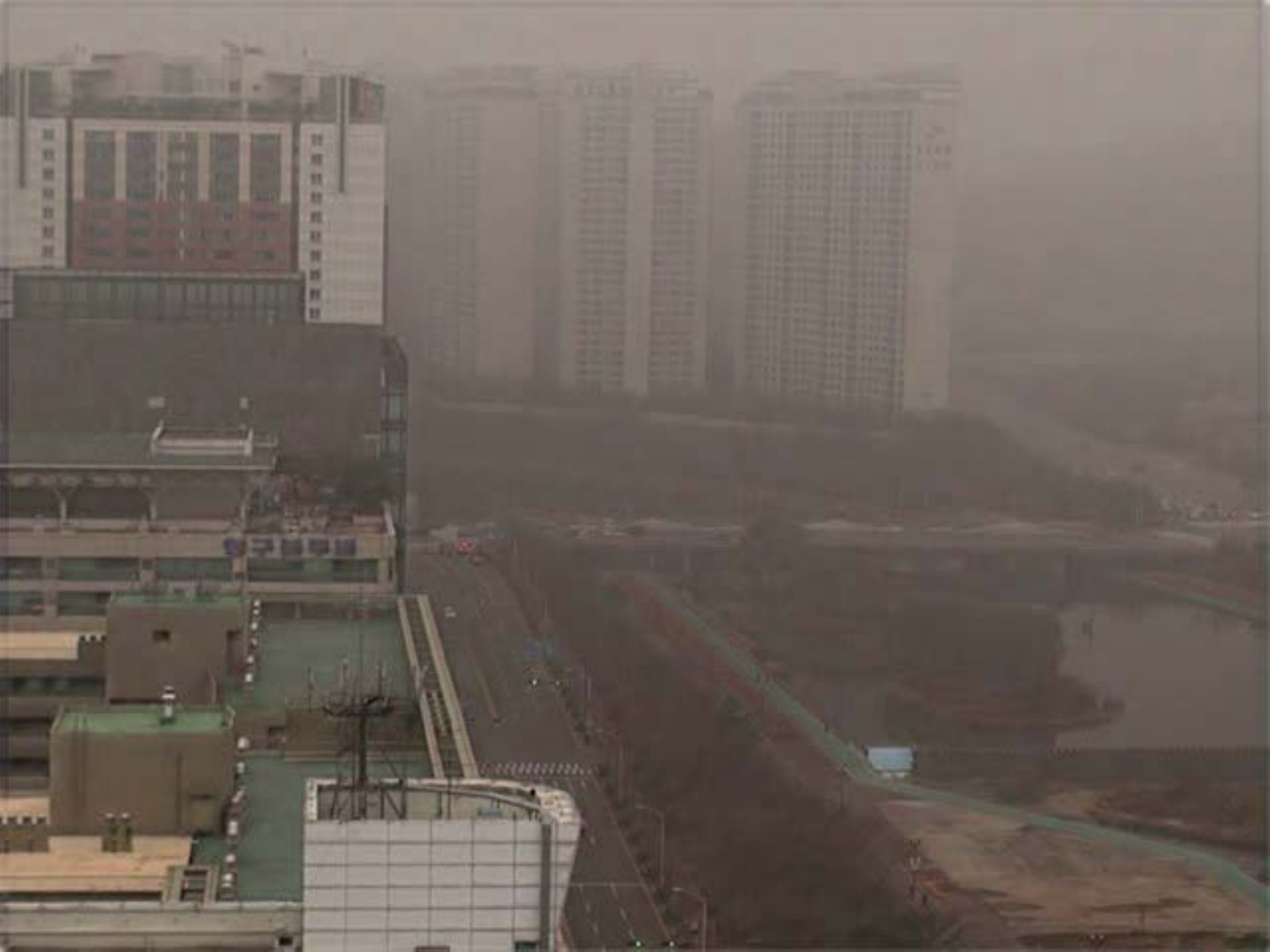} & \hspace{-0.46cm}
			\includegraphics[width = 0.14\linewidth, height = 0.13\linewidth]{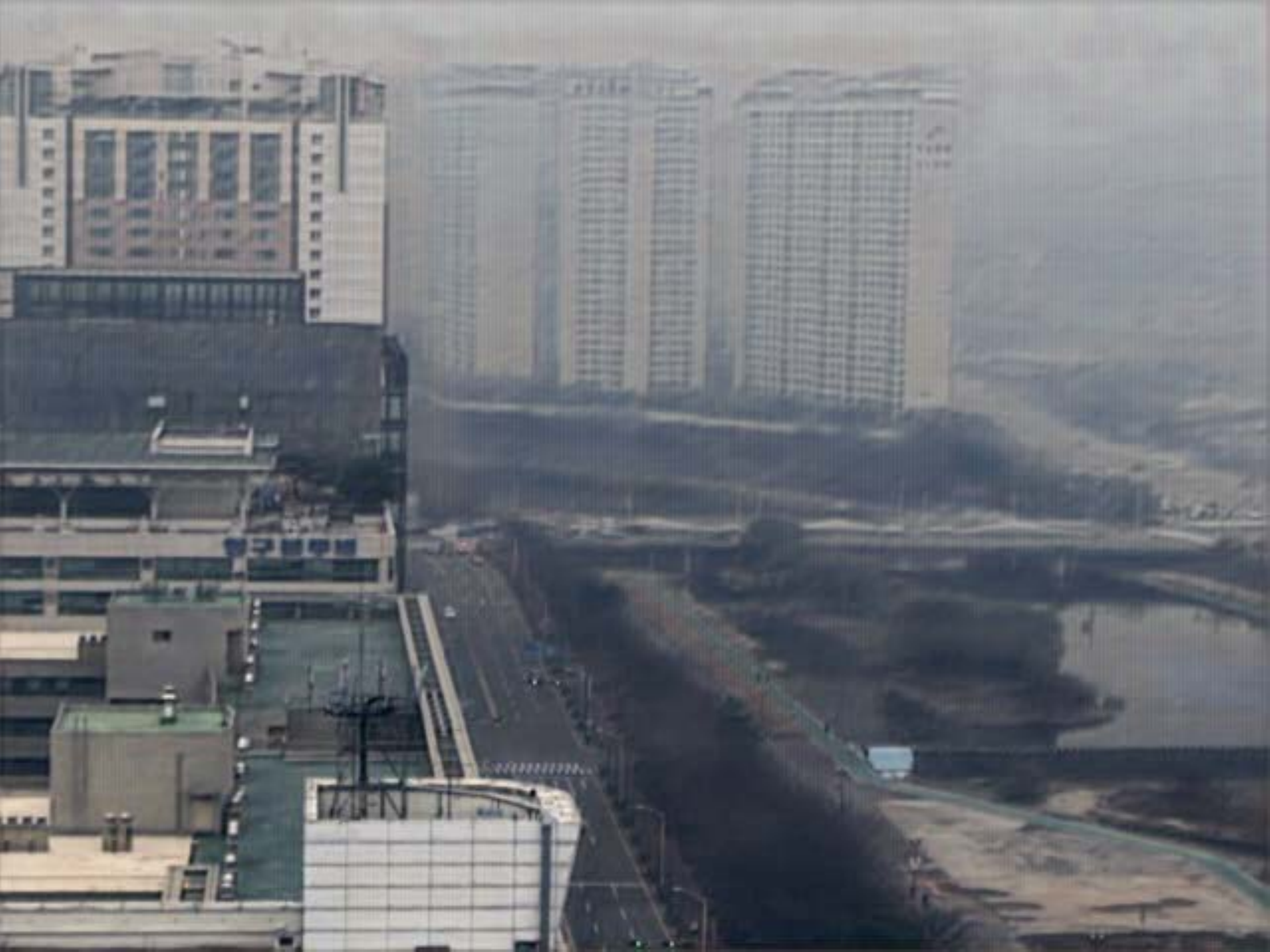} & \hspace{-0.46cm}
			\includegraphics[width = 0.14\linewidth, height = 0.13\linewidth]{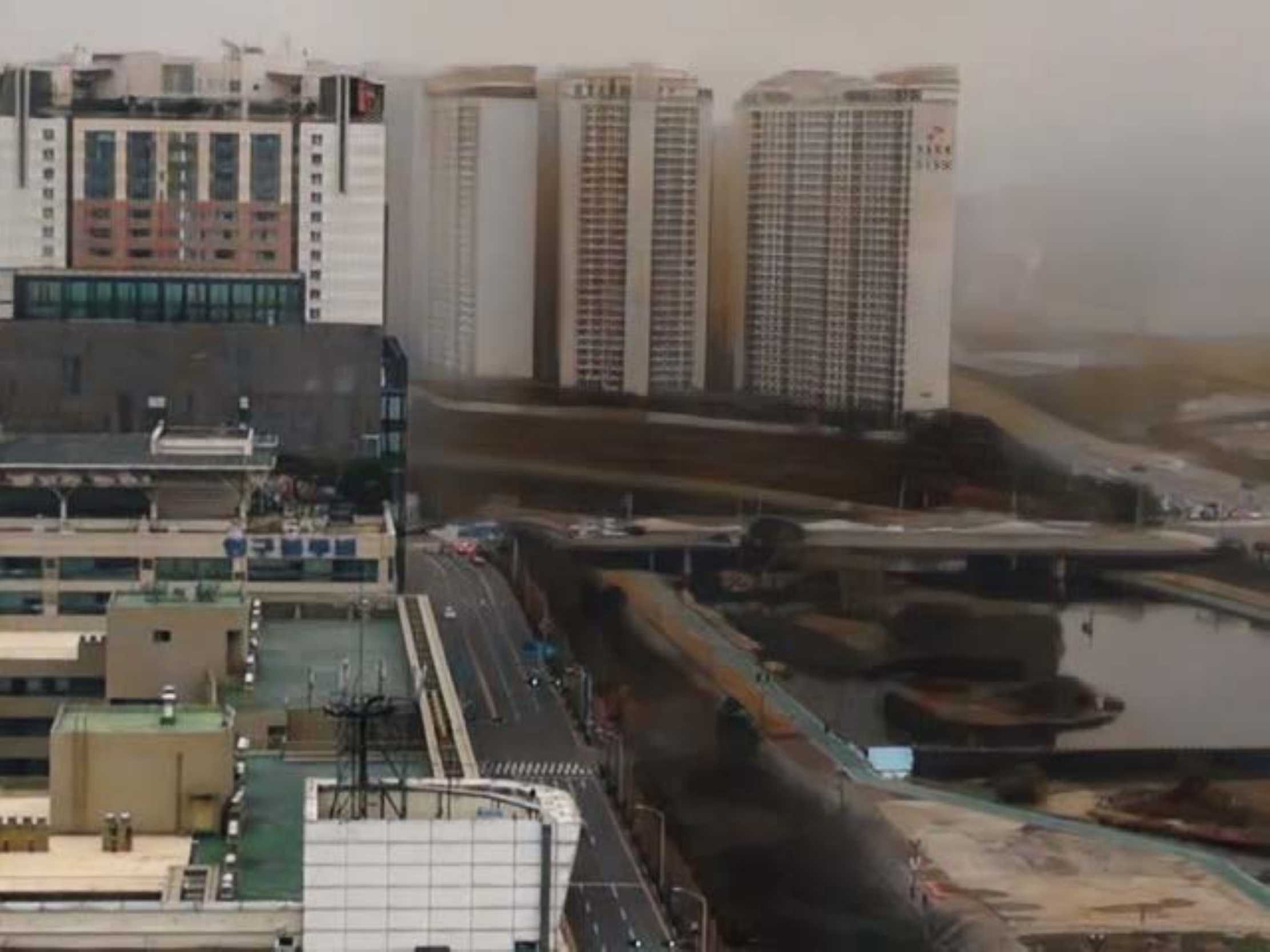}\\
			\includegraphics[width = 0.14\linewidth, height = 0.13\linewidth]{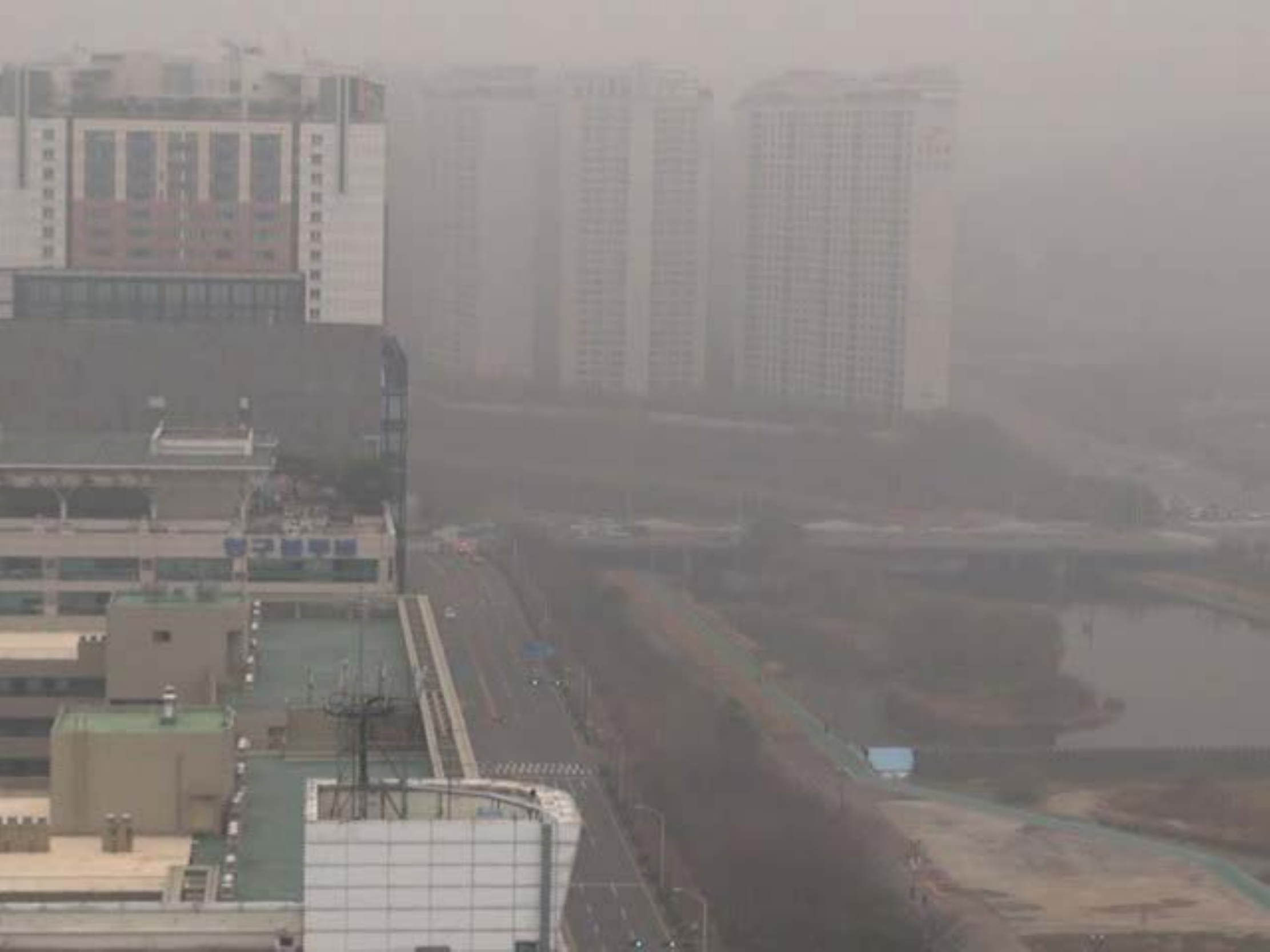}& \hspace{-0.46cm}
			\includegraphics[width = 0.14\linewidth, height = 0.13\linewidth]{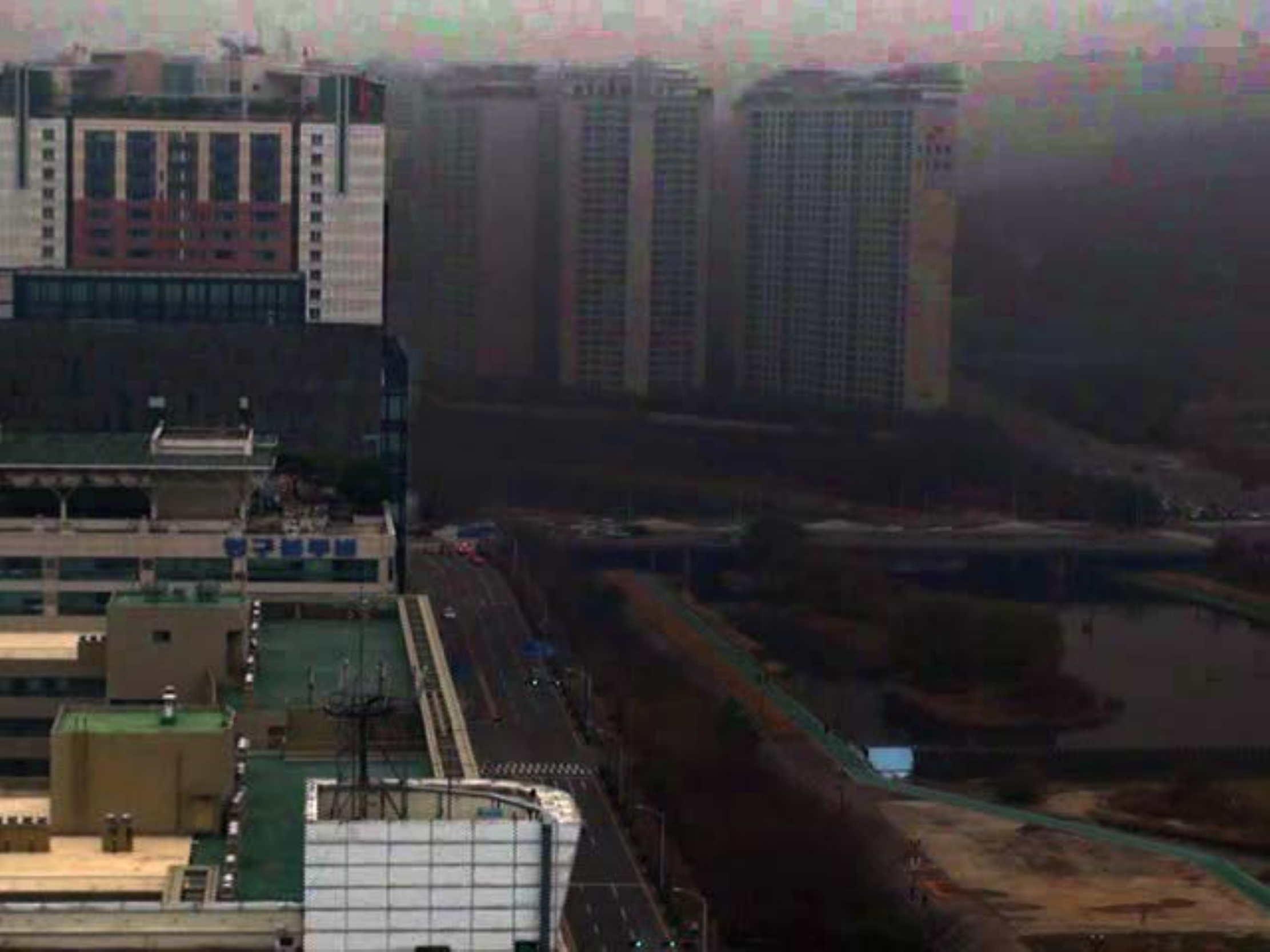} & \hspace{-0.46cm}
			\includegraphics[width = 0.14\linewidth, height = 0.13\linewidth]{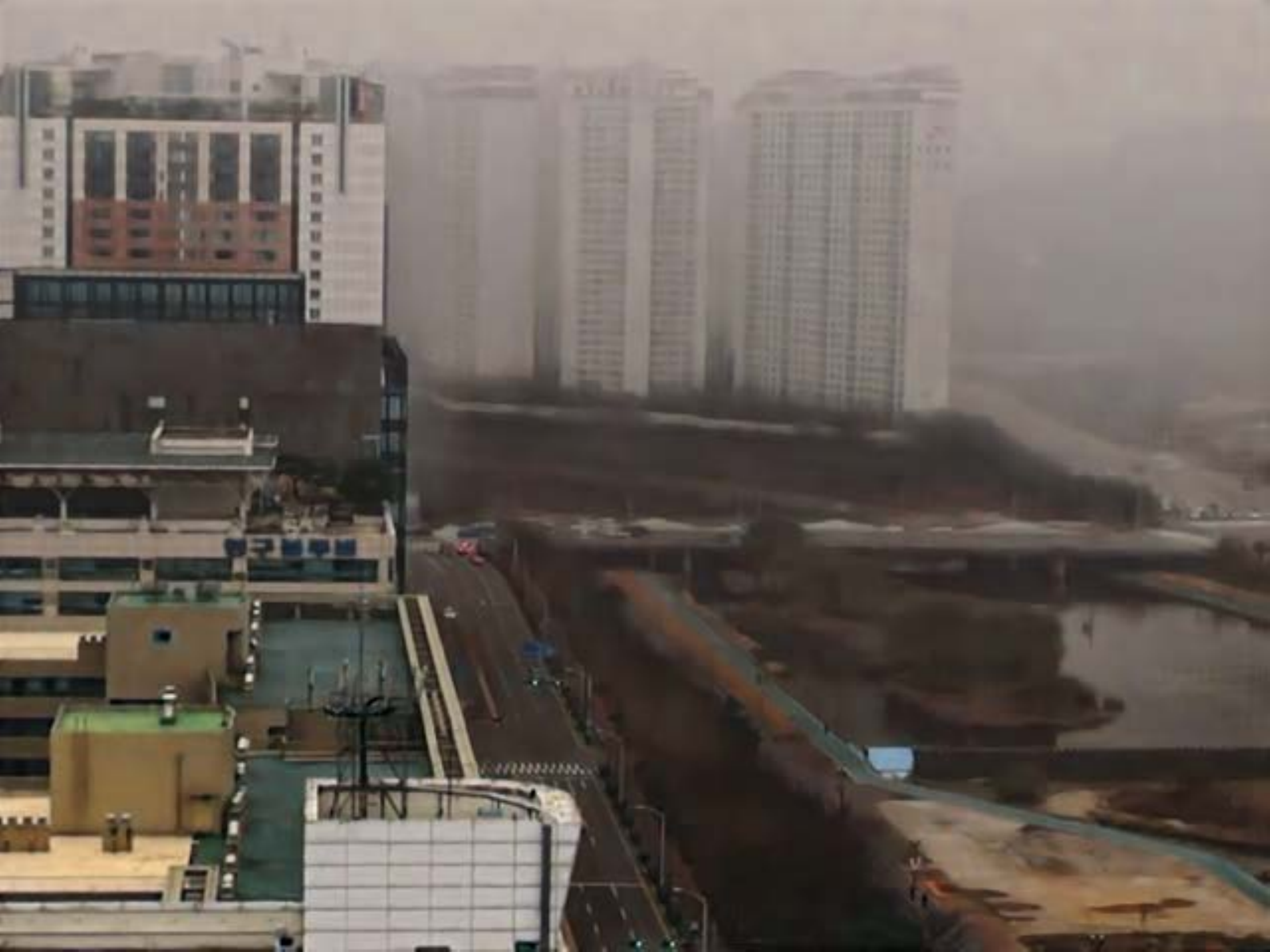} & \hspace{-0.46cm}
			\includegraphics[width = 0.14\linewidth, height = 0.13\linewidth]{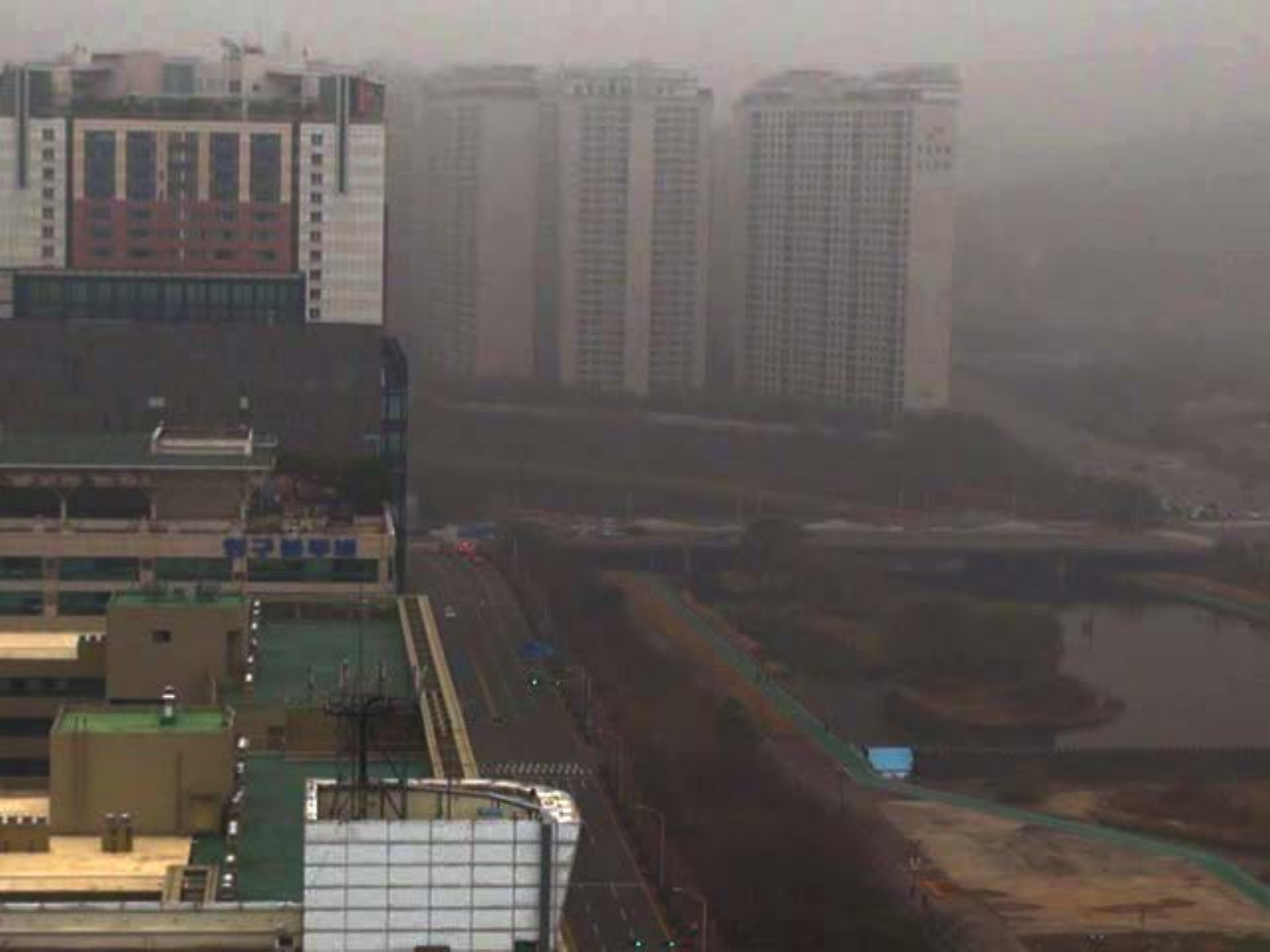} & \hspace{-0.46cm}
			\includegraphics[width = 0.14\linewidth, height = 0.13\linewidth]{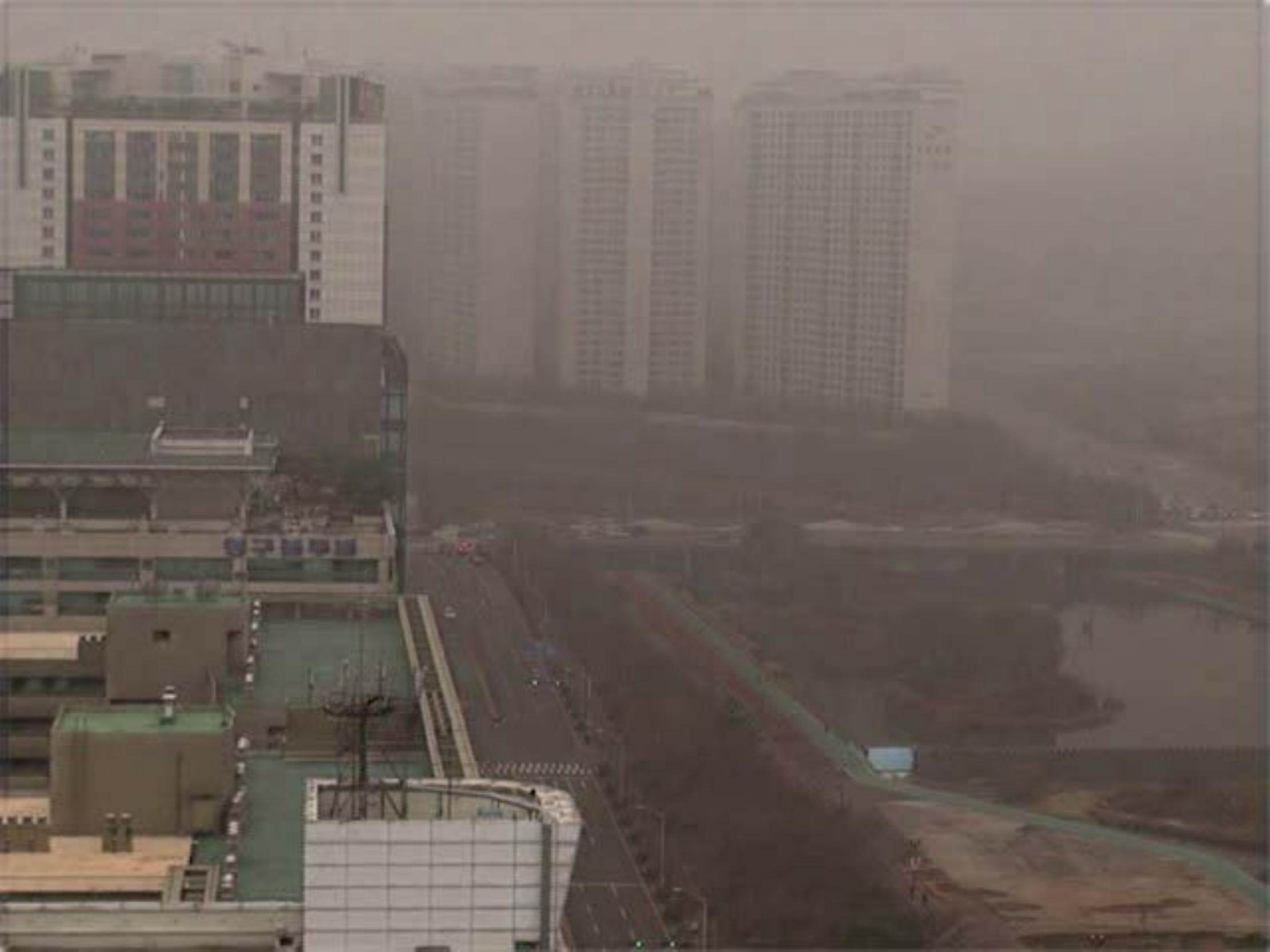} & \hspace{-0.46cm}
			\includegraphics[width = 0.14\linewidth, height = 0.13\linewidth]{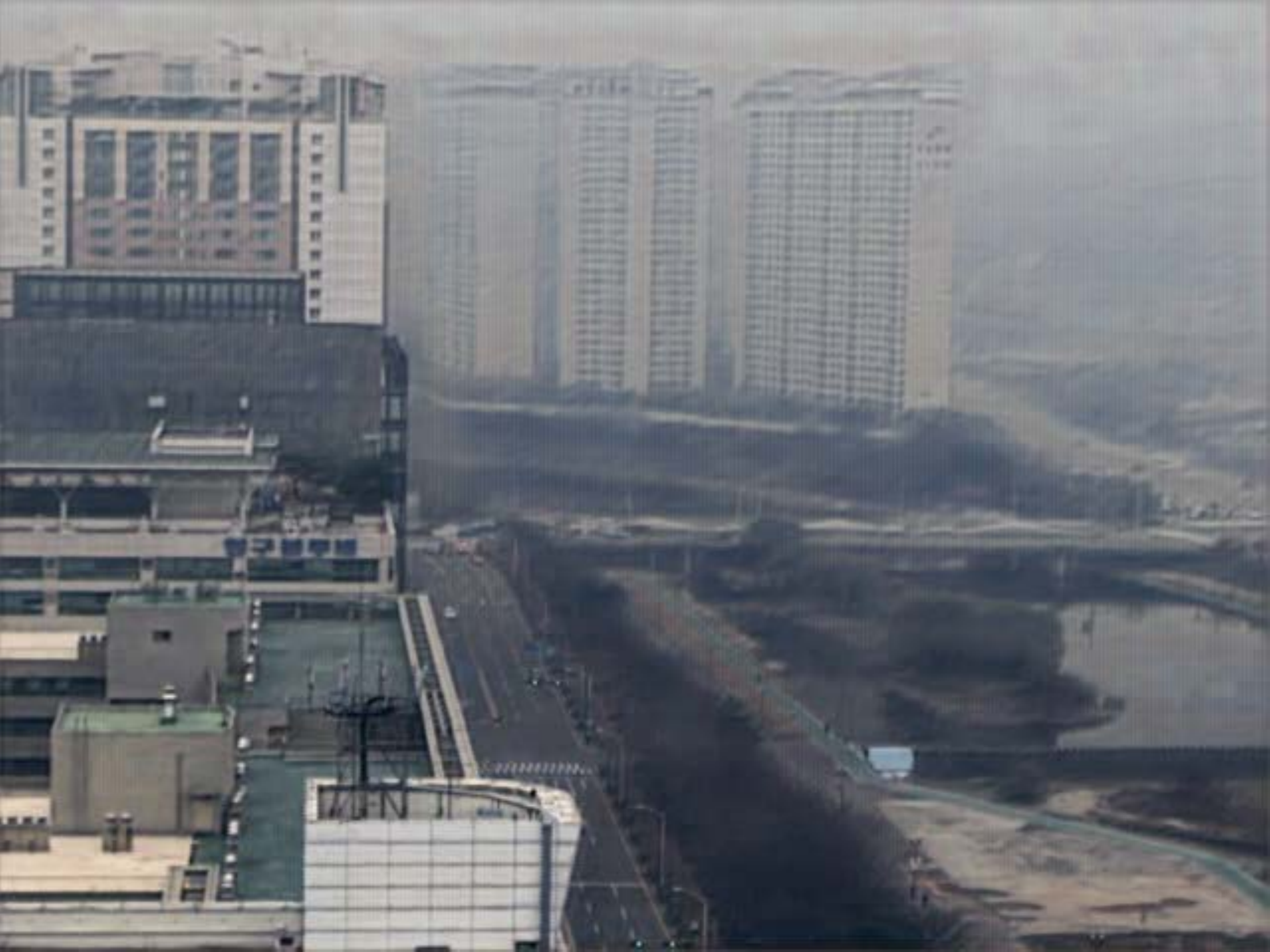} & \hspace{-0.46cm}
			\includegraphics[width = 0.14\linewidth, height = 0.13\linewidth]{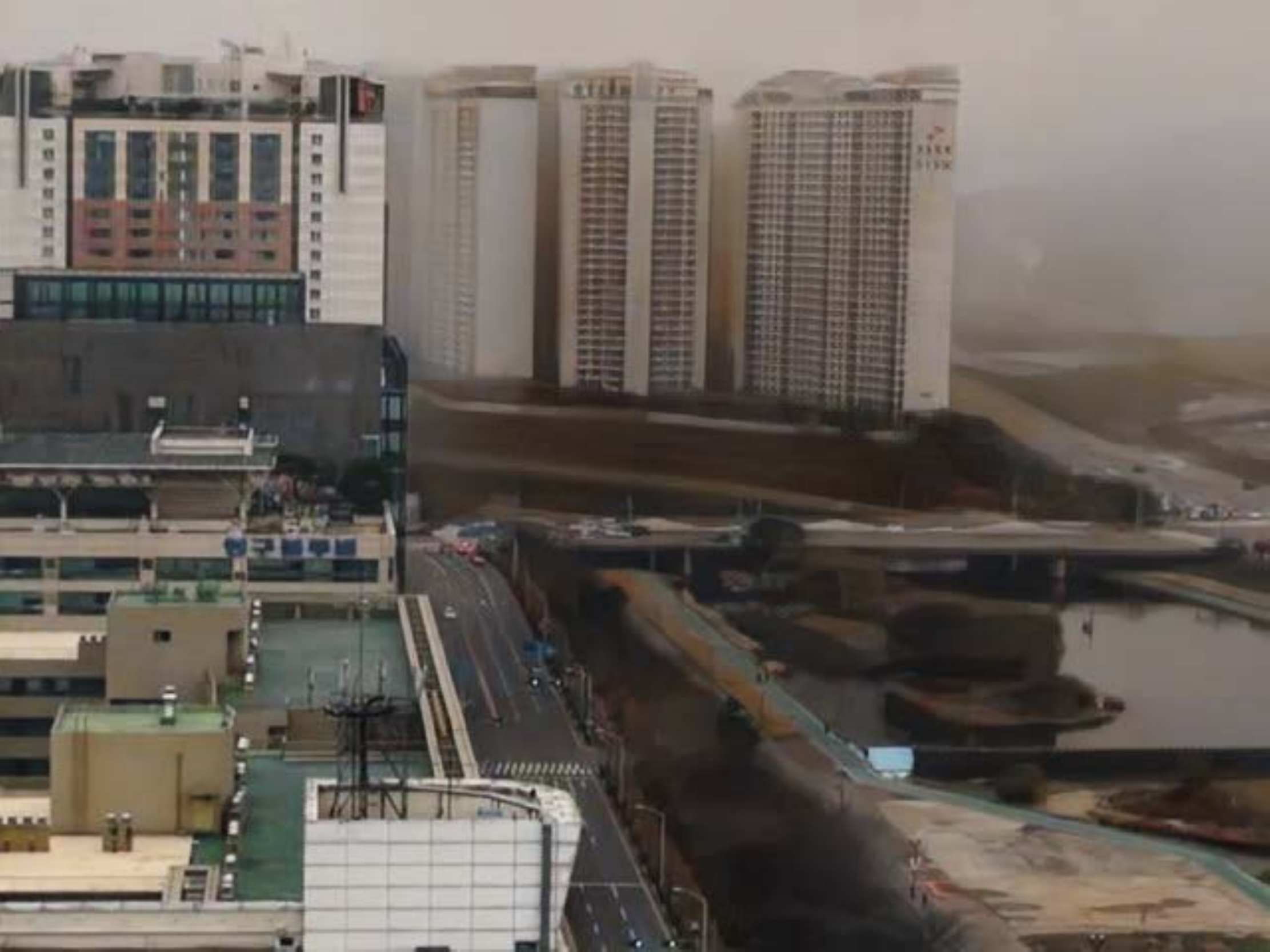}\\
			\includegraphics[width = 0.14\linewidth, height = 0.13\linewidth]{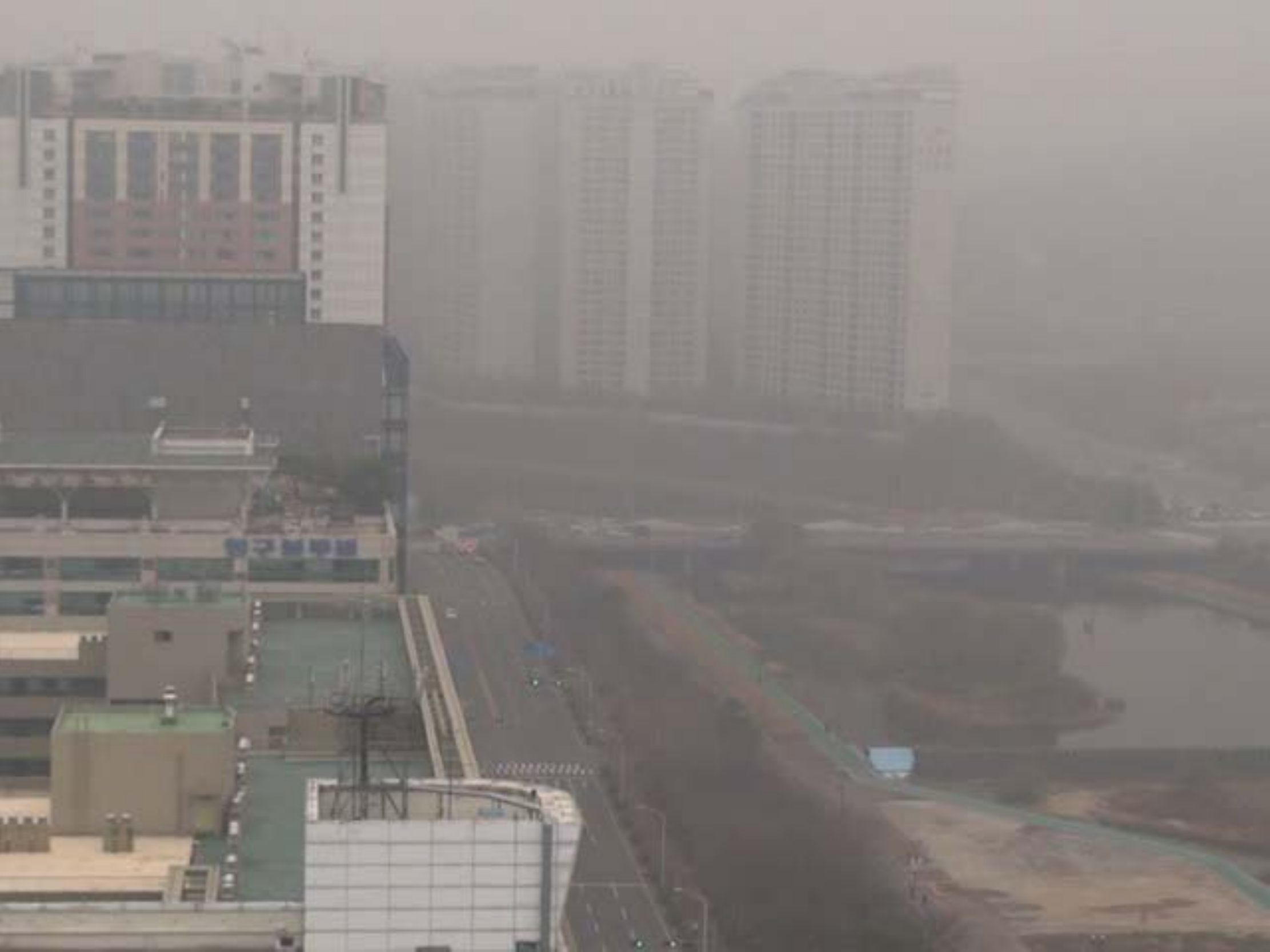}& \hspace{-0.46cm}
			\includegraphics[width = 0.14\linewidth, height = 0.13\linewidth]{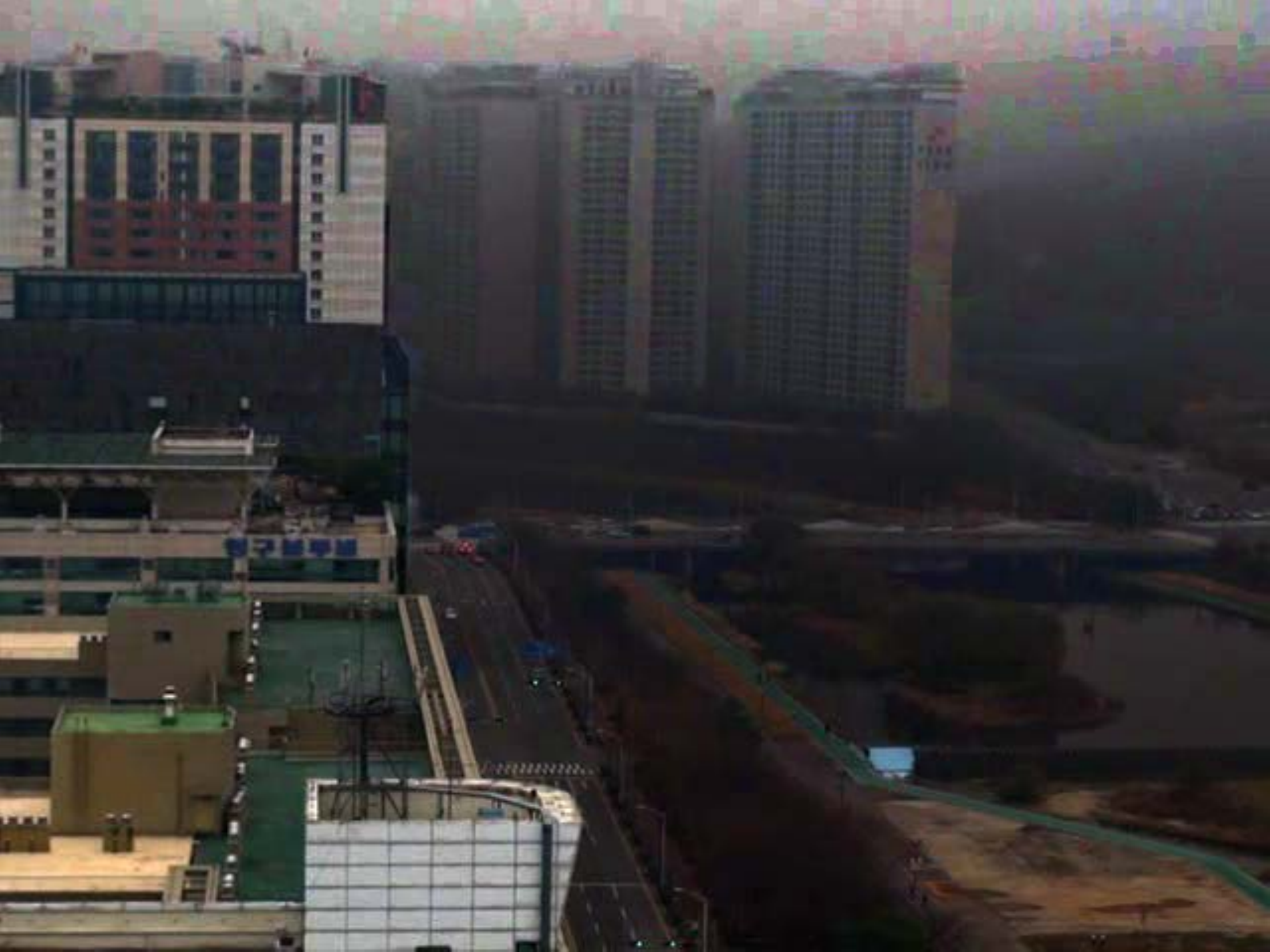} & \hspace{-0.46cm}
			\includegraphics[width = 0.14\linewidth, height = 0.13\linewidth]{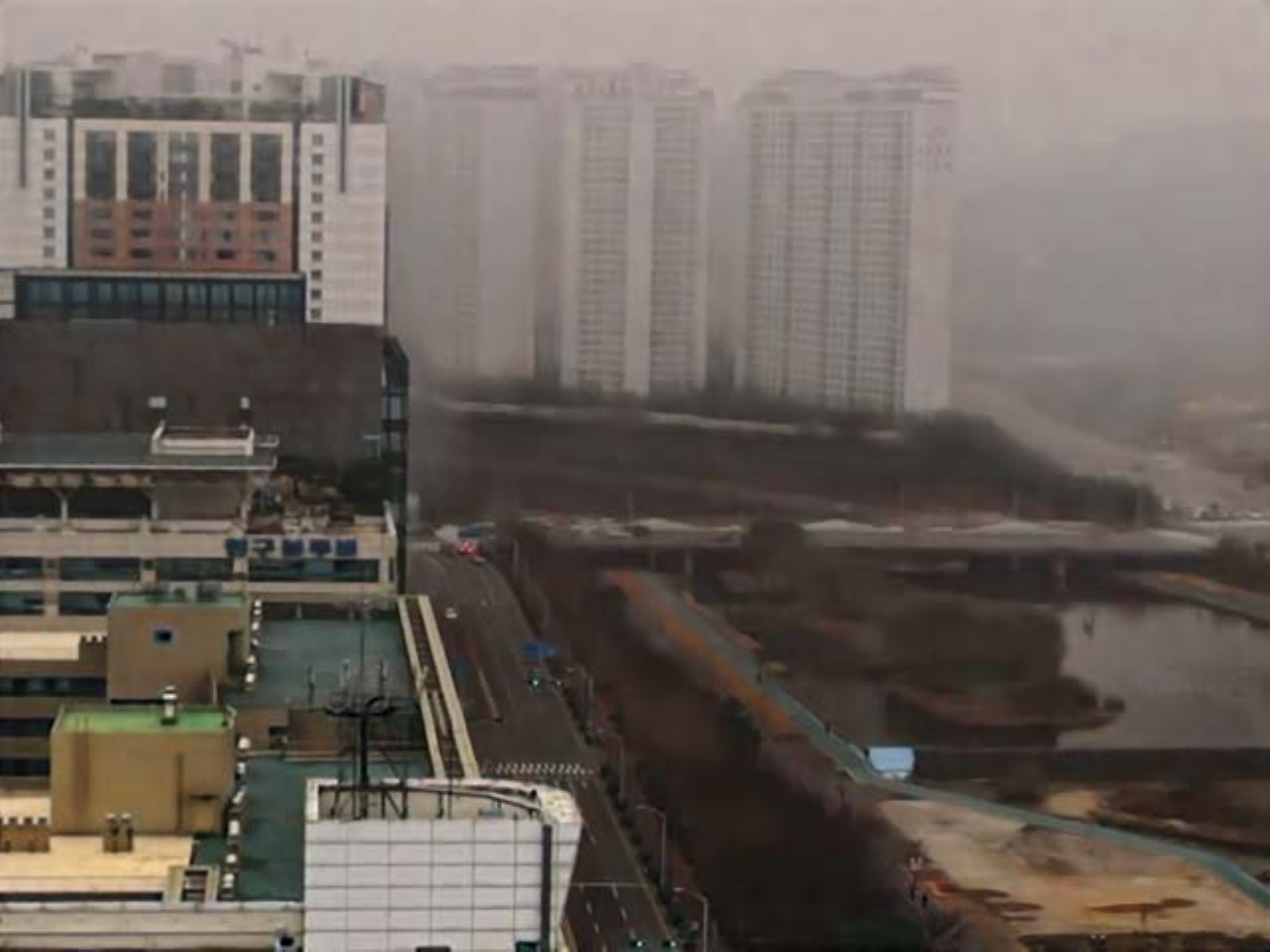} & \hspace{-0.46cm}
			\includegraphics[width = 0.14\linewidth, height = 0.13\linewidth]{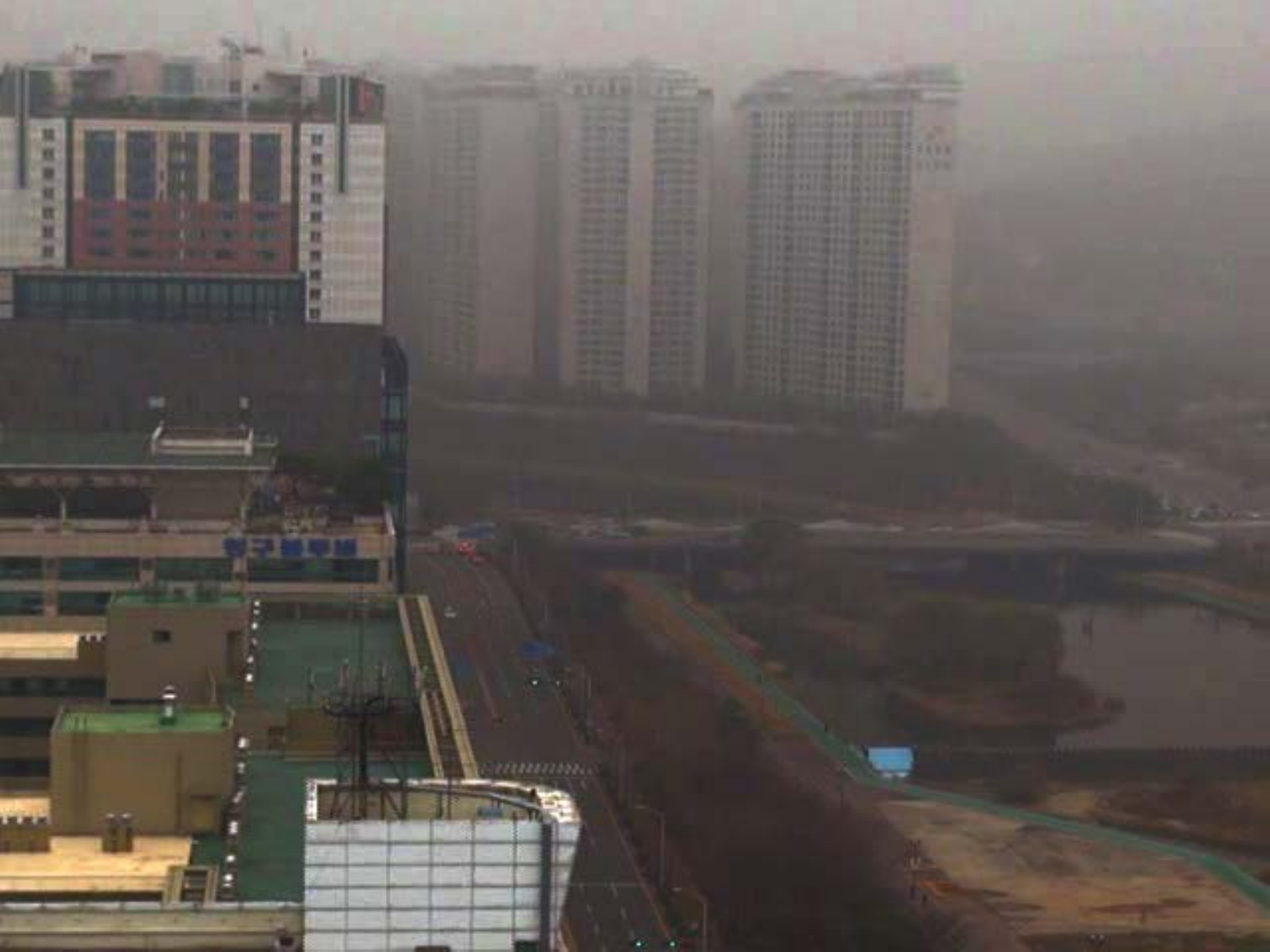} & \hspace{-0.46cm}
			\includegraphics[width = 0.14\linewidth, height = 0.13\linewidth]{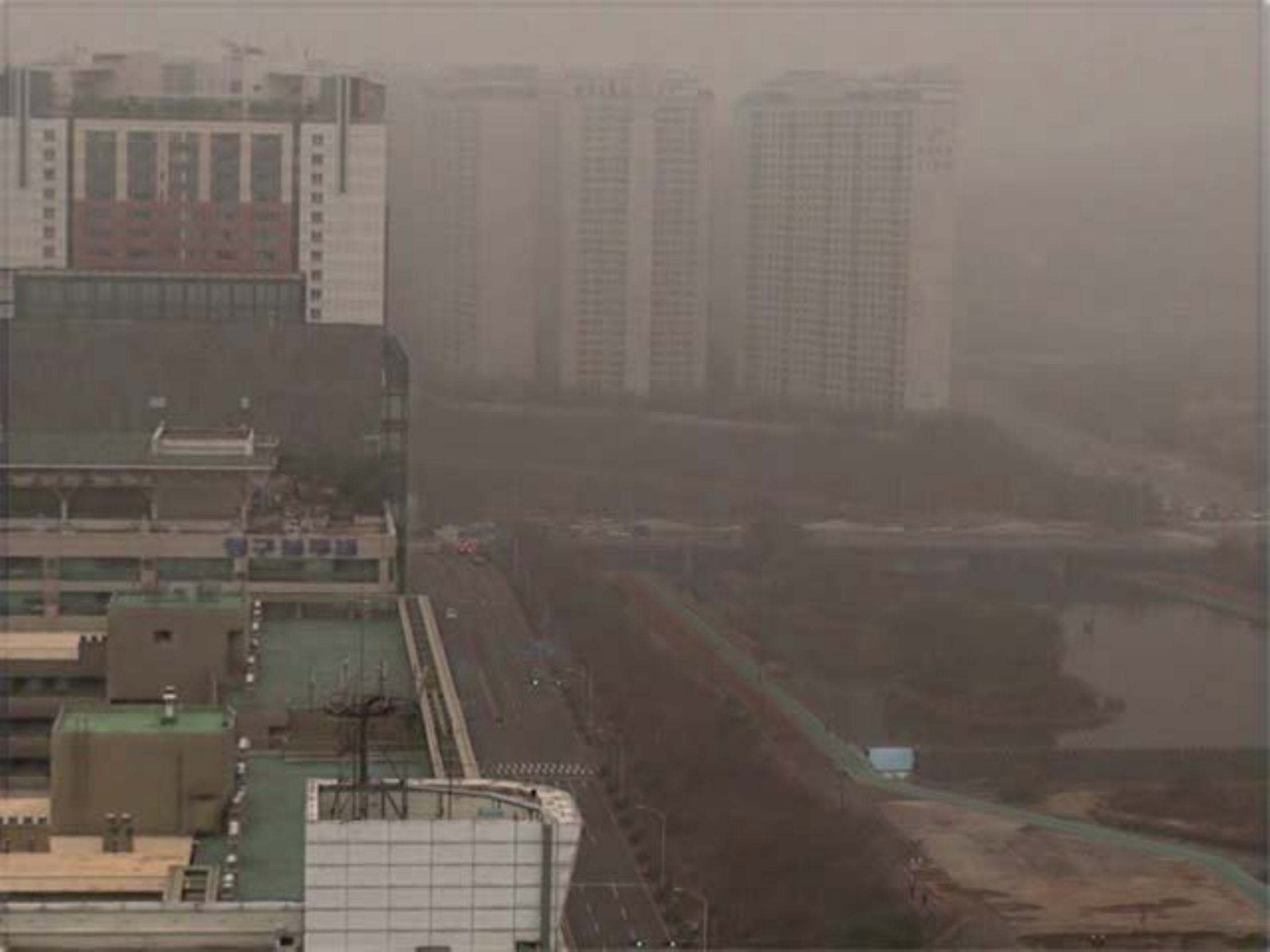} & \hspace{-0.46cm}
			\includegraphics[width = 0.14\linewidth, height = 0.13\linewidth]{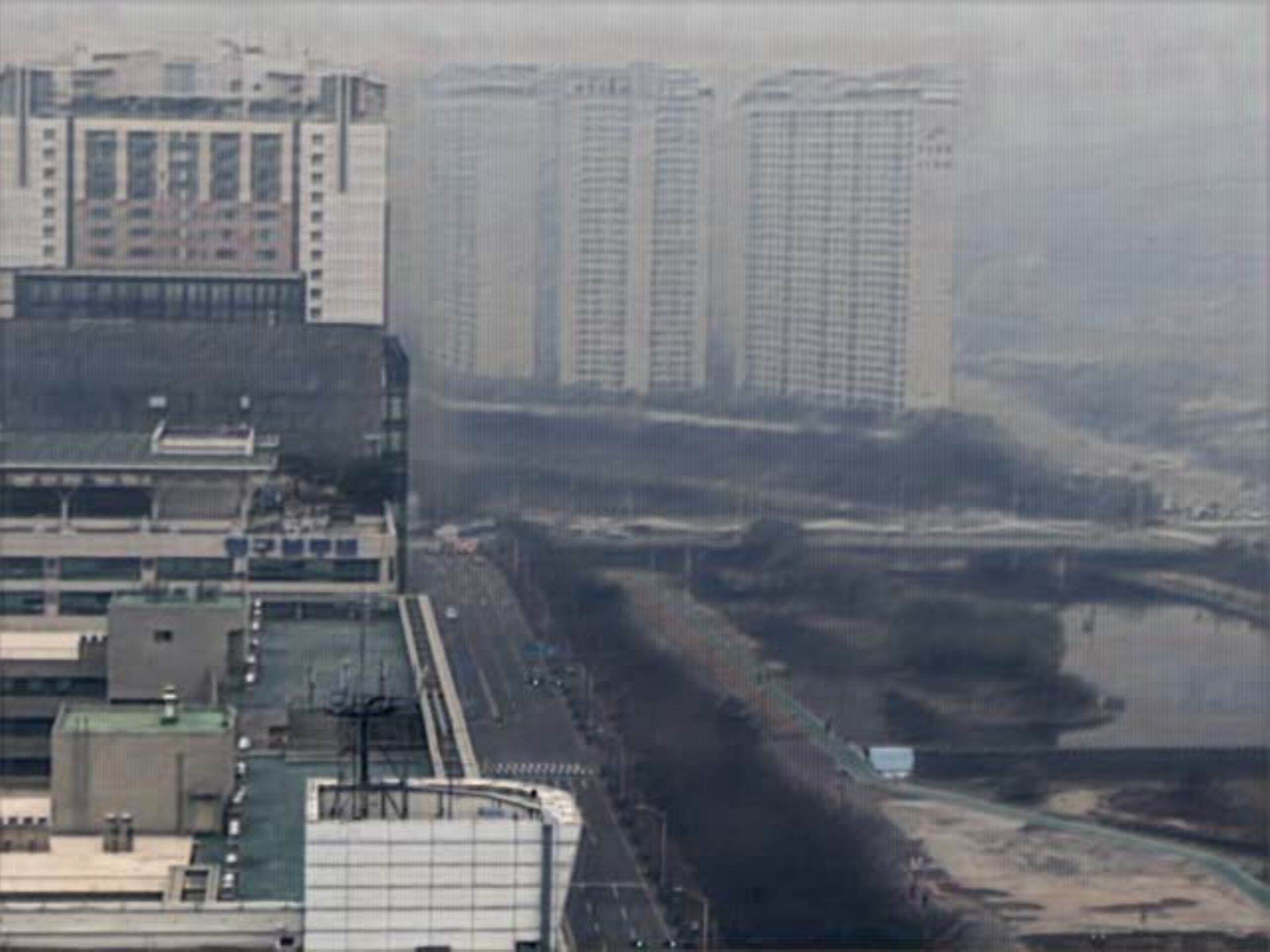} & \hspace{-0.46cm}
			\includegraphics[width = 0.14\linewidth, height = 0.13\linewidth]{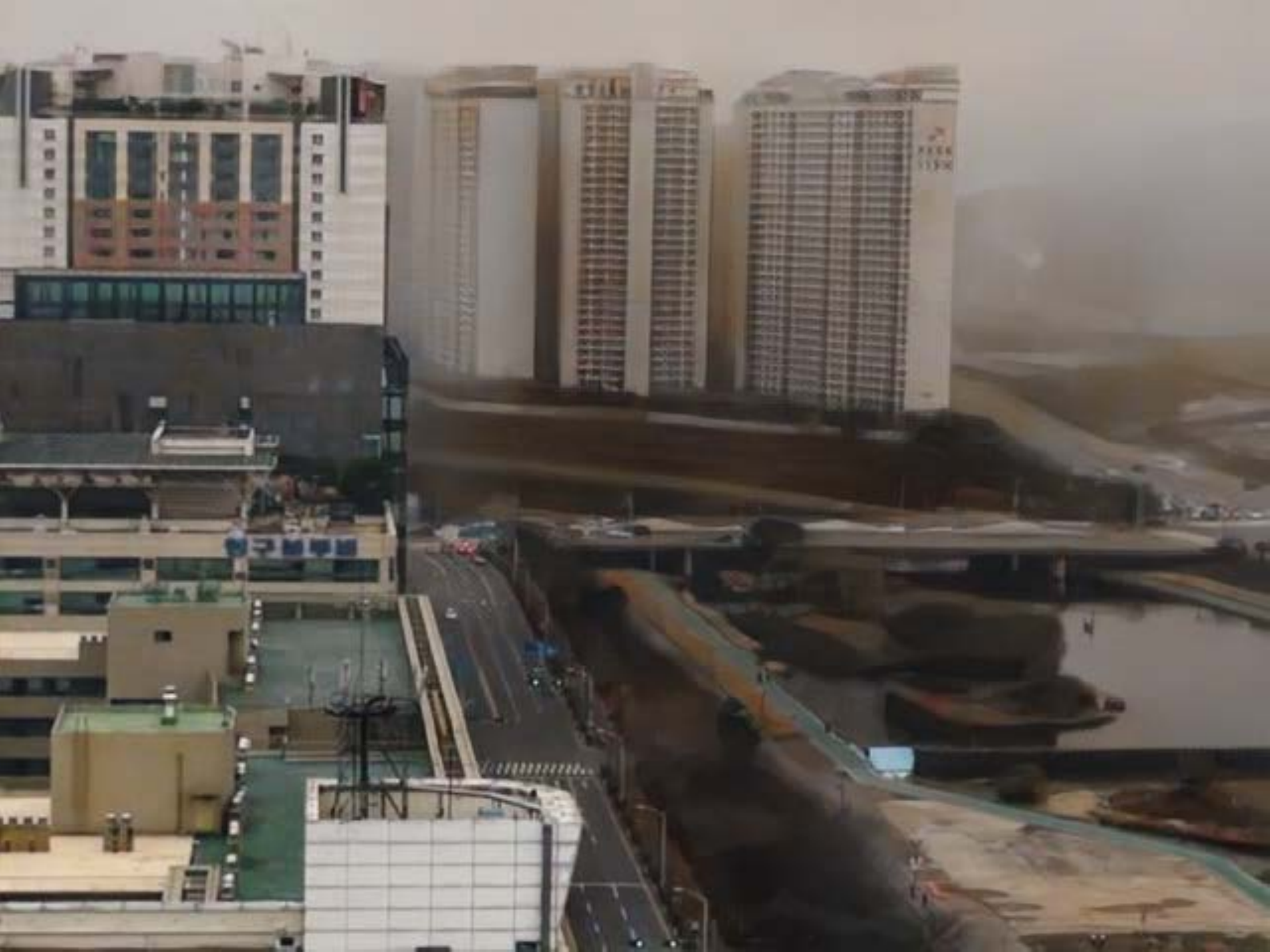}\\			
			(a) Original  & \hspace{-0.46cm} (b) DCP \cite{DBLP:journals/pami/He0T11} & \hspace{-0.46cm} (c) FFA \cite{DBLP:conf/aaai/QinWBXJ20} & \hspace{-0.46cm} (d) STMRF \cite{DBLP:conf/pcm/CaiXT16}& \hspace{-0.46cm} (e) EDVNet \cite{DBLP:conf/aaai/LiPWXF18}& \hspace{-0.46cm}  (f) FastDVD \cite{DBLP:conf/cvpr/TassanoDV20}& \hspace{-0.46cm}  (g) Ours\\
		\end{tabular}
	\end{center}
	%\vspace{-0.5cm}
	\caption{Examples from consecutive adjacent frames and corresponding dehazed results by several state-of-the-art methods.}
	\label{fig:frame2}
\end{figure*}

\subsection{Evaluations on Real Videos}
%%%%%%%%%%%%%%%%%%
\begin{figure*}[!htp]
	\footnotesize
	\begin{center}
		\begin{tabular}{ccccc}
			%\vspace{-0.2cm}
			\includegraphics[width = 0.24\linewidth, height = 0.16\linewidth]{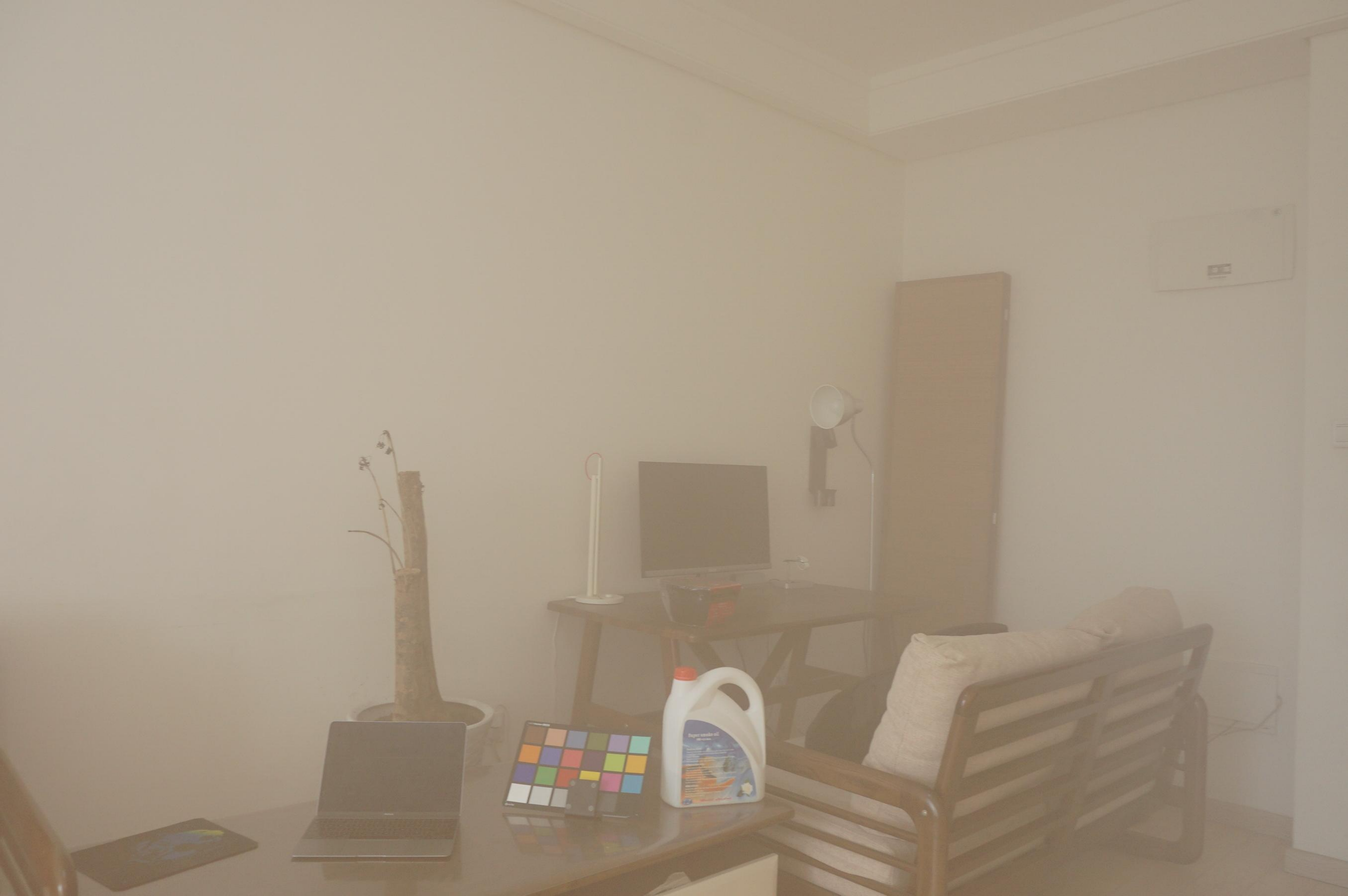}& \hspace{-0.46cm}
			\includegraphics[width = 0.24\linewidth, height = 0.16\linewidth]{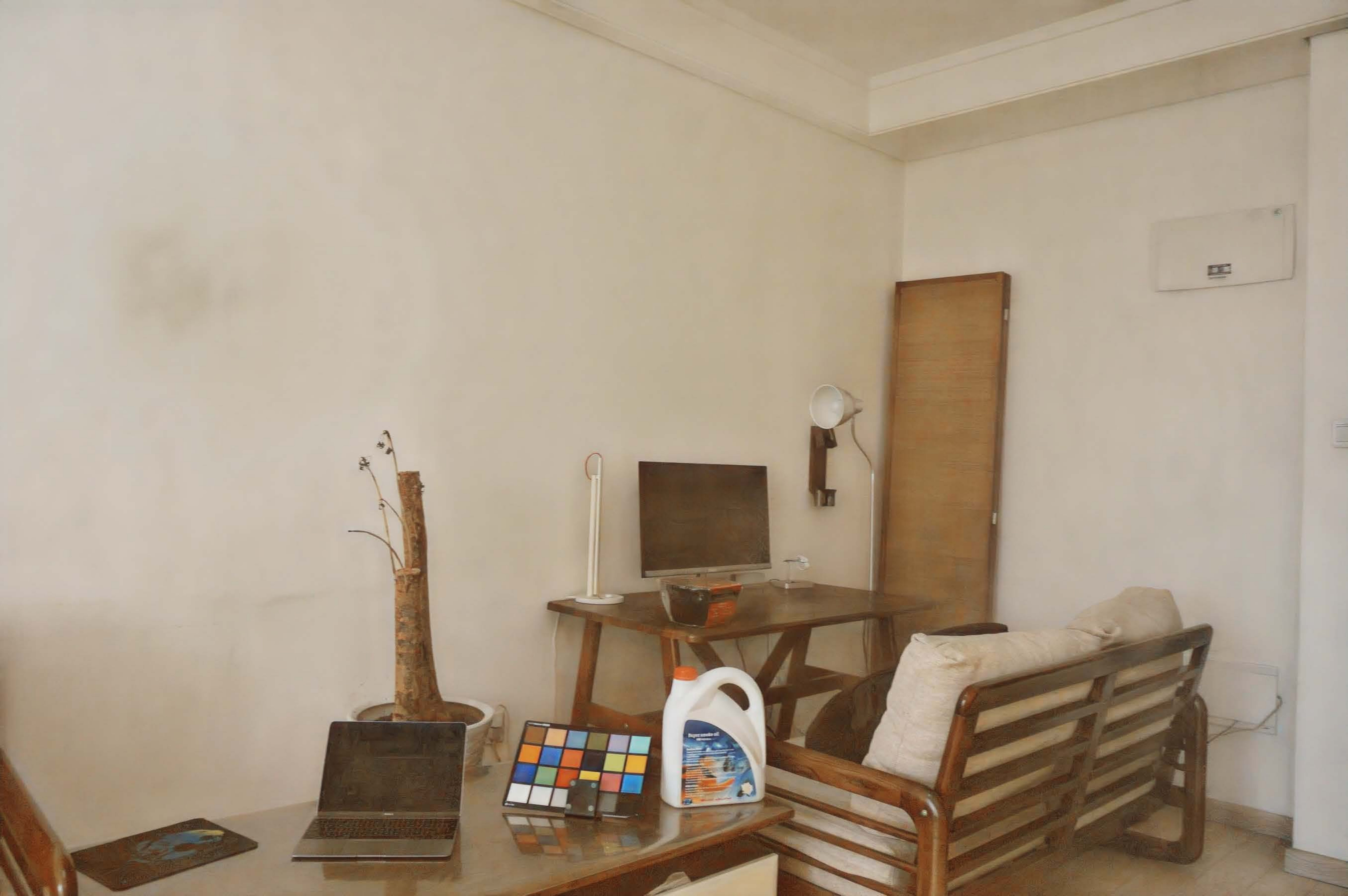} & \hspace{-0.46cm}
			\includegraphics[width = 0.24\linewidth, height = 0.16\linewidth]{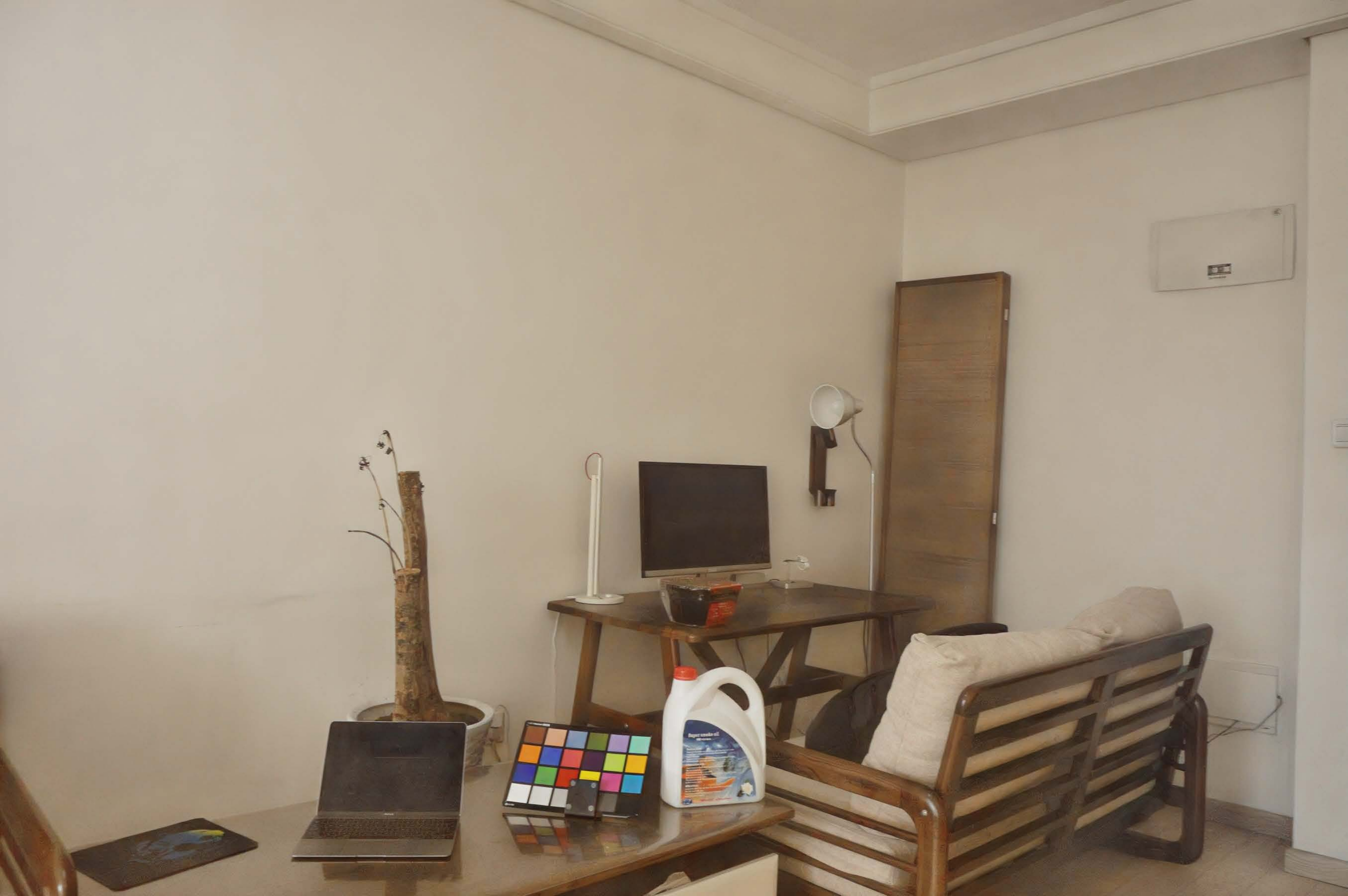} & \hspace{-0.46cm}
			\includegraphics[width = 0.24\linewidth, height = 0.16\linewidth]{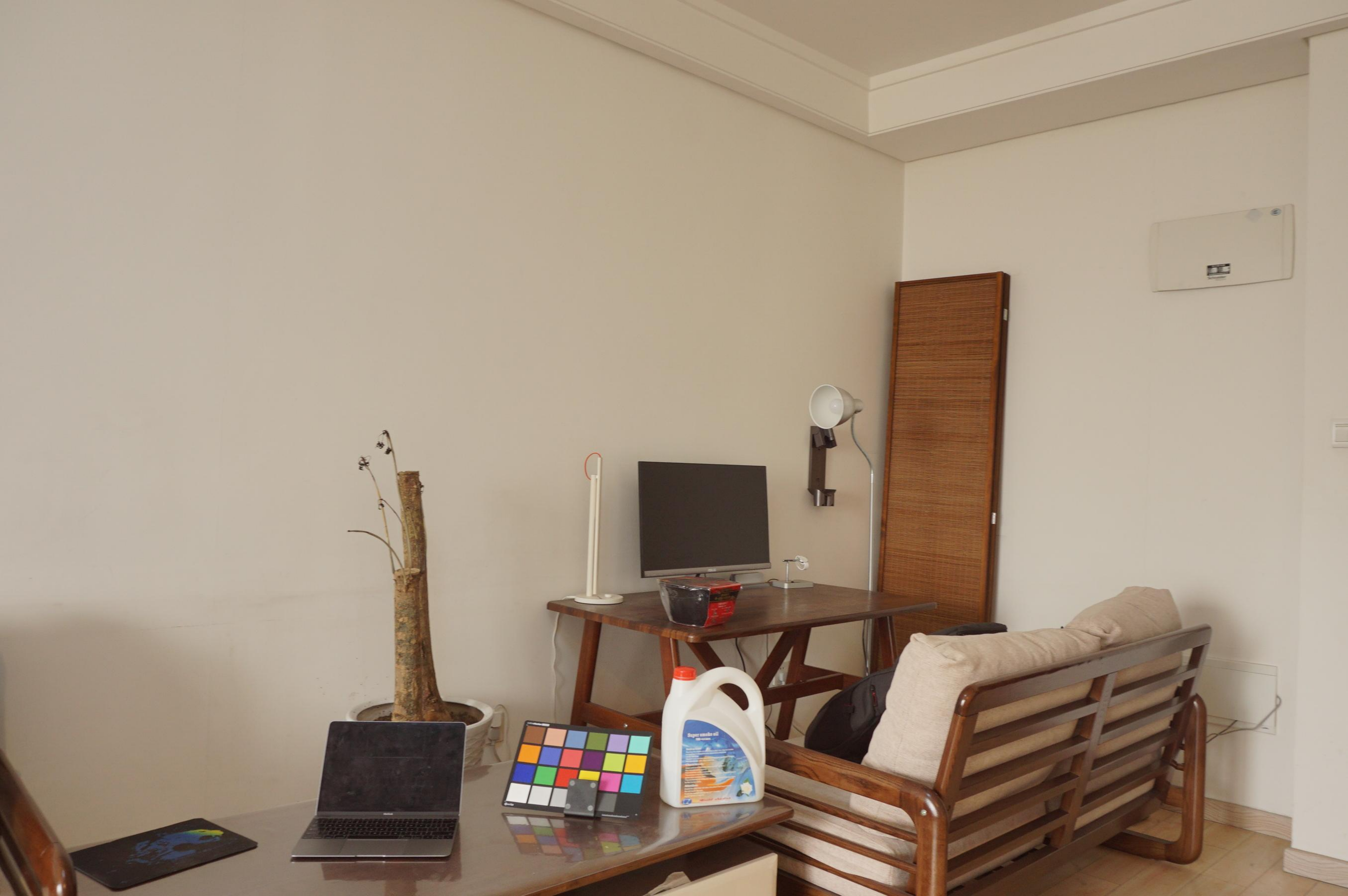}\\
			18.99/0.911  & \hspace{-0.46cm} 24.22/0.949  & \hspace{-0.46cm} 24.39/0.956   & \hspace{-0.46cm}  $+\infty$/1 \\
			\includegraphics[width = 0.24\linewidth, height = 0.16\linewidth]{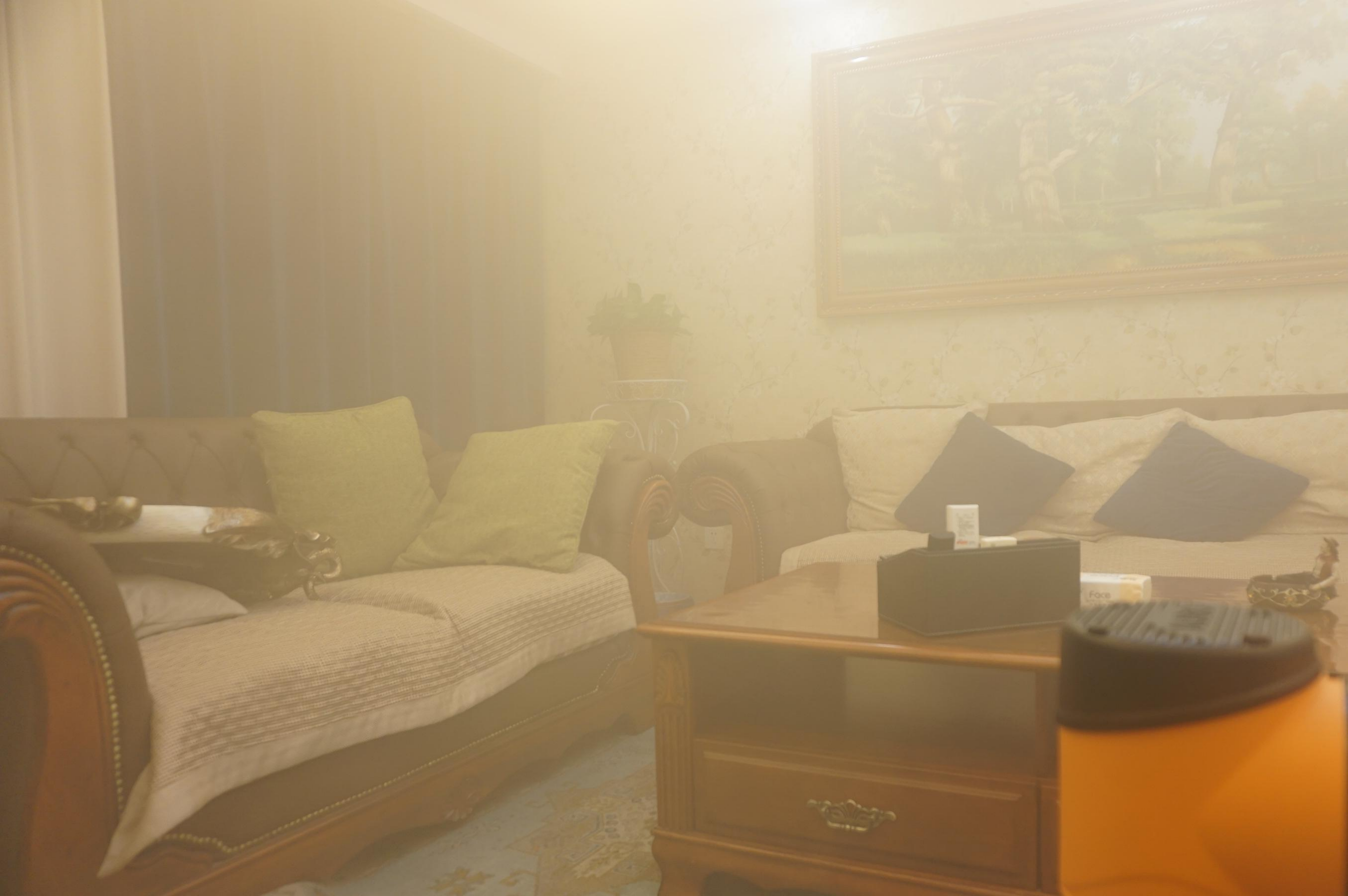}& \hspace{-0.46cm}
			\includegraphics[width = 0.24\linewidth, height = 0.16\linewidth]{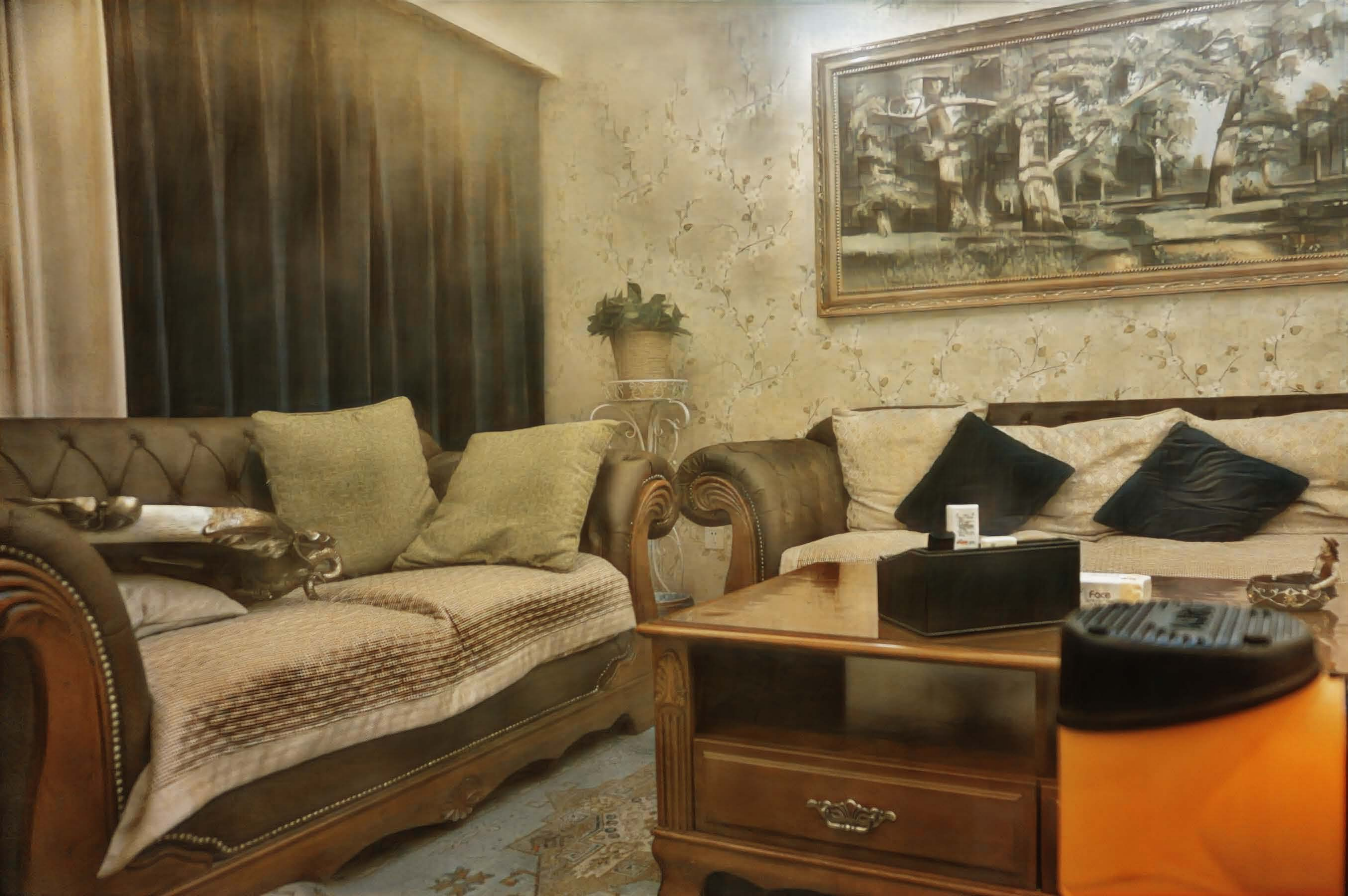} & \hspace{-0.46cm}
			\includegraphics[width = 0.24\linewidth, height = 0.16\linewidth]{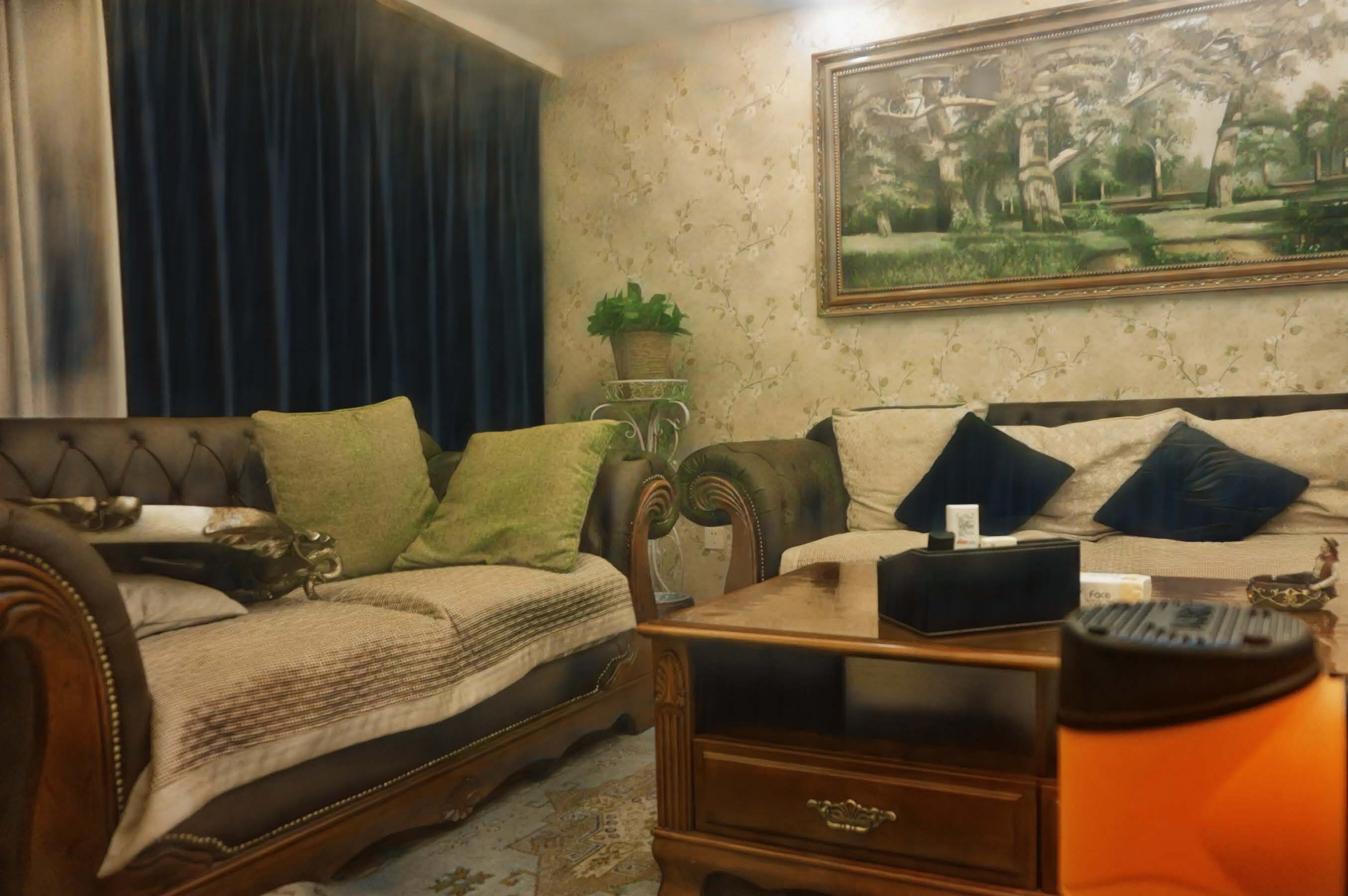} & \hspace{-0.46cm}
			\includegraphics[width = 0.24\linewidth, height = 0.16\linewidth]{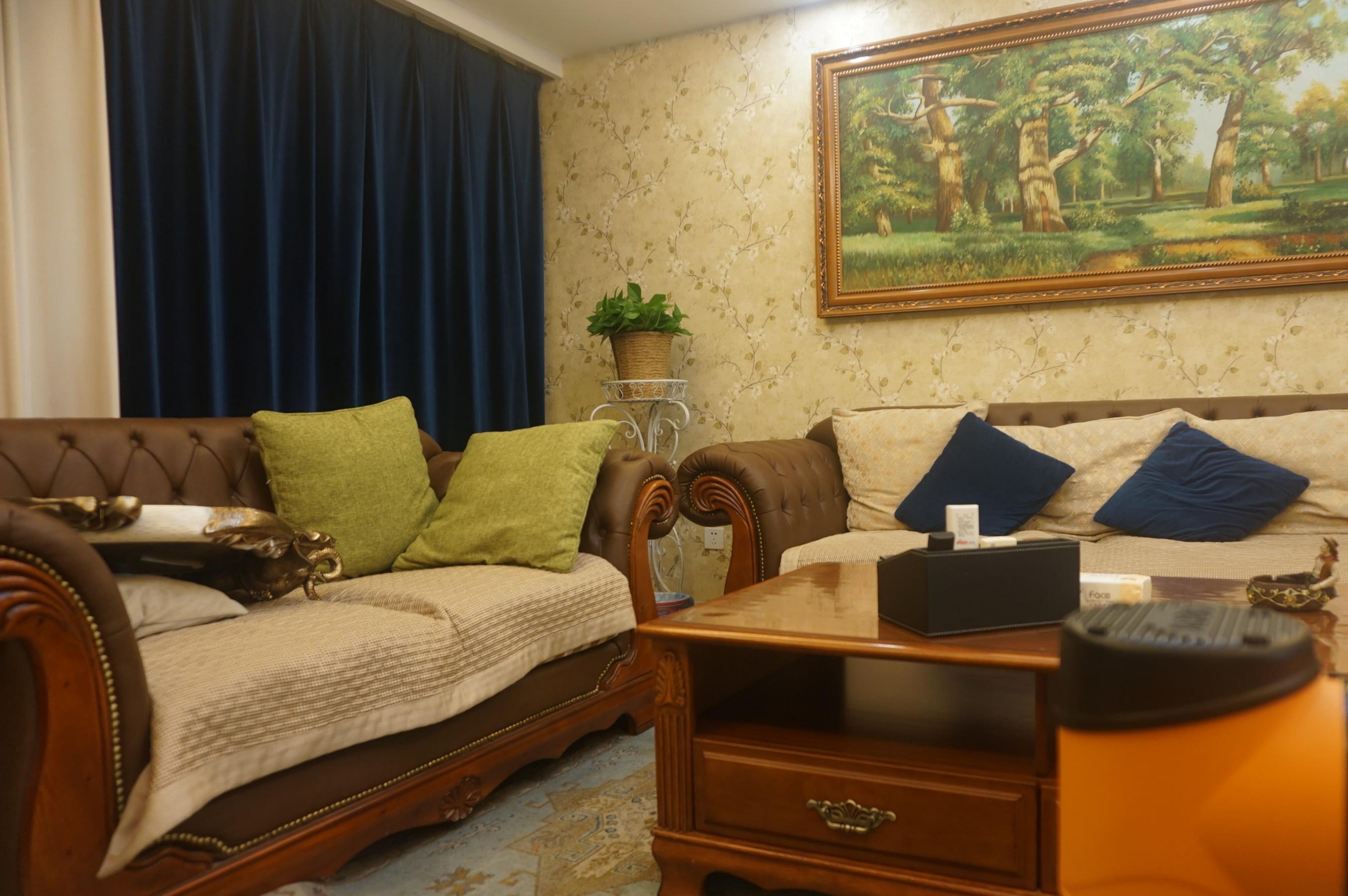}\\
			11.42/0.648  & \hspace{-0.46cm} 20.82/0.758  & \hspace{-0.46cm} 21.35/0.884   &\hspace{-0.46cm}  $+\infty$/1 \\	 	
			(a) Original  & \hspace{-0.46cm} (b)  & \hspace{-0.46cm} (c)  & \hspace{-0.46cm}  (d)  Ground Truth \\
		\end{tabular}
	\end{center}
	%\vspace{-0.5cm}
	\caption{Examples from the REVIDE test dataset and corresponding dehazed frames by different baseline methods. PSNR/SSIM values are below each frame.}
	\label{fig:ablation}
\end{figure*}
To evaluate the effect of the proposed method, we compare our video dehazing method with the state-of-the-art image and video dehazing methods including: DCP~\cite{DBLP:journals/pami/He0T11}, FFA~\cite{DBLP:conf/aaai/QinWBXJ20}, GridDehazeNet~\cite{DBLP:conf/iccv/LiuMSC19}, RIVD~\cite{DBLP:conf/eccv/ChenDW16}, STMRF~\cite{DBLP:conf/pcm/CaiXT16}, EVDNet~\cite{DBLP:conf/aaai/LiPWXF18}.
We retrain the FastDVDnet~\cite{DBLP:conf/cvpr/TassanoDV20}  on the REVIDE training dataset and replace the noise map with the haze map  and set it as a baseline experiment.
The FFA~\cite{DBLP:conf/aaai/QinWBXJ20} and GridDehazeNet~\cite{DBLP:conf/iccv/LiuMSC19} are single image dehazing methods which are retrained on the REVIDE training dataset.
The EUVD~\cite{DBLP:conf/aaai/LiPWXF18} is a video dehazing method which is retrained on the REVIDE training dataset.
We use the same patch size for each method retrained on the haze video dataset in the training process for fair comparison.
To  quantitatively evaluate each method, we select Peak Signal to Noise Ratio (PSNR) and Structural Similarity Index (SSIM) as the reference criteria.

Table~\ref{table:REVIDE} shows the quantitative evaluation on the REVIDE test dataset.
We note that some prior-based image/video dehazing methods do not achieve higher PSNR/SSIM values that those of the original haze frames. some deep learning-based methods including single image dehazing and video dehazing  improve the quantitative results by training on the real haze video dataset.
The proposed multi-stage video dehazing method achieves much higher PSNR/SSIM values than other stat-of-the-art methods.
Fig.~\ref{fig:frame1} shows four hazy frames from the REVIDE test dataset and corresponding dehazed frames by different comparative methods.
We note that the dehazed frames by DCP method contain color distortion and haze residuals, as shown in Fig.~\ref{fig:frame1}(b).
The STMRF method is not able to remove haze from the high definition haze frames and the dehazed frames contain many haze residuals.
The single stage methods including EDVNet method and FastDVD method are not effect to remove haze in dense haze scenes.
The dehazed frames by our method contain fewer haze residuals and achieve higher PSNR/SSIM values than other comparative methods.

Fig.~\ref{fig:frame2} shows five consecutive adjacent frames from a real haze video and corresponding dehazed frames by different comparative methods.
The dehazed frames by our method contain fewer haze resudials and artifacts.

\subsection{Ablation Experiments}
To prove the effectiveness of the proposed deep video dehazing method, we develop several ablation experiments: (a) Only using two fusion stages; (b) Using two fusion stages and the refinement stage (the proposed method).
We use the same settings for each ablation experiment.
Table~\ref{table:ablation1} show the quantitative results of different baseline methods on the REVIDE test dataset.
We note that the PSNR/SSIM values of the proposed method are much higher that those of the original haze frames, which proves that our video dehazing method is able to remove haze from frames.
The PSNR/SSIM values of the baseline (b) are higher than those of the baseline (a), which proves that the refinement stage helps to improve dehazing performance.
Fig.~\ref{fig:ablation} shows three haze frames from the REVIDE test datset and corresponding dehazed frames by different baseline methods.
We note that the dehazed frames by our method achieve higher PSNR/SSIM values and holds fewer haze residuals and artifacts.

\begin{table}[!htp]
	\footnotesize
	%\scriptsize
	\begin{center}
		\caption{Quantitative evaluations of different baseline methods on the REVIDE test dataset.}
		%\vspace{-0.3cm}
		\begin{tabular}{cccc}
			\toprule
			Methods&Input& (a) & (b)\\
			\hline
			PSNR&15.05 &22.12  &22.69   \\
			SSIM&0.770 &0.862  &0.875    \\
			\bottomrule		\end{tabular}
		%\vspace{2mm}
		\label{table:ablation1}
		%\vspace{-0.5cm}
	\end{center}
\end{table}
%%%%%%%%%%%%%%%%%%

%------------------------------------------------------------------------
\section{Conclusion}
In this paper, we propose a progressive video dehazing network fusing the alignment and restoration processes.
The alignment process aligns consecutive neighboring frames stage by stage without using the optical flow estimation.
The restoration process is not only implemented under the alignment process but also  uses a refinement network to improve the dehazing performance of the whole network.
The proposed networks include four fusion networks and one refinement network.
To decrease the parameters of networks, three fusion networks in the first fusion stage share the same parameters.
Extensive experiments demonstrate that the proposed video dehazing method achieves favorable performance against the-state-of-art methods.

%\section*{Acknowledgements}
%This work has been supported in part by the National Key R\&D Program of China (No. 2018AAA0102002), the National Natural Science Foundation of China (Nos. 61922043, 61925204, 61872421), and the Natural Science Foundation of Jiangsu Province (No. BK20180471).

%\appendices
%% you can choose not to have a title for an appendix
%% if you want by leaving the argument blank
%\section{}

%\section*{Acknowledgment}

\ifCLASSOPTIONcaptionsoff
  \newpage
\fi

\bibliographystyle{IEEEtran}
\bibliography{egbib}
%\begin{thebibliography}{1}
%\end{thebibliography}



%\vfill


\end{document}